\documentclass{ecai} 
\usepackage[dvipsnames]{xcolor}

\usepackage{latexsym}
\usepackage{amssymb}
\usepackage{amsmath}
\usepackage{amsthm}
\usepackage{booktabs}
\usepackage{enumitem}
\usepackage{graphicx}
\usepackage{color}
\usepackage{subcaption}
\expandafter\def\csname ver@subfig.sty\endcsname{}

\usepackage{subfig}
\usepackage{graphicx}
\usepackage{caption}
\usepackage{bm}

\usepackage{algorithm} 
\usepackage{algpseudocode}  
\usepackage{wrapfig}
\usepackage{makecell} 
\usepackage{multirow}
\usepackage{tikz}
\usepackage{hyperref}

\newcommand{\tikzxmark}{%
\tikz[scale=0.23] {
    \draw[line width=0.7,line cap=round] (0,0) to [bend left=6] (1,1);
    \draw[line width=0.7,line cap=round] (0.2,0.95) to [bend right=3] (0.8,0.05);
}}

\newtheorem{definition}{Definition}
\newcommand{\Lagr}{\mathcal{L}}

\newcommand{\BibTeX}{B\kern-.05em{\sc i\kern-.025em b}\kern-.08em\TeX}

\begin{document}


\begin{frontmatter}

\title{PUFFLE: Balancing Privacy, Utility, and Fairness in Federated Learning}


\author[A]{\fnms{Luca}~\snm{Corbucci}\thanks{Corresponding Author Email: luca.corbucci@phd.unipi.it}
}
\author[B]{\fnms{Mikko}~\snm{Heikkilä}
}
\author[B]{\fnms{David Solans}~\snm{Noguero}
} 
\author[A]{\fnms{Anna}~\snm{Monreale}
}
\author[B]{\fnms{Nicolas}~\snm{Kourtellis}
} 

\address[A]{University of Pisa}
\address[B]{Telefónica Research}


\begin{abstract}
Training and deploying Machine Learning models that simultaneously adhere to principles of fairness and privacy while ensuring good utility poses a significant challenge. The interplay between these three factors of trustworthiness is frequently underestimated and remains insufficiently explored. Consequently, many efforts focus on ensuring only two of these factors, neglecting one in the process. The decentralization of the datasets and the variations in distributions among the clients exacerbate the complexity of achieving this ethical trade-off in the context of Federated Learning (FL).
For the first time in FL literature, we address these three factors of trustworthiness.
We introduce PUFFLE, a high-level parameterised approach that can help in the exploration of the balance between utility, privacy, and fairness in FL scenarios. We prove that PUFFLE can be effective across diverse datasets, models, and data distributions, reducing the model unfairness up to 75\%, with a maximum reduction in the utility of 17\% in the worst-case scenario, while maintaining strict privacy guarantees during the FL training.
\end{abstract}

\end{frontmatter}


\section{Introduction}

In recent years, Machine Learning (ML) models have been deployed in a wide variety of fields. In earlier years, the emphasis was primarily on optimizing the utility of these models. Nowadays, the current and upcoming AI legislation~\cite{aiact, uk_whitepaper_2023,  biden2023executive} demand equal attention to other trustworthiness requirements \cite{trustworthyAIEU}, including but not limited to fairness and privacy. Fairness refers to the goal of reducing the algorithmic biases that models might exhibit in their predictions and/or representations. Privacy refers to the goal of keeping training data safe and preventing any information leakage through the use of the model. Although privacy and fairness are ideally sought at the same time, it is unfortunately challenging to find a balance between them~\cite{10.1145/3314183.3323847}.
Often, the outcome is a model exhibiting strong utility, measured in model accuracy, but deficient in terms of privacy protection and fairness. 
Conversely, efforts to address the model's unfairness may compromise privacy protection and utility requirements.

This problem becomes even more complex to solve when we consider a Federated Learning (FL) scenario. With FL, various clients aim to train a model without transferring their local training data to a central server. Instead, they only exchange the locally trained model with a central aggregator. The decentralization of the training datasets introduces several issues that do not need to be considered when working in a centralized learning context. 
In this paper, we investigate the interplay between privacy, fairness, and utility within the context of FL, and present a methodology that enables informed decisions on the ethical trade-offs with the goal of training models that can strike a balance between these three dimensions.

\noindent
\textbf{Our contributions:}
We propose PUFFLE, a first-of-its-kind methodology for finding an optimal trade-off between privacy, utility and fairness while training an FL model. Our approach, inspired by the method in~\cite{yaghini2023learning} for a centralised setting, aims to train a model that can satisfy specific fairness and privacy requirements through the active contribution of each client. In particular, the clients incorporate an additional regularization term into the local loss function during the model's training to reduce unfairness. Furthermore, to ensure the privacy of the model, each client employs Differential Privacy (DP)~\cite{10.1007/11787006_1} by introducing controlled noise during the training process. We summarize our key contributions below:

\begin{itemize}
    \item We show how to mitigate model unfairness under privacy constraints in the FL setting.
    \item Unlike the method proposed in~\cite{yaghini2023learning}, PUFFLE does not require access to any external public dataset.
    \item We propose an approach that does not require the clients to delve into the technical details of the methodology. Our approach automatically computes the necessary parameters based on the client's fairness and privacy preferences, ensuring a high level of parameterisation compared to past methods. 
    \item We offer both local and global computation of the fairness metrics. The former considers the fairness of the local clients on the local datasets. The latter, instead, considers the fairness of the aggregated, global model.
    \item We validate our approach through comprehensive experimentation using three datasets of two different modalities, two different model architectures and three real-life distributions. Moreover, we examine various combinations of privacy and fairness preferences to showcase the impact on model utility. With model utility, in this paper, we mean model accuracy.
    \item We release PUFFLE's source code to ensure reproducibility. \footnote{Code available at: \url{https://github.com/lucacorbucci/PUFFLE}}.
\end{itemize}

\section{Related work}
\label{sec:related}
Training a single model that achieves fairness and privacy objectives is always a challenging task~\cite{10.1145/3314183.3323847} because of the incompatibility in the simultaneous use of these trustworthiness requirements. Several papers have demonstrated how using privacy mitigation techniques could lead to more unfair models w.r.t. the underrepresented groups~\cite{bagdasaryan2019differential, kuppam2020fair, esipova2023disparate}. Similarly, employing fairness mitigation techniques can increase the privacy risk for involved clients~\cite{chang2021privacy}. 

There is a large body of research on techniques to mitigate the unfairness of models trained in centralised learning~\cite{kamiran2012data, zemel2013learning, feldman2015certifying, hardt2016equality, calmon2017optimized, pleiss2017fairness}. One idea, initially proposed in~\cite{6137441} with a logistic regression model, is the use of a dedicated regularization term. Recently, Yaghhini et al.~\cite{yaghini2023learning} extended this method to train models that could be both private and fair. They modified the standard Differentially Private Stochastic Gradient Descent (DP-SGD) algorithm~\cite{Abadi_2016}, which is itself a variant of Stochastic Gradient Descent (SGD). In their work, they introduced two methods: FairDP-SGD and FairPATE. The goal is to train ``impartial'' ML models that do not favour one goal over another. They recover a Pareto Frontier to show the trade-off between the different objectives. The method in~\cite{yaghini2023learning} has been the primary inspiration for our work. However, in our setting, data decentralization requires managing the presence of many clients with different distributions, preventing a direct application of the method in~\cite{yaghini2023learning}. Moreover, the variety of fairness definitions leads to different mitigation strategies to achieve different goals.

In FL, the literature firstly focused on fostering \textit{Accuracy parity}, i.e., achieving consistent model performance across clients regardless of their data distribution~\cite{9746300,donahue2023models, li2021ditto, wang2021federated,mohri2019agnostic,li2020fair}, then on mitigating \textit{Group Fairness}~\cite{dwork2011fairness}, i.e., achieving consistent model performance across different demographic groups represented in the dataset. Current literature on FL often addresses privacy and fairness as separate constraints, rather than considering them jointly~\cite{9378043,ezzeldin2022fairfed,cui2021addressing,abay2020mitigating}.
Recently, some works have started to consider the interplay between these two requirements. Borja Rodríguez-Gálvez et al.~\cite{mmdm} introduced the Fair and Private Federated Learning (FPFL) algorithm to enforce group fairness in FL. The algorithm extends the modified method of differential multipliers to empirical risk minimization with fairness constraints and uses DP to guarantee privacy protection. However, a limitation is that clients can only perform a single local step before sharing the model with the server, increasing communication costs and slowing down the training. Our solution relies on a more advanced FL aggregation algorithm, allowing multiple local steps before model sharing, thus reducing communication costs.

A two-step approach proposed in~\cite{padala2021federated} tackles this problem training a fair model that balances fairness and utility. Then, a second model with DP guarantees aligns its predictions with the fair predictions of the first one. However, focusing exclusively on scenarios with few clients overlooks potential issues with multiple clients involved in training. Instead, our solution also works as clients increase addressing problems that may occur. A similar solution has been proposed in ~\cite{10.1007/978-981-99-9614-8_8}.
First, the clients train a fair proxy model and then a privacy-protected model based on the proxy. However, this approach shares similar limitations to~\cite{padala2021federated}. Additionally, its applicability is restricted to scenarios involving convex and Lipschitz loss functions. Furthermore, a few crucial details regarding privacy accounting mechanisms are missing, making it challenging to evaluate the privacy guarantees.

A different family of solutions to mitigate the model unfairness consists of making the model fair either in a pre-or a post-processing phase. Sikha Pentyala et al.~\cite{pentyala2022privfairfl} introduced a pre-processing and a post-processing algorithm to reduce the unfairness in FL while preserving the client's privacy by adopting DP during the training. Abay et al.~\cite{abay2020mitigating} presented a similar approach using a pre-processing method to mitigate the model unfairness. However, to the best of our knowledge, none of these methods allow the user to set a desired unfairness level that the model should exhibit at the end of the training. Table~\ref{tab:rw} summarizes the main papers that addressed the issue of fairness and privacy in FL, and how PUFFLE goes beyond these papers.

\section{Background}

\begin{table}[t]
    \centering
    \resizebox{\columnwidth}{!}{
    \begin{tabular}{ccccc}
        \toprule
        \textbf{Fairness Notion} & \textbf{Fairness Target} & \textbf{Privacy} & \textbf{Utility} & \textbf{Reference} \\
        \midrule
        Client Fairness & \tikzxmark & \checkmark & \checkmark  & ~\cite{9746300}~\cite{donahue2023models}~\cite{li2021ditto}~\cite{wang2021federated}~\cite{mohri2019agnostic}~\cite{li2020fair} \\\midrule
         
         \multirow{3}{*}{Group Fairness} & \tikzxmark & \tikzxmark &\checkmark &~\cite{9378043}~\cite{ezzeldin2022fairfed}~\cite{cui2021addressing}~\cite{abay2020mitigating}\\ \cmidrule{2-5}
          & \tikzxmark & \checkmark &\checkmark  &~\cite{mmdm}~\cite{pentyala2022privfairfl}~\cite{padala2021federated}~\cite{10.1007/978-981-99-9614-8_8} \\ \cmidrule{2-5}
          & \checkmark & \checkmark &\checkmark & PUFFLE (present study) \\
          \bottomrule
    \end{tabular}
    }
    \caption{A summary of literature that studied the interplay between fairness and privacy in FL.}
    \label{tab:rw}
\end{table}

\subsection{Federated Learning}

Federated Learning (FL)~\cite{mcmahan2023communicationefficient} is a technique for training ML models while minimizing the sharing of information required for the learning process. In FL, $K$ clients train a global model $\theta$ for a total number of \textit{R} rounds without sharing the local datasets $D_{k}$ of size $n_k$ with a central server. The orchestration of the model training is managed by a server $S$ that selects a subset $\chi$ of clients for each round $r \in [0, R]$.
Each client $k \in \chi$ executes $E$ local training steps on their training dataset $D_k$. The server is also responsible for the aggregation of the models trained by the clients. A possible aggregation algorithm is Federated-SGD (FedSGD)~\cite{mcmahan2023communicationefficient} that assumes a number $E=1$ of local gradient update steps. In this case, at the end of the local step, the selected clients $\chi$ share with the server the gradient $g_k = \nabla \Lagr(\theta_k, b_i)$ of the model $\theta_k$ computed on the batch $b_i$. The server updates the global model $\theta_{r+1} \leftarrow \theta_r - \eta \sum^{C}_{k=1} \frac{n_k}{n}g_k$ where $\eta$ is a fixed learning rate and $n = \sum_{k=1}^{C} n_k$. In this work, we assume the use of Federated Average (FedAvg)~\cite{mcmahan2023communicationefficient}, a more advanced aggregation algorithm that improves the performance of FedSGD. In this case, the clients $\chi$ can perform more than a single local step before communicating with the server. Each client $k \in \chi$ updates the local model $\theta_{k} = \theta_{k} - \eta \nabla \Lagr(\theta_k, b_i)$, then, after $E$ local steps, it shares the model $\theta_k$ with the server that will aggregate all the received $\theta_{k}$ producing the model $\theta_{r+1}$ for the next round $\theta_{r+1} \leftarrow \sum_{k=1}^{C} \frac{n_k}{n} \theta_{k}$.

Despite the original goal of protecting clients' privacy, it has been proven that FL is vulnerable to various attacks~\cite{10190537,lyu2020threats,boenisch2023reconstructing}. Therefore, privacy-preserving techniques like DP are usually applied to guarantee increased client protection.

\subsection{Fairness in ML}

The problem of training fair ML models has received growing attention in the past few years. In literature, various fairness definitions and measures exist to evaluate ML model unfairness~\cite{10.1145/3194770.3194776,10.1145/3457607}. 

In this paper, we focus on \textit{Group Fairness} in the FL setting aiming to train a model that is fair with all the different demographic groups represented in the dataset, defined by sensitive attributes like gender or race. In FL, \textit{Group Fairness} can be assessed either by aggregating results at the individual client level or at the server, where client prediction statistics are aggregated. Among the growing number of fairness metrics available in the state of the art, we use Demographic Parity~\cite{dwork2011fairness}, as previously done in~\cite{yaghini2023learning} for the same goal in centralised learning. However, we would like to emphasize that PUFFLE can be used with any differentiable fairness metric, such as Equalised Odds \cite{pleiss2017fairness}, and Error Rate Difference \cite{pleiss2017fairness}.
\begin{definition}\label{def:dem_parity}
Demographic Parity is a notion of \textit{Independence}~\cite{barocas2017nips} that requires the probability of a certain prediction to be independent of the sensitive group membership. For instance, a sensitive group could be the gender of the sample taken into account. More formally, the Demographic Parity is defined as:
\begin{equation}
\mathbb P(\hat{Y}=y \mid Z=z) = \mathbb P(\hat{Y}=y \mid Z \neq z)
\end{equation}
\end{definition}
\noindent
In other words, the ML model should produce similar success rates for all different groups: the model is fairer as the probabilities become more similar.

Definition ~\ref{def:dem_parity} can be rewritten to represent the difference between the probabilities of obtaining the same output given different sensitive groups:

\begin{definition}
Demographic Disparity $\Gamma(y, z)$ is the difference between the probability of predicting class $y$ for samples with sensitive value $z$ and the probability of predicting class $y$ for samples with sensitive value different than $z$: 
\begin{equation}
\Gamma(y, z) = \mathbb P(\hat{Y}=y \mid Z=z) - \mathbb P(\hat{Y}=y \mid Z \neq z)
\end{equation}
\end{definition}

\textbf{Fairness through Regularization:} The concept of leveraging regularization to mitigate the unfairness of an ML model has previously been introduced in~\cite{yaghini2023learning}. In their approach, they proposed computing an additional loss called Demographic Parity Loss ($DPL$) and using it as a regularization term. $DPL$ is an extension of Demographic Parity which is computed in the following way:

\begin{equation} \label{eq:DPL}
DPL(\theta; D) = max_y max_z \Gamma(y, z)
\end{equation}

where $\Gamma(y,z)=\bigl\{\frac{|\{\hat{Y}=y, Z=z\}|}{|\{Z=z\}|} - \frac{|\{\hat{Y}=y, Z \neq z\}|}{|\{Z \neq z \}|}\bigr\}$, $D$ is the dataset whose $DPL$ is to be calculated, $\hat{Y}=y$ is the prediction of the private model $\theta$ for a sample and $Z$ is the sensitive value of a sample, i.e., the attribute towards which the model may be biased. For instance, a sensitive value could be the gender or the area of origin of an individual.
For each mini-batch $b_{i}$, the regularization term is summed to the classic loss function of the model $\Lagr(\theta_t, b_i) + \lambda DPL(\theta_t; D_{public})$.
To balance between the original and the regularization loss, \cite{yaghini2023learning} multiplied the regularization loss by a fixed parameter $\lambda$ set at the beginning of the training and maintained constant throughout the entire process. In their methodology, they tried different $\lambda$ values and then selected the optimal model based on the trade-off between privacy, utility, and fairness.

Although the idea of using regularisation to reduce unfairness in the model is effective, it cannot be directly applied within the context of FL. This is due to the decentralization of the datasets across clients, which may result in local client distributions where the $DPL$ cannot be computed. Furthermore,~\cite{yaghini2023learning} assumed the existence of a public dataset $D_{public}$ with a distribution akin to the training dataset, employed for calculating the $DPL$ at each training step. The assumption of the existence of a public dataset $D_{public}$ and the need to select a $\lambda$ parameter contribute to the level of complexity, making the approach unintuitive, and unsuitable for real-world scenarios.
On the contrary, PUFFLE solves all these issues making the approach proposed in~\cite{yaghini2023learning} not only compatible with FL but also improving interpretability and parameterization.

\subsection{Differential Privacy}

Differential Privacy (DP) is a formal definition of privacy~\cite{Dwork_McSherry_Nissim_Smith_2006, Dwork_Kenthapadi_McSherry_Mironov_Naor_2006}:

\begin{definition}
\label{def:dp}
Given $\varepsilon > 0$ and $\delta \in [0,1]$, 
a randomised algorithm $\mathcal{M}: \mathcal X \rightarrow \mathcal O$ is $(\varepsilon,\delta)$-DP, if for all neighbouring $x,x' \in \mathcal X$, $S \subset \mathcal O$, 
    \begin{equation}
        \mathbb P[\mathcal{M}(x) \in S] \leq e^{\varepsilon} \mathbb P [\mathcal{M}(x') \in S] + \delta .
    \end{equation}
\end{definition}

We use the so-called ``add/remove neighbourhood'' definition, i.e., $x,x' \in \mathcal X$ are neighbours if $x$ can be transformed into $x'$ by adding or removing a single element. 
For privacy accounting, we use an existing accountant~\cite{yousefpour2022opacus} based on R\'enyi DP~\cite{Mironov_2017}, a relaxation of DP (Definition~\ref{def:dp}).

\textbf{Local and Central DP:} In the FL setting, an important detail to consider is when and by whom DP is applied. A common high trust-high utility approach is called Central Differential Privacy (CDP)~\cite{Dwork_McSherry_Nissim_Smith_2006}: $K$ clients share raw data with an aggregator $S$ who is responsible for aggregating the data and then applying DP to the result. This approach has the advantage of maintaining good utility since we introduce a limited amount of noise into the result. However, all the clients have to trust that $S$ will adhere to the established protocol. In contrast, Local Differential Privacy (LDP)~\cite{4690986} does not assume that the clients trust $S$. Instead of sharing raw data, clients individually apply DP before sharing it with the aggregator. The drawback is that the total noise introduced during the process is larger than with CDP. In this paper, we assume that the clients might not trust the aggregator, and hence we adopt LDP.

\subsection{Privacy-Preserving ML}

Training ML models require a large amount of data. These data may contain sensitive information regarding the individuals whose data were collected. Hence, the model trained on these data could leak sensitive information. DP is a common solution for mitigating privacy risks in ML training. One standard algorithm to train DP ML models is DP-SGD~\cite{Song_Chaudhuri_Sarwate_2013, Abadi_2016}, a modification of SGD. DP-SGD differs from SGD in the way gradients are calculated: given a random sample of training data $L_{e}$, it computes the gradient of each $i \in L_{e}$. These gradients are clipped and then some noise is added to guarantee privacy protection. Since this is a well-known algorithm from literature, its pseudocode is reported in Algorithm ~\ref{alg:dp-sgd}, Appendix ~\ref{sec:appendix:dp-sgd}.

\textbf{Differentially Private FL:} In the context of a model trained using FL, LDP can be applied in different ways depending on what you want to protect. Common choices include, e.g., instance- or sample-level privacy, where the objective is to protect the privacy of individual samples within the training dataset, and user-level privacy, when the DP neighbourhood considers the entire contribution from a single client. In our paper, we aim to ensure \textit{instance-level DP}.

\section{Fair \& Differentially Private Federated Learning}

Our proposed methodology aims to incorporate DP and a fairness regularization term into the training process of an FL model. The goal is twofold: to mitigate the privacy risks encountered by clients involved in the model training, and concurrently, to diminish the algorithmic biases of the model increasing its fairness. Being in an FL setting, we cannot directly apply the techniques utilized in centralized learning. The decentralization of data introduces several challenges that must be considered and addressed when implementing a solution like the one we propose. Our final goal is to offer an easy-to-use approach to train fair and private FL models that can balance fairness, privacy and utility meeting pre-defined trustworthiness requirements. We tested our approach on multiple datasets, proving that it can be effective regardless of the data modality, distribution, and model architecture. We report in Table~\ref{tab:symbols} a list of the symbols used in the paper to increase clarity and readability.

\begin{table*}[]
    \centering
    \captionsetup{justification=justified}
    {\small
    \begin{tabular}{|c|c|c|c|}
        \hline
        \textbf{Symbol} & \textbf{Meaning} & \textbf{Symbol} & \textbf{Meaning} \\
        \hline
        K & FL Clients & R & FL Rounds\\ \hline
        $D_k$ & Dataset of each client $k \in K$ & $n_k$ & Size of the dataset $D_k$ \\ \hline
        S & Server &  $\chi$ & Subset of selected clients for an FL Round\\ \hline
        $\omega$ & Statistics shared by the clients & E & Local training steps of clients\\ \hline
        $\theta$ & Trained model & $\eta$ & Learning Rate\\ \hline
        Z & Sensitive Value & $\lambda$ & Lambda Used for regularization\\ \hline
        $\mathcal{M}$ & Randomised Algorithm used in DP & $D$  & Training Dataset\\ \hline
        $\varepsilon$ & Privacy Budget of DP & $q$ & data subsampling probability\\ \hline
        $\delta$ & Probability of DP compromise & $L_e$ & Random sample of training data \\ \hline
        $\sigma$ & Noise parameter for Differential Privacy & B & clipping value\\ \hline
        T & Target disparity & $\rho$ & Regularization Parameter\\ \hline    
    \end{tabular}
    }
    \caption{Table of notations}
    \label{tab:symbols}
\end{table*}

\begin{algorithm}
\caption{PUFFLE (Client-Side)}
\label{alg:client}
{\small
\begin{algorithmic}[1]
\Require $E$ Number of local steps. $P$ prob. $\langle$\textit{$z$, $y$}$\rangle$ combinations, $z$ sensitive value, $y$ output of the model. $D_{k}$ train dataset. Target Disparity $T$. Target  $\varepsilon$ and $\delta$. $\sigma_{1}$, $\sigma_{2}$ and $\sigma_3$ noise multipliers based on $\epsilon$ to guarantee DP. $momentum$ hyperparameter.  $velocity=0$ Initial velocity value. $r$ fl round.
\State Receive model $\theta$ from server and initialize local model $\theta_k \leftarrow \theta$
\State $\lambda_{0} \leftarrow 0$ if $(T - DPL(\theta_k, D_{k})) \geq 0$ or ($r$=0) else 1 \label{line:initial_lambda}  
\For{$e \in E$}
\For {$b_{i} \in D_{k}$} \# use Poisson sampling with probability $q$
\State $DPL(\theta_k, b_{i}) \leftarrow max_{z} max_{y} \mathbb P(\hat{Y}=y \mid Z=z) - \mathbb P(\hat{Y}=y \mid Z \neq z)$. \# Use the statistics $P$ if necessary. \label{line:reg_loss}
\State $g_{dpl} \leftarrow \nabla DPL$   \label{line:dpl_gradient}
\State $g_{loss} \leftarrow \nabla \Lagr(\theta_k, b_i)$\label{line:gradient}
\State $g \leftarrow (1-\lambda)\times g_{loss} + (\lambda) \times g_{dpl}$ \label{line:sum}
\State DP-SGD: clip the per-sample gradients and add noise $\sigma_{1}$
\State $\Delta \leftarrow T - (DPL(\theta_k, b_{i}) + \mathcal{N}(0, \sigma_{2}^2))$ \label{line:delta}
\State $velocity \leftarrow momentum \times  velocity + \Delta$ \label{line:momentum}
\State $\lambda_{t+1} \leftarrow clip(\lambda_{t} - \rho \times velocity$) \label{line:new_lambda}
\EndFor
\EndFor
\State $R_k \leftarrow compute\_metrics(\theta_k, D_{k})$
\State $\omega_k \leftarrow$ compute\_statistics($\theta_k$, $d_{k}$) + $\mathcal{N}(0, \sigma_{3}^2)$
\State Send \{$\theta_k$, $\omega_k$, $R_k$\} to the server
\end{algorithmic}
}
\end{algorithm}

\subsection{Fair Approach}

The approach we adopt to mitigate the unfairness of the model is based on the use of a regularization term represented by the $DPL$. The idea was originally proposed in~\cite{yaghini2023learning} in the centralized setting, but we need to apply several changes to make it suitable to an FL setting.
The client-side training algorithm that we propose (Algorithm~\ref{alg:client}) is a modification of the standard DP-SGD algorithm that adds a regularization term $DPL$ to the regular model loss. These two losses are weighted by a parameter $\lambda$. In our approach, $\lambda$ can be tuned during the hyperparameter search and remain fixed for the entire training on each node or, differently from the original paper, $\lambda$ can vary during the training (we provide more details next). In the latter case, at the beginning of the training, each client $k$ computes $\lambda_0$ (Line~\ref{line:initial_lambda}, Algorithm~\ref{alg:client}). In the first FL round, the model is random, resulting in random predictions and disparity typically close to 0, allowing $\lambda_0$ to be 0. Then, in the following FL rounds, we compute $\lambda_0$, we consider the difference between the target disparity $T$ and the $DPL(\theta_k, D_{k})$ computed using the model $\theta_k$ on the training dataset $D_{k}$. If this difference is positive then $\lambda_0 = 0$, else if the actual disparity is higher than the target, we fix its value to $\lambda_0 = 1$. Then, the classic training of the model can start. For each mini-batch $b_i$, sampled using Poisson sampling with probability $q$, each client computes a regularization term $DPL(\theta_k, b_{i}) = max_{z} max_{y} \mathbb P(\hat{Y}=y \mid Z=z) - \mathbb P(\hat{Y}=y \mid Z \neq z)$ (Line~\ref{line:reg_loss}, Algorithm~\ref{alg:client}). 

Furthermore, and in contrast to the approach proposed in~\cite{yaghini2023learning}, we do not rely on a public dataset $D_{public}$ for computing the $DPL$. The assumption of having a public dataset $D_{public}$ matching the distribution of the training data is often impractical in real-life scenarios. Therefore, we propose to compute the $DPL$ directly on the batches $b_i$ using the disparity of batch $b_i$ as a regularization term. Upon computing the regularization term, the algorithm proceeds to calculate the \textit{per-sample gradients} w.r.t. both the regularization term (Line~\ref{line:dpl_gradient}, Algorithm~\ref{alg:client}) and the regular model loss (Line~\ref{line:gradient}, Algorithm ~\ref{alg:client}).

Since we want to have a Differentially Private model, we use DP-SGD to train the model. This means that the gradients computed are \textit{per-sample gradients}. Therefore, to combine the regularization term with the regular loss, we sum the \textit{per-sample gradients} $g = (1-\lambda) \times g_{loss} + (\lambda) \times g_{dpl}$ (Line~\ref{line:sum}, Algorithm ~\ref{alg:client}). This is where $\lambda$ comes in. This parameter balances the weight of the regular loss with the weight of the regularization. Setting $\lambda \sim 1$ assigns more weight to the fairness regularization than to the standard model loss, resulting in an extremely fair model but with low utility. Conversely, opting for $\lambda \sim 0$ prioritizes utility over fairness in the model optimization process.
After this sum the regular DP-SGD algorithm is executed, the gradients are first clipped and the noise is added to guarantee DP.

\subsection{Tunable \texorpdfstring{$\lambda$}{Lg}} 

As mentioned earlier, interpreting the parameter $\lambda$ is not straightforward as it provides no clear insight into the resulting model disparity after training. Instead of using the same fixed parameter $\lambda$ for each client, our approach uses a more interpretable parameter $T$ which represents the demographic disparity target for the trained model. Ranging between 0 and 1, this parameter is easily interpretable and can be selected based on current and upcoming AI legislations~\cite{madiega2021artificial, uk_whitepaper_2023, biden2023executive}. Consequently to the choice of target disparity that the final trained model should exhibit, each client calculates its own $\lambda$ based on $T$ and the local model disparity. The value will increase or decrease throughout the training process to make the trained model meet the target disparity value $T$. Our Tunable $\lambda$ will range between 0 and 1: the more the model is unfair, the higher the $\lambda$ will be to reduce the unfairness. 
The value of the Tunable $\lambda$ is adjusted after each local training step on each client $k$. 
First of all, the difference $\Delta$ between the desired disparity $T$ and the actual disparity is computed $\Delta = T - (DPL_{b_{i}} + \mathcal{N}(0, \sigma_{2}^2))$ (Line~\ref{line:delta} and~\ref{line:momentum}, Algorithm ~\ref{alg:client}). When computing the $\Delta$ we introduce noise from a Gaussian Distribution $\mathcal{N}(0, \sigma_{2}^2)$ to guarantee DP in the computation of the next $\lambda$. Using the $\Delta$ we can compute the parameter $velocity$ which is 0 in the first batch and then is updated in the following way: $velocity = momentum \times velocity + \Delta$ (Line~\ref{line:momentum}, Algorithm ~\ref{alg:client})
where \textit{momentum} is a hyperparameter learned during the hyperparameter tuning phase.
Then, we can finally compute the next $\lambda$, clipping it to $[0,1]$: $\lambda_{t+1} = clip(\lambda_{t} - \rho \times velocity)$ (Line~\ref{line:new_lambda}, Algorithm~\ref{alg:client}) where $\rho$ is a learned hyperparameter indicating how aggressively we want to make the model fair.

\subsection{Sharing Local Distribution Statistics} 
\label{sec:sharing_local_stats}
In a real-life scenario, clients involved in the training process usually have different data distributions, posing a challenge that requires a solution before applying the approach proposed in \cite{yaghini2023learning} in an FL setting. For instance, consider $K$ clients aiming to train a model predicting individuals' eligibility for bank loans. Each client's dataset includes data regarding different individuals, including sensitive information like gender. We must prevent the training of a model that is biased toward a specific gender. DPL regularization offers a potential solution, challenges arise in an FL setting where some clients possess data exclusively for one gender.
In such cases, when these clients compute the $DPL$ using Formula~\ref{eq:DPL}, they can only calculate either the part of the formula where $Z=Male$, or the part where $Z=Female$. To handle this case, we propose a solution that involves the sharing of some statistics from the client to the server. Without loss of generality, we can assume a classification problem with a binary label and binary sensitive feature, at the end of each training round, each client computes the following statistics and shares them with the server: the number of samples with $Z=1$ and $\hat{Y} = 1$ and the number of samples with $Z=0$ and $\hat{Y} = 1$.

The shared information is sufficient for the server to compute all the other necessary statistics and the probabilities $P(\hat{Y}=y \mid Z = z)$. The clients do not need to compute and share the number of samples with sensitive values $Z=1$ and with $Z=0$ for each FL round, because these values do not change during the training; thus, sharing them just at the beginning of the training process is sufficient. This, along with the ability to compute the missing statistics, allows us to decrease the amount of privacy budget required for this part of the process. After computing the missing statistics (Line~\ref{line:missing_stats}, Algorithm~\ref{alg:server}) and aggregating them (Line~\ref{line:aggregation}, Algorithm~\ref{alg:server}), the server shares these values with the clients selected for the next FL round. To clarify, and without loss of generality, assuming a dataset with a binary label $Y$ and sensitive value $Z$, each selected client will receive $P(\hat{Y}=1 \mid Z = 1)$, $P(\hat{Y}=1 \mid Z = 0)$, $P(\hat{Y}=0 \mid Z = 1)$, $P(\hat{Y}=0 \mid Z = 0)$. 
The selected clients that lack the necessary data to calculate the DPL can rely on the statistics shared by the server.

In addition to managing edge cases, client-shared statistics play a crucial role in understanding the trained model's disparity. A global view of the model fairness involves aggregating the various counters from individual clients, allowing for the computation of demographic disparity as if the server had access to the entire dataset. Conversely, the opposite approach involves gaining a local understanding of demographic disparity by computing the mean of each client's demographic disparity. This dual approach enables a global view across the entire dataset and a focused client fairness examination.

\begin{algorithm}[] \caption{PUFFLE (Server-Side)}
\label{alg:server}
{\small
\begin{algorithmic}[1]
\Require $R$ Number of training rounds. $K$ number of clients. $Z$ possible sensitive values of client k. $Y$ possible labels of client k.
\State Initialize the model $\theta$
\For{$r \in R$}
\State Sample a group $\chi$ of clients, send them $\theta$. If $r > 0$ send $P$.
\For{$k \in K$}
\State Receive \{$\theta_k$, $\omega_k$, $R_k$\} {model, statistics, metrics}
\EndFor
\State $\theta_{r+1} \leftarrow \sum_{k=1}^{C} \frac{n_k}{n} \theta_{k}$
\State $Acc = \sum_{k=1}^{C} \frac{n_k}{n} Acc_{k}$ 

\State For each $\omega_k$ compute the missing statistics using $Z_k$ and $Y_k$ \label{line:missing_stats}
\State $\omega = \sum_{y \in Y, z \in Z} \omega_{k}[Y=y, Z=z]$ \label{line:aggregation}
\For{$z \in Z$ and $y \in Y$}
\State $P_{(z, y)} = \frac{\omega_{(Z=z, Y=y)}}{\omega_{(Z=z, Y=y)} + \omega_{(Z \neq z, Y=y)}}$
\EndFor
\State Compute the Global demographic disparity
\EndFor
\end{algorithmic}
}
\end{algorithm}

\subsection{Differentially Private Approach} To guarantee a privacy-preserving learning process we propose a solution which is $(\varepsilon, \delta)$-DP. Ensuring this requires the implementation of DP in three distinct phases of our process:
\begin{itemize}
    \item The model training must be differentially private. To this end, we adopt a local DP approach based on DP-SGD (Algorithm~\ref{alg:dp-sgd} in the Appendix). 
    Denote the privacy budget for this phase by ($\varepsilon_1, \delta_1$) and the corresponding noise needed to guarantee DP as $\sigma_{1}$. 
    \item The tunable $\lambda$ depends on the disparity of the model computed using the training dataset on each client. Therefore, the computation of this value must be differentially private.
    Let the budget for this phase be ($\varepsilon_2, \delta_2$) corresponding to a noise denoted as $\sigma_{2}$. 
    \item Sharing statistics with the server could potentially leak information about the model's predictions and the data distribution of each client. Therefore, we apply DP to shared statistics using a privacy budget ($\varepsilon_3, \delta_3$) that corresponds to a noise $\sigma_{3}$. 
    
\end{itemize}

For each phase, we use the well-known Gaussian mechanism~\cite{Dwork_Kenthapadi_McSherry_Mironov_Naor_2006} to provide privacy. To track the total privacy budget for each phase, we use the RDP accountant implemented in Opacus~\cite{yousefpour2022opacus}.

The privacy of model training follows immediately since we use DP-SGD with minor variations; the minibatch is selected via Poisson sampling with a given sampling ratio, and we still clip the per-example gradients and add noise distributed as $\mathcal N (0,B^2 \sigma_1^2)$ to each dimension (step 6-9, Algorithm~\ref{alg:client}). Therefore, the total number of local optimization steps gives the client-specific number of compositions for privacy accounting.

For updating $\lambda$ and for sharing the statistics, we can bound the sensitivity without clipping: since by definition $\lambda \in [0,1]$ and the statistics are essentially counts, the sensitivity under add/remove neighbourhood for both is equal to 1. 
For updating $\lambda$ and for sharing the statistics, we therefore add noise distributed as $\mathcal{N}(0, \sigma_2^2)$ and $\mathcal{N}(0, \sigma_3^2)$, respectively. In the case of $\lambda$, the resulting noisy value is clipped to $[0,1]$ (Line~\ref{line:new_lambda}, Algorithm~\ref{alg:client}), which does not affect DP due to the post-processing guarantees of DP~\cite{Dwork_Roth_2014}. The number of compositions for sharing the statistics is the total number of FL rounds where a given client is selected, while for $\lambda$ it matches the total number of local optimisation steps. Note that sharing the statistics does not use data sub-sampling while updating $\lambda$ uses Poisson sub-sampling.
From the basic composition theorem of DP~\cite{Dwork_Roth_2014} it follows that our protocol achieves $(\varepsilon, \delta)$-DP, where: $\varepsilon = \varepsilon_1 + \varepsilon_2 + \varepsilon_3$ and $\delta = \delta_1 + \delta_2 + \delta_3$.

\subsection{Complexity Analysis}
In this section, we analyze PUFFLE's complexity considering both the client side and the communication complexity.
\subsubsection{Local Computational Complexity}
From the perspective of local client complexity, DP-SGD is used to train the local models. While DP-SGD maintains the same theoretical time complexity as standard SGD, it is important to note that it is practically slower due to the explicit per-sample gradient computation requirement.
The computation of the fairness loss in PUFFLE does not introduce additional computational overhead. This is because the fairness loss calculation uses predictions already available from the training process. 
\subsubsection{Communication Complexity}
As explained in Section\ref{sec:sharing_local_stats}, PUFFLE requires the clients to share additional statistics with the server compared to the standard FedAvg. This causes an increase in the communication cost. For a classification task with $|\hat{Y}|$ possible classes and $|Z|$ sensitive groups, the additional information exchanged scales as $O(|\hat{Y}||Z|)$ per client per FL round. This extra communication is constant to the number of model parameters and the number of participating clients.

\section{Experimental Setup}
\label{sec:exp_setup}

\subsection{Datasets}
\label{sec:dataset}
We demonstrate the effectiveness of PUFFLE across various models, datasets, and data distributions, showing the efficacy on both tabular and image datasets. This highlights the methodology's robustness regardless of all these factors, proving its applicability in different scenarios. We conduct tests using three distinct datasets: CelebA~\cite{liu2015faceattributes}, Dutch~\cite{dutch} and ACS Income~\cite{ding2022retiring}.
$CelebA$ is an image dataset with more than 200.000 celebrity images with annotations. The annotations are crucial in our scenario since they contain information about the gender of the person depicted in the photos. In our experiment, we focus on a classification task, predicting whether the celebrity in the image is smiling or not. $Dutch$ is a tabular dataset collected in the Netherlands, providing information about individuals. Compared with other common fairness tabular datasets like Compas~\cite{Compas}, $Dutch$ is bigger, making it suitable for our cross-device FL setting. The objective here is to predict whether the participants' salary is above or below \$50,000. $ACS Income$ is also a tabular dataset built from the American Community Survey (ACS) with data coming from over all 50 states and Puerto Rico in 2018. The task involves predicting whether the samples in the dataset have a salary higher than \$50,000.
Since it is split into 51 parts based on the regions, it is a natural choice for FL experimentation allowing us to test PUFFLE on an inherently partitioned dataset.
Table~\ref{tab:dataset} provides a detailed description of the datasets.

\begin{table}[]
    \centering
    \resizebox{\columnwidth}{!}{
    \begin{tabular}{ccccc}
        \toprule
        \textbf{Dataset} & \textbf{Samples} & \textbf{Label} & \textbf{Sensitive Value} & \textbf{Clients} \\
        \midrule
        CelebA & 202.599 & Smiling & Gender & 150\\
        \hline
        Dutch & 60.420 & Salary (>=50K) & Gender & 150\\
        \hline
        ACS Income & 1.664.500 & Salary (>=50K) & Gender & 51\\
        \bottomrule
    \end{tabular}
    }
    \caption{Details of the dataset used for our experiments}
    \label{tab:dataset}
\end{table}

Table~\ref{tab:models} in Appendix \ref{sec:appendix:model_architecture} provides details about the model architecture used in our experiments. For CelebA, we use a Convolutional Neural Network (CNN) consisting of three convolutional layers (CL) with 8, 16, and 32 channels, respectively. We use kernel size of value 3 and the stride parameter with value 1. Each CL is followed by max pooling with kernel size 2 and stride value of 2. After the CL, we use a fully connected layer with 2048 hidden units, with ReLU as the activation function. For Dutch and ACS Income, instead, we use a simple neural network with a single linear layer to mimic the behaviour of a Logistic Regressor.
We used Flower~\cite{beutel2020flower} to simulate the FL process and developed the neural networks using PyTorch~\cite{paszke2019pytorch}.

In Dutch and CelebA experiments, we used 150 clients during the simulation. Initially, data was distributed using a representative diversity approach, maintaining the same ratio per node as in the dataset. Then, to exaggerate client unfairness, we increase the amount of data from a specific group of sensitive values and reduce the amount of data from the opposite group. The experiments reported here present an even more realistic scenario to prove how PUFFLE can address edge-case situations. Specifically, for the Dutch dataset, half of the clients lack samples from the group $\langle sensitive\ value = 1, label = 1 \rangle$. For CelebA, half of the clients have samples removed with $\langle sensitive\ value = 1\rangle$.
In the case of the ACS Income dataset, we leverage the natural division of this dataset into states using each of the 51 states as a client during the training. We add more information about the strategy used to split the dataset into clients in Appendix~\ref{sec:appendix:split}.

In the experiments, we assume to be in a cross-device scenario dividing clients into two groups: train and test. For Dutch and CelebA, we had 100 train clients and 50 test clients, while for Income we had 40 training clients and 11 test clients. 
On each FL round, we sample 30\% of the training clients with CelebA and Dutch while we select half of the training clients in the case of ACS Income Dataset. 
We tune all hyperparameters using Bayesian optimization, trying to maximize model utility while staying under the target disparity $T$.

To ensure the reproducibility of the results, we release the source code for reproducing all the experiments.

\subsection{Hyperparameter Tuning}
\label{sec:hyp_tuning}
The final goal of PUFFLE is to train models that offer good utility while mitigating the disparity. We defined the best hyperparameters as those that maximized the formula:
$$\psi = validation \ accuracy + \Delta$$
with $\Delta = 0$ when validation disparity $\leq$ $T$ and $\Delta = -\infty$ otherwise. 
The validation accuracy is computed independently by the validation clients on their local datasets and aggregated by the central server $S$ using a weighted average. By maximizing $\psi$, we incorporate both accuracy and disparity into a single formula ensuring that the hyperparameters lead to a model disparity below the fairness target $T$ at least on the validation set.
We performed a hyperparameter search using the Bayesian search method dividing the $K$ clients into 3 groups: train, validation and test. To ensure consistency throughout the hyperparameter search, we kept the set of test clients constant. On the contrary, we shuffled the assignment of clients to the training and validation sets for each of the different hyperparameter searches ran. For the Dutch and CelebA dataset, we used 150 clients: 50 as test clients, while the remaining 100 were split into 60\% training and 40\% validation clients. With ACS Income dataset, we had 51 clients in total: 11 used as test, 30 as train and 10 as validation. After completing the hyperparameter tuning, we retrained the model using the best hyperparameters, combining training and validation clients into a single training group.

\section{Experimental Results}
\label{sec:results}

In this section, we present the results of the application of PUFFLE on the CelebA Dataset. Corresponding results on the other datasets with PUFFLE can be found in Appendix~\ref{sec:appendix:results}.

\subsection{Utility vs. Fairness vs. Privacy Trade-offs}

To explore the interaction between these trustworthiness requirements, we examine combinations of two privacy targets ($\epsilon$, $\delta$) and five fairness disparities targets $T$. We follow the common approach in deep learning of using $\varepsilon \leq 10$~\cite{Ponomareva_2023}, we select as target $(\varepsilon=5, \delta=8 \times  10^{-4})$ and $(\varepsilon=8, \delta=8 \times 10^{-4})$ in the experiment with CelebA. The parameter $\delta$ is determined by considering the local parameter $\delta_{k} = \frac{1}{n_{k}}$ computed by each client $k \in K$ using the local dataset size $n_k$, as recommended in the literature~\cite{Ponomareva_2023}. In our method, $\delta$ is computed as the maximum $\delta_k$ among all clients, denoted as $max_{k} \delta_{k}$.

The target disparity $T$ is based on a percentage reduction from the baseline. We experiment with 10\%, 25\%, 50\%, 65\%, and 75\% reductions in disparity to demonstrate the efficacy of our methodology in achieving a trade-off, for both high and extremely low targets. Since the baseline model trained with FL on the CelebA dataset has a disparity of $\sim 0.17$, we set the target disparities for our experiment at $0.15$, $0.12$, $0.09$, $0.06$, $0.04$. Note that the parameter $T$ is only a desired value for the model disparity, unlike Differential Privacy, which provides formal guarantees w.r.t. training data, Demographic Parity does not guarantee any formal bound w.r.t the test data. However, our experiments empirically show that we can achieve test disparity close to our fairness target under diverse settings. Results presented here are the averages and standard errors of 5 executions per experiment. Each execution is conducted with a different seed for shuffling training clients, while test clients remain constant. For brevity, only results with 25\%, 50\% and 75\% reduction in disparity are shown; other privacy and target disparity combinations can be found in Appendix~\ref{sec:appendix:results}.

Figure~\ref{fig:celeba_1} provides a valuable insight into the intricate trade-off between fairness, utility, and privacy when the privacy requirement is ($\varepsilon=5, \delta=8 \times 10^{-4}$).
In Figures~\ref{fig:accuracy_1},~\ref{fig:accuracy_2} and ~\ref{fig:accuracy_3}, we present the test accuracy achieved by varying the target disparity. In Figures~\ref{fig:disparity_1},~\ref{fig:disparity_2} and~\ref{fig:disparity_3}, we present the corresponding achieved test disparity. In these plots, we show different, escalating configurations: 1) Baseline model without fairness or DP; 2) Model with Unfairness Reduction with Fixed $\lambda$; 3) Model with Unfairness Reduction with Tunable $\lambda$; 4) Model with DP; 5) Fair and Private model with Fixed $\lambda$; and 6) Fair and Private model trained with Tunable $\lambda$ (PUFFLE includes \#5 and \#6).
The horizontal line in the test disparity plots in Figures~\ref{fig:disparity_1},~\ref{fig:disparity_2} and~\ref{fig:disparity_3} represents the target disparity fixed in each of the experiments. Test Disparity is computed by the server using statistics shared by test clients, providing a global view of the aggregated model's disparity.

\begin{figure*}[t]
    \centering
    \captionsetup{justification=justified}
    \begin{subfigure}[b]{0.33\textwidth}
         \centering
         \includegraphics[width=\textwidth]{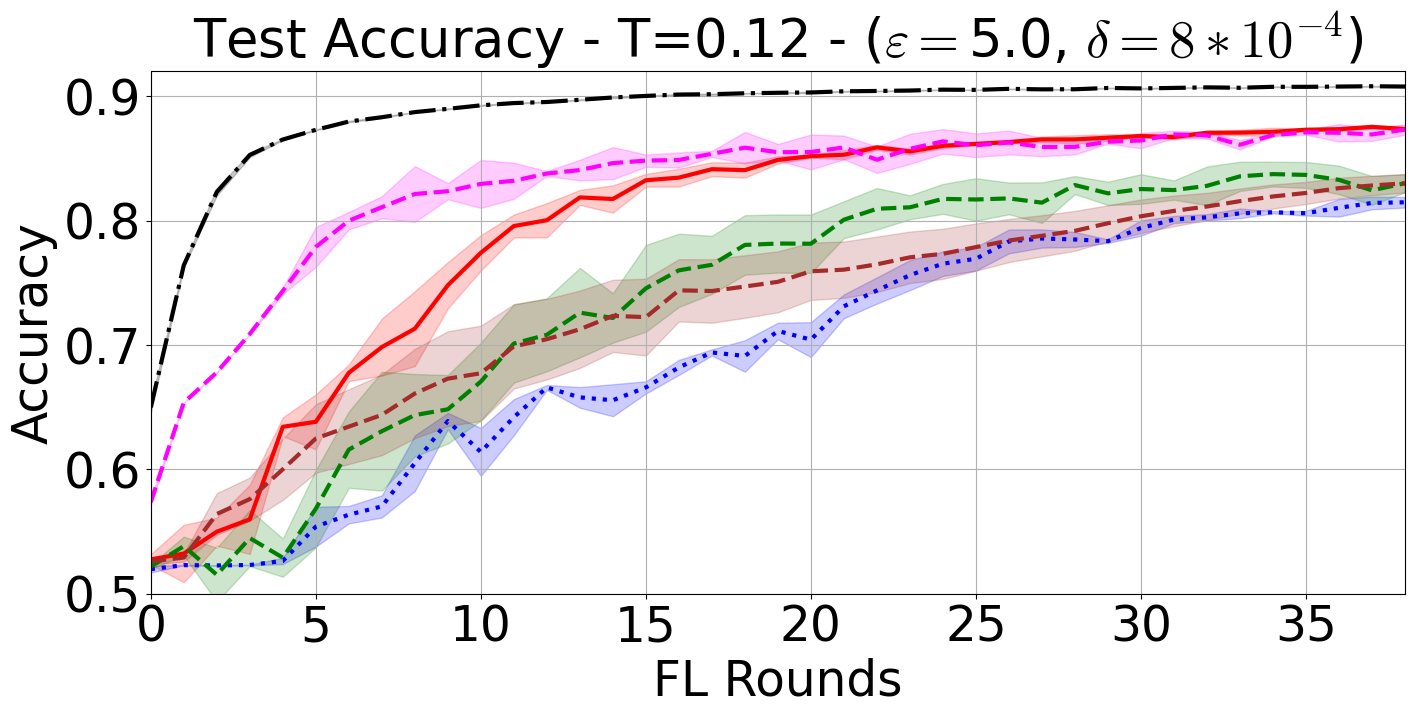}
         \caption{}
         \label{fig:accuracy_1}
     \end{subfigure}
     \hfill
      \begin{subfigure}[b]{0.33\textwidth}
         \centering
         \includegraphics[width=\textwidth]{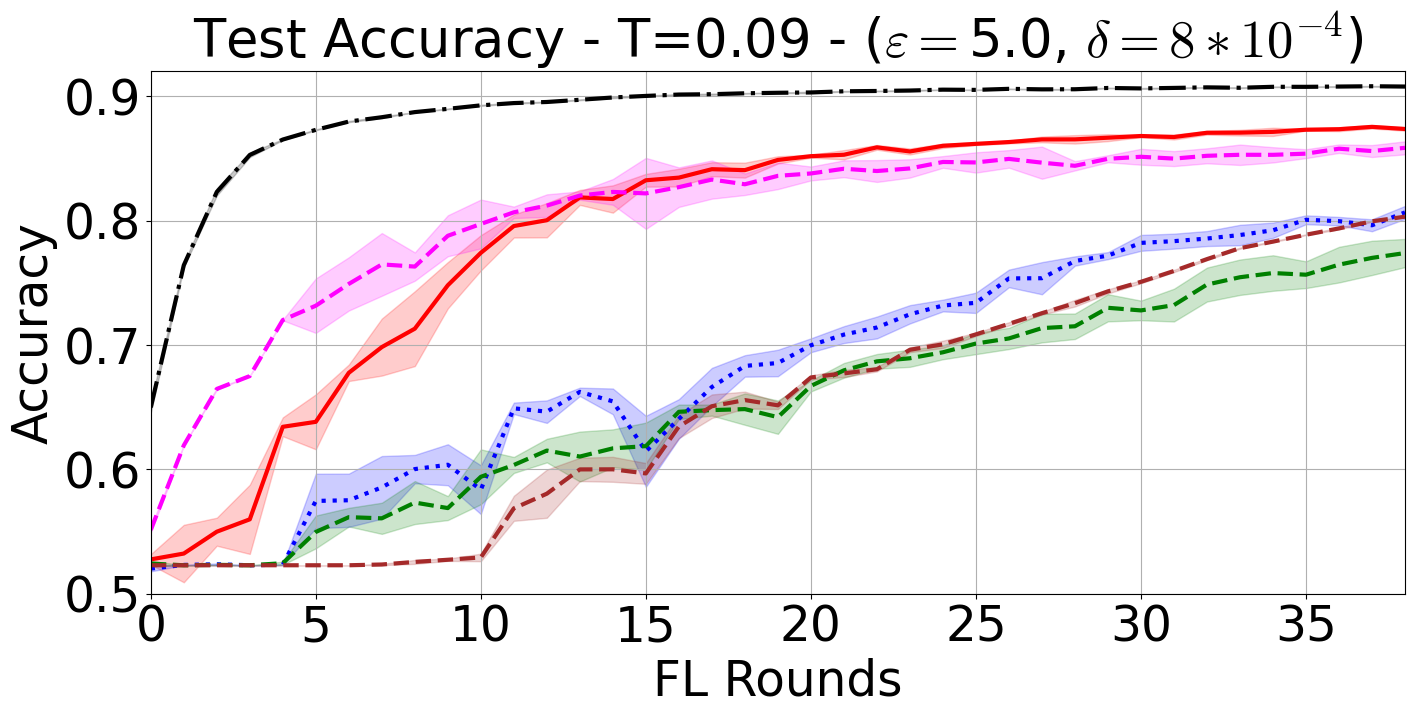}
         \caption{}
         \label{fig:accuracy_2}
     \end{subfigure}
     \hfill
     \begin{subfigure}[b]{0.33\textwidth}
         \centering
         \includegraphics[width=\textwidth]{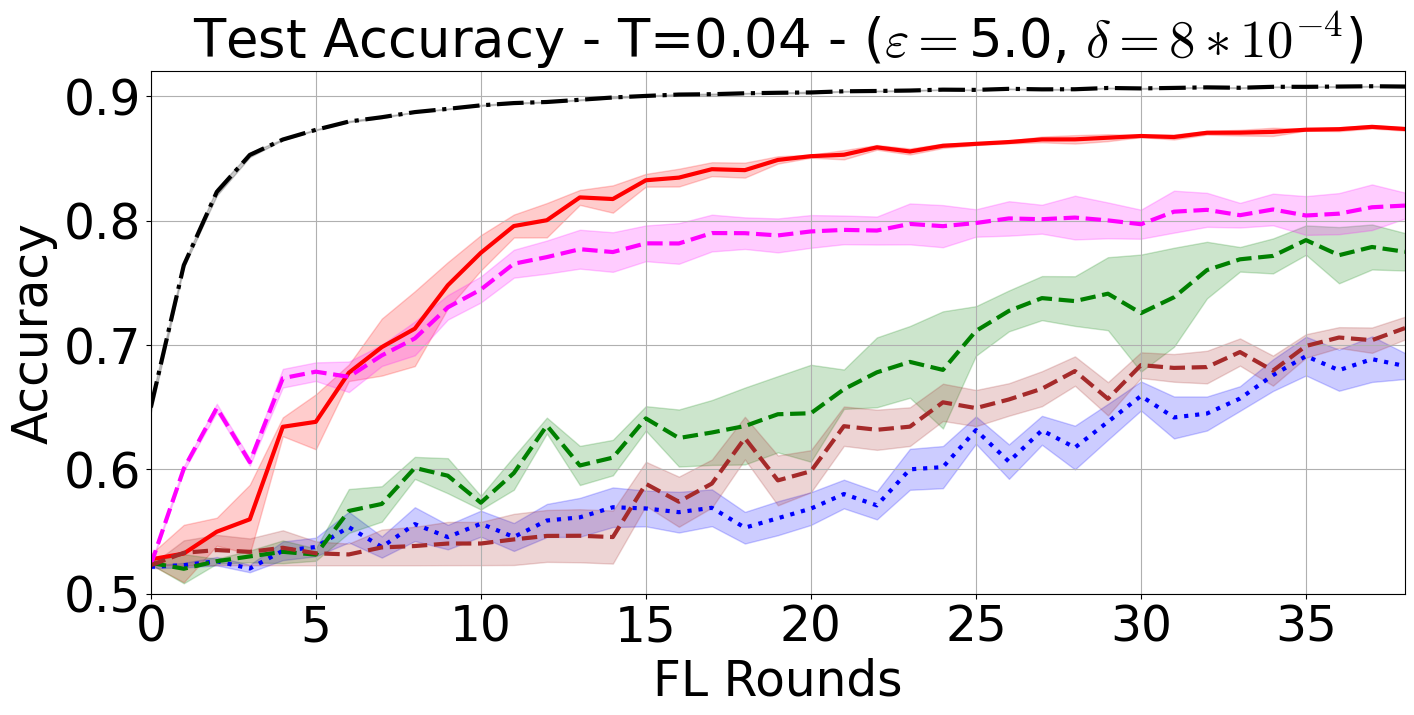}
         \caption{}
         \label{fig:accuracy_3}
     \end{subfigure}\\
    \vspace{0.30cm}
     \begin{subfigure}[b]{0.33\textwidth}
         \centering
         \includegraphics[width=\textwidth]{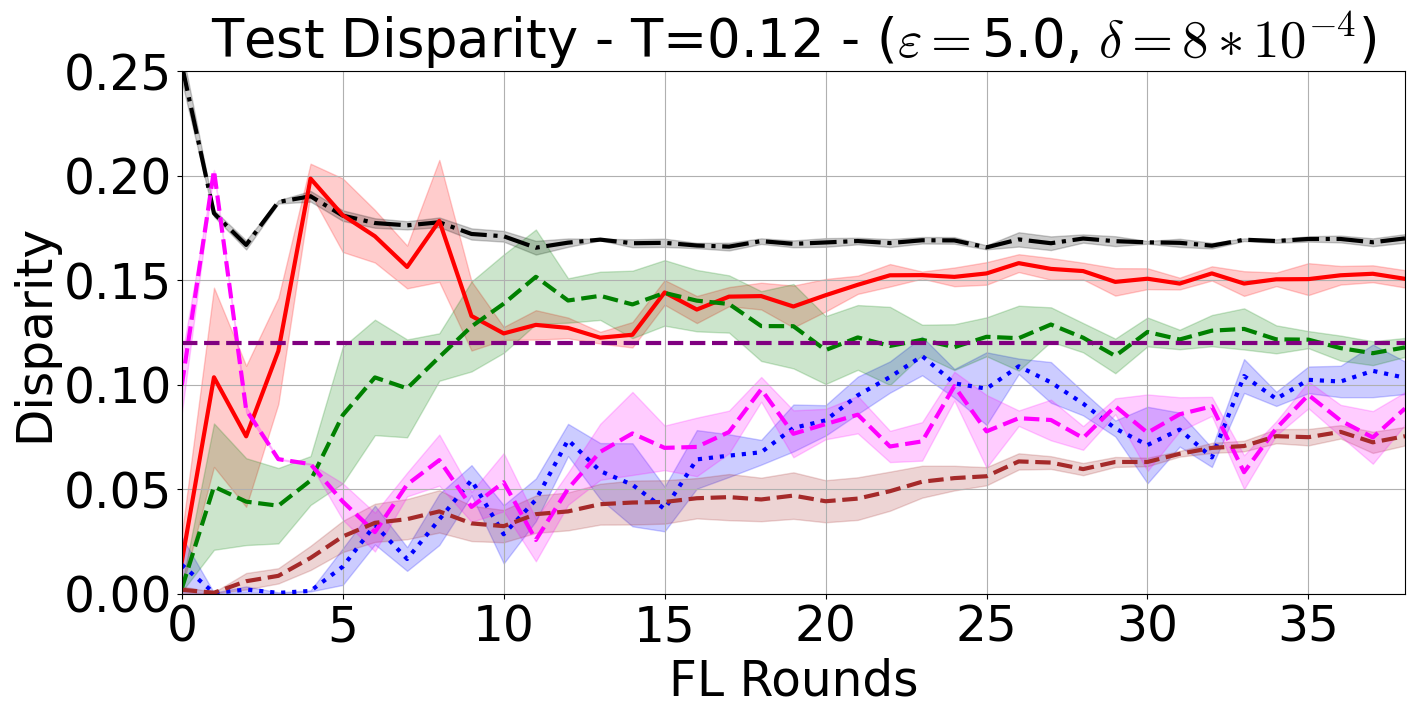}
         \caption{}
         \label{fig:disparity_1}
     \end{subfigure}
     \hfill
     \begin{subfigure}[b]{0.33\textwidth}
         \centering
         \includegraphics[width=\textwidth]{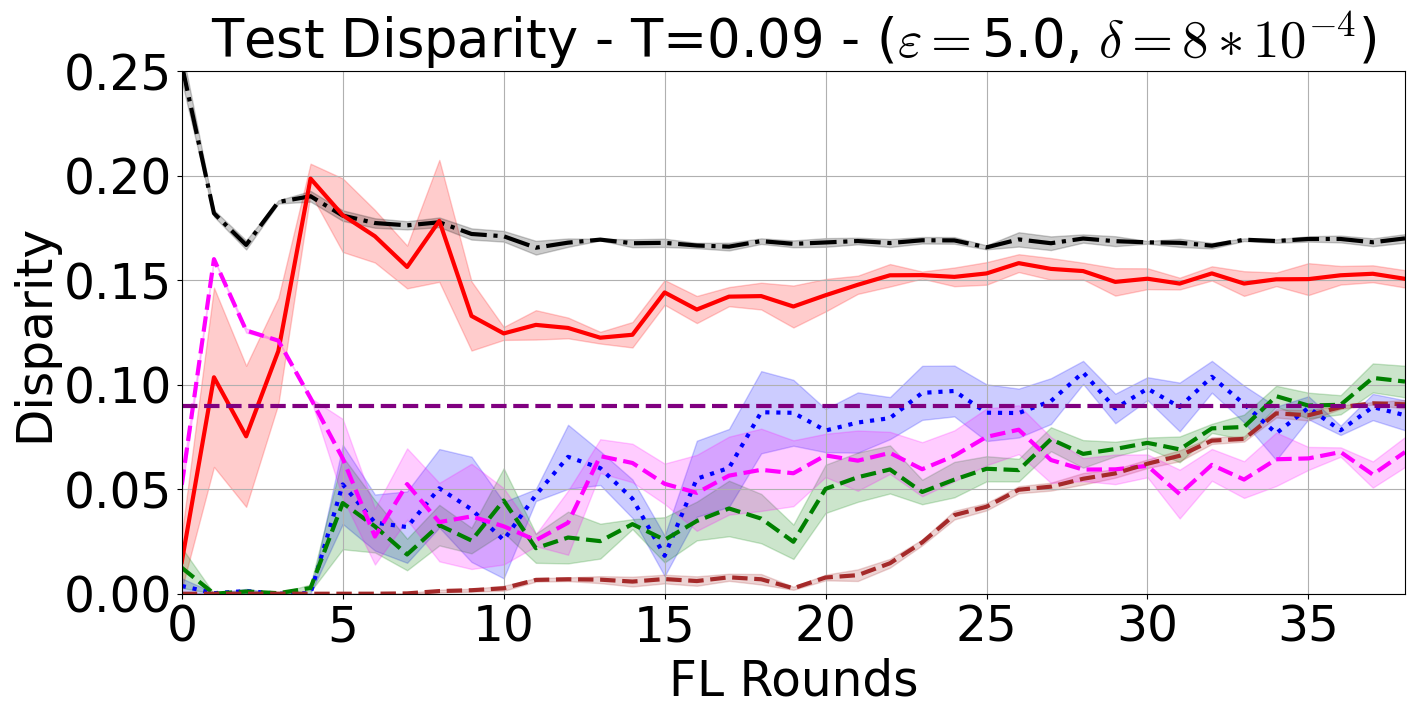}
         \caption{}
         \label{fig:disparity_2}
     \end{subfigure}\hfill
     \begin{subfigure}[b]{0.33\textwidth}
         \centering
         \includegraphics[width=\textwidth]{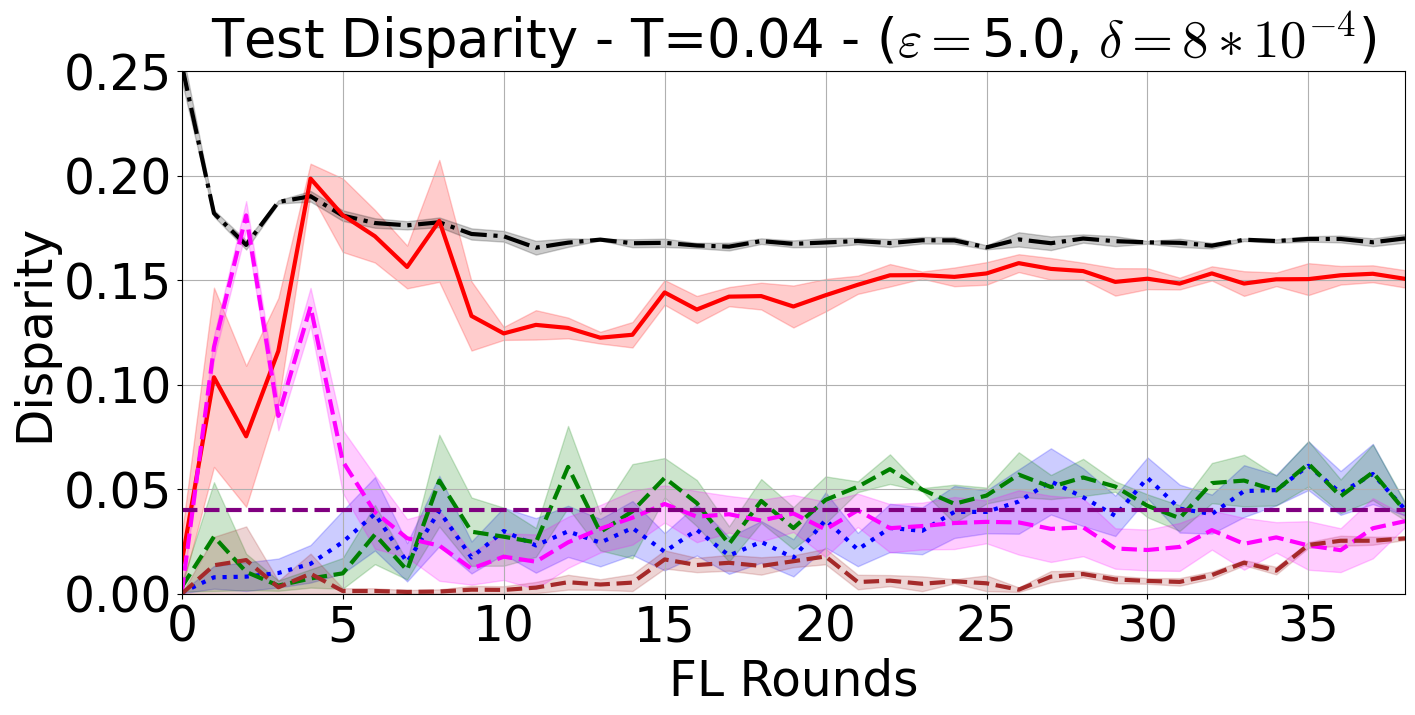}
         \caption{}
         \label{fig:disparity_3}
     \end{subfigure}\\
     \vspace{0.30cm}
     \includegraphics[width=0.75\linewidth]{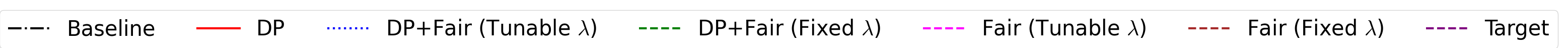}\\
     \vspace{0.15cm}
        \caption{
        Fig.~\ref{fig:accuracy_1}, ~\ref{fig:accuracy_2}, ~\ref{fig:accuracy_3} show Test Accuracy. Fig.~\ref{fig:disparity_1},~\ref{fig:disparity_2}, \ref{fig:disparity_3} show test disparity. We use one privacy parameter ($\varepsilon=5$,$\delta=8$$\times$$10^{-4}$) and different levels $T$ ($0.12$, $0.09$, $0.04$) of target disparity. These target parameters correspond to reductions of 25\%, 50\%, and 75\% compared to the Baseline's disparity. As we decrease the target disparity, we degrade the utility of the trained model using both Tunable and Fixed $\lambda$.}
        \label{fig:celeba_1}
\end{figure*}
\begin{figure*}
     \centering
     \captionsetup{justification=justified}
     \begin{subfigure}[b]{0.30\textwidth}
         \centering
         \includegraphics[width=\textwidth]{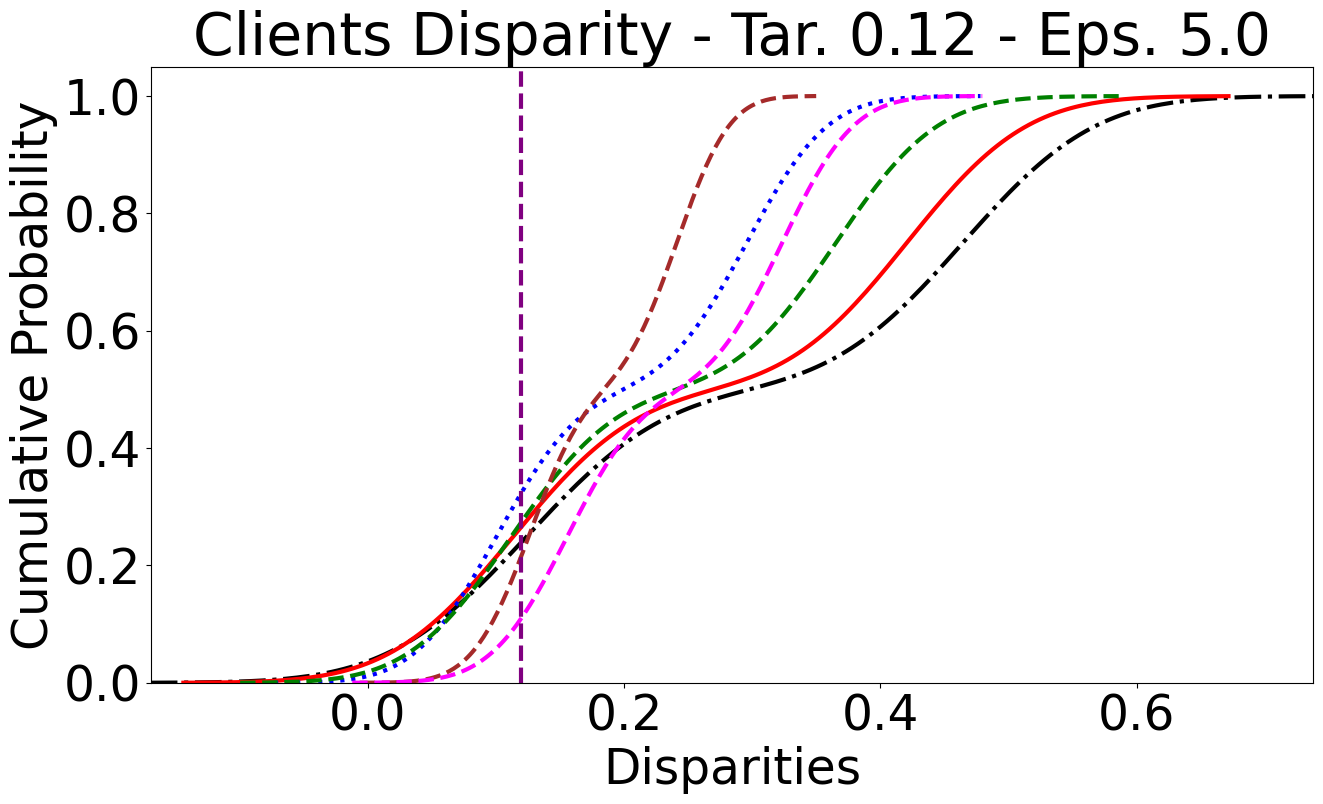}
         \caption{}
         \label{fig:cdf_1}
     \end{subfigure}
     \hfill
     \begin{subfigure}[b]{0.30\textwidth}
         \centering
         \includegraphics[width=\textwidth]{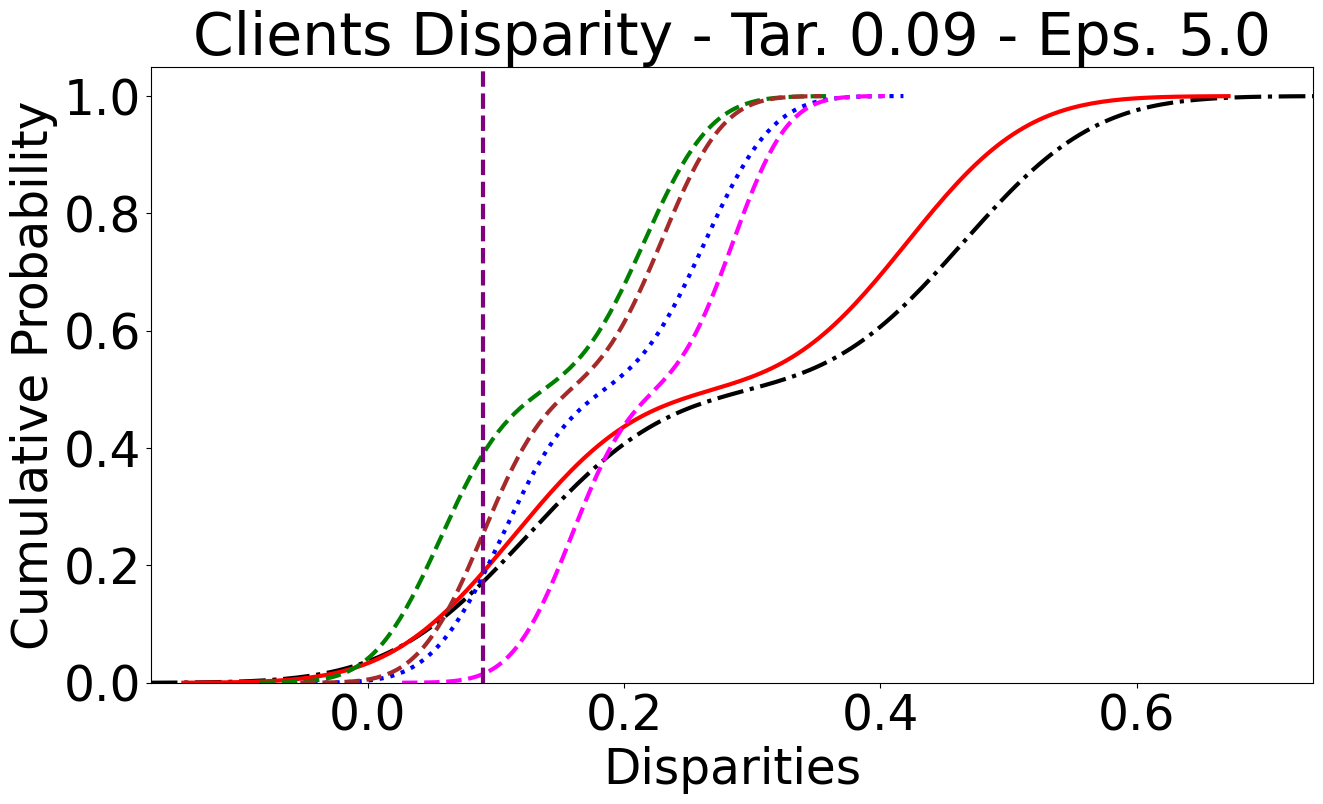}
         \caption{}
         \label{fig:cdf_2}
     \end{subfigure}\hfill
     \begin{subfigure}[b]{0.30\textwidth}
         \centering
         \includegraphics[width=\textwidth]{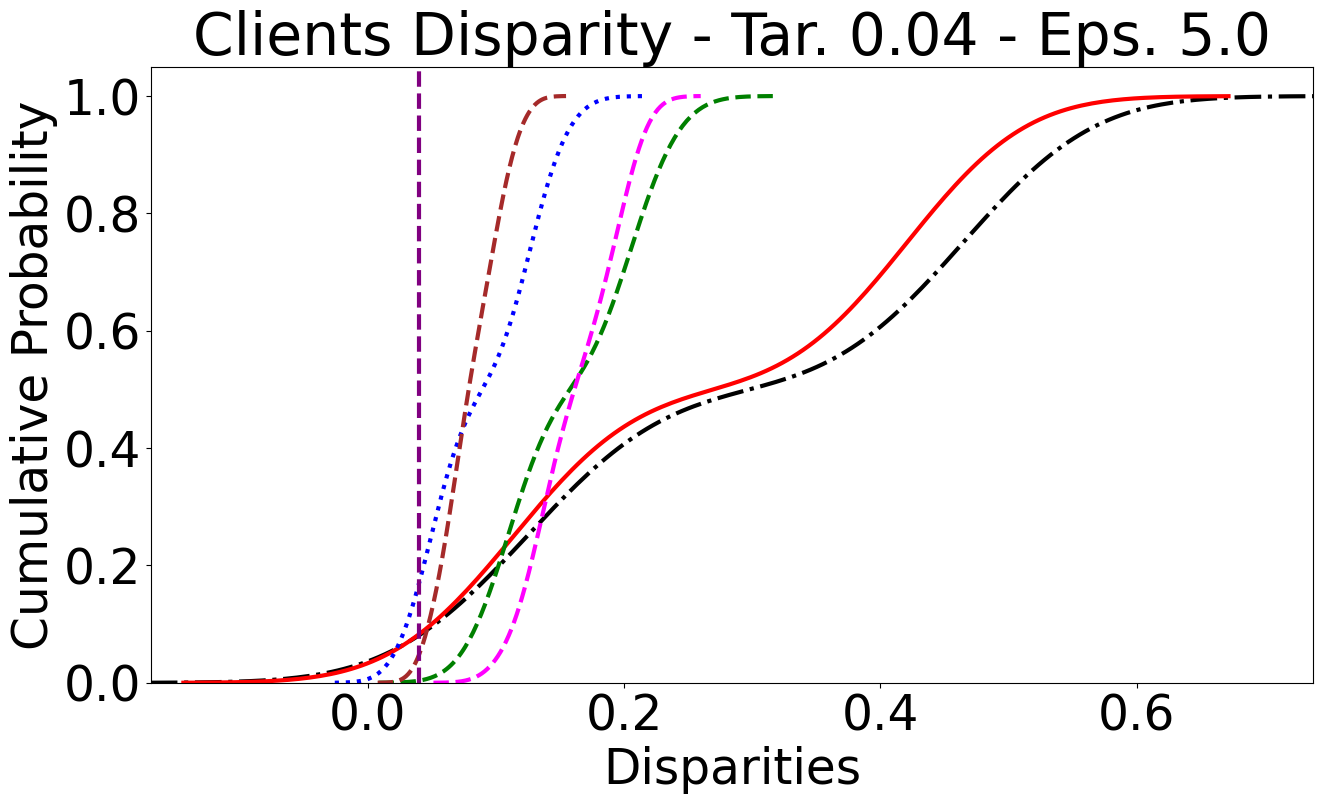}
         \caption{}
         \label{fig:cdf_3}
     \end{subfigure}\\
     \vspace{0.30cm}
     \includegraphics[width=0.75\linewidth]{images/legend_disparity_1.png}\\
     \vspace{0.15cm}
    \caption{CDF of the local clients' disparity with targets $T$ ($0.12$, $0.09$ $0.04$) and privacy parameter ($\varepsilon=5, \delta=8 \times  
 10^{-4}$)}
        \label{fig:celeba_local_1}
\end{figure*}

First, we find that the use of PUFFLE with the regularization with Tunable $\lambda$, leads to comparable results, in terms of accuracy and disparity, to the experiment with fixed $\lambda$. However, for the same results, our approach with tunable $\lambda$ turns out to be easier to interpret and use. Indeed, it enables selecting the target disparity $T$ and understanding the corresponding model accuracy facilitating the identification of optimal trade-offs among trustworthiness requirements.

Second, looking at the three accuracy plots, we notice that as we decrease the target fairness disparity $T$, there is a corresponding reduction in the accuracy of the trained model. There is an unavoidable trade-off between accuracy, fairness and privacy: good privacy and reduced disparity cost accuracy. The reduction of $T$ leads to an increase of the $\lambda$ used in the computation of the gradient $g = (1-\lambda)\times g_{loss} + (\lambda) \times g_{dpl}$. Increasing $\lambda$ implies assigning more weight to the fairness regularization term than to the model loss. Consequently, the model utility experiences a degradation. In Table \ref{tab:accuracy_degradation} we highlight the impact that the choice of target $T$ has on the accuracy degradation. In experiments with Fixed $\lambda$, accuracy drops by 6\% to 11\% compared to the Baseline, with a fairness reduction of at least 25\% to 75\%. It's even less than the Baseline with DP. With Tunable $\lambda$, the drop is similar, ranging from 8\% to 17\% compared to the Baseline and 6\% to 15\% compared to the Baseline with DP.

\begin{table}[t]
    \centering
    \resizebox{\columnwidth}{!}{
    \begin{tabular}{|c|c|c|c|}
        \hline
        \textbf{$\lambda$} & \textbf{T}(reduction from baseline) &  \textbf{Utility Drop} & \textbf{Utility Drop (from DP)} \\
        \hline
        Fixed & 0.12 (25\%)  &  6\% & 5\% \\\hline
        Tunable & 0.12 (25\%)  &  8\% & 6\% \\\hline
        Fixed & 0.09 (50\%)  &  13\% & 11\% \\\hline
        Tunable & 0.09 (50\%)  &  9\% & 7\% \\\hline
        Fixed & 0.04 (75\%)  &  11\% & 9\% \\\hline
        Tunable & 0.04 (75\%)  &  17\% & 15\% \\\hline
    \end{tabular}
    }
    \caption{A summary of how PUFFLE with privacy parameter ($\varepsilon=5, \delta=8 \times 10^{-4}$) affects the final model's utility compared to the Baseline and Baseline with DP.}
    \label{tab:accuracy_degradation}
\end{table}

Third, PUFFLE aims to train a model ensuring privacy while keeping unfairness below the target $T$. Moving away from the target, further reducing the unfairness, results in a fairer model, albeit with greater accuracy degradation, as shown in Figures \ref{fig:accuracy_1} and \ref{fig:accuracy_3}.

Table~\ref{tab:results_celeba} in Appendix~\ref{sec:appendix:results} lists model accuracy and disparity for all privacy and fairness target combinations across all datasets.

\subsection{Fairness for Local Clients}

In Figure~\ref{fig:disparity_1},~\ref{fig:disparity_2} and~\ref{fig:disparity_3}, we present a \textit{global view} of the test disparity computed by the server using differentially private statistics shared by the clients. Our approach also allows us to examine the clients' local disparity.
A \textit{local view} of clients' disparity aids in understanding how PUFFLE systematically reduces individual client unfairness in FL training.
Figure~\ref{fig:celeba_local_1} shows the Cumulative Distribution Function (CDF) of the final model disparity among test clients, under different targets $T$. Despite our methodology focusing on minimizing the \textit{global} disparity, the results demonstrate how the regularization positively influences local clients' disparity. The CDF describes the fraction of clients with a model disparity at most the value specified on the horizontal axis.  
We note that the curves for the Baseline and the DP models are lower than the ones for the models with Fairness+DP (PUFFLE) meaning that our approach is useful for reducing the unfairness of the local models across clients, while still providing privacy. Both the CDF of DP+Fair (Fixed $\lambda$) and the DP+Fair (Tunable $\lambda$) show similar reactions after using the Regularization. This mitigation can reduce the maximum disparity across clients and results in a more vertical distribution.

\subsection{Key Takeaways}
We summarize our experimental findings below:
\begin{itemize}
    \item We improved the concept of using regularization to alleviate the unfairness of ML models by introducing the notion of a Tunable $\lambda$. PUFFLE is more interpretable and parameterized compared to past work~\cite{yaghini2023learning}.
    \item Experiments on CelebA (Figure~\ref{fig:celeba_1}) demonstrate that PUFFLE reduces unfairness by at least 75\% while lowering accuracy by 17\% in the worst-case compared with the Baseline. These results are consistent with the other datasets. Compared with the Baseline, PUFFLE reduces model disparity by at least 75\%, with an accuracy degradation of only 21\% in the worst-case scenario with Dutch, and 1.4\% with ACS Income. An extensive evaluation of these two datasets is reported in Appendix~\ref{sec:appendix:results}.
    \item Despite our approach primarily targeting global model fairness, our results demonstrate that regularization positively impacts clients' local model disparity.
\end{itemize}

\section{Conclusion}

The changes in AI legislations~\cite {madiega2021artificial,  uk_whitepaper_2023,  biden2023executive} are pushing the development and deployment of ML models that can respect trustworthiness factors such as privacy and fairness. This paper has explored the relatively unexplored interplay between these factors and the utility of models in the context of FL. PUFFLE, our high-level parameterised methodology, enables clients to make informed decisions by expressing fairness, privacy, and utility preferences. This is crucial for models that must comply with AI regulations. 

As one of the first explorations in studying the relationship between utility, fairness and privacy in the context of FL, our work opens up many possibilities for future research. A limitation that our methodology currently has is that it is not able to handle the unfairness of clients who have a bias toward different sensitive groups, e.g., half of the clients unfair toward the group of ``Female'' while the other half toward the group of ``Male''. Thus, we think an extension of the methodology to handle scenarios where clients have biases toward different groups is needed. In addition, our future efforts aim to explore the use of alternative fairness metrics for regularization. We used the Demographic Disparity but there exists a lot of fairness metrics that could be used instead. This would make our system even more customized and suitable for use in many contexts. Another interesting problem that we did not consider in PUFFLE is how to balance these three trustworthiness requirements when the clients involved in the training have different requirements in terms of fairness or privacy.
Our contribution represents a significant step toward developing FL models capable of ensuring high accuracy, the privacy of the clients, and fairness to underrepresented groups.

\section{Acknowledgements}

This research was partially supported by:
The European Commission under the NextGeneration EU programme – National Recovery and Resilience Plan (PNRR), under agreements: PNRR - M4C2 - Investimento 1.3, Partenariato Esteso PE00000013 - "FAIR - Future Artificial Intelligence Research" - Spoke 1 "Human-centered AI", and SoBigData.it – "Strengthening the Italian RI for Social Mining and Big Data Analytics” – Prot. IR0000013 – Avviso n. 3264 del 28/12/2021
The Ministry of Economic Affairs and Digital Transformation of Spain and the European Union-NextGenerationEU programme for the "Recovery, Transformation and Resilience Plan" and the "Recovery and Resilience Mechanism" under agreements TSI-063000-2021-142 and TSI-063000-2021-147 (6G-RIEMANN).
The European Union Horizon 2020 program under grant agreements No. 101021808 (SPATIAL) and No. 101120763 (TANGO).
The views and opinions expressed are those of the authors only and do not necessarily reflect those of the European Union or the European Health and Digital Executive Agency (HaDEA). Neither the European Union nor the granting authority can be held responsible for them.

\clearpage

\bibliography{mybibfile}

\clearpage

\appendix

\section{DP-SGD}
\label{sec:appendix:dp-sgd}

We report in Algorithm ~\ref{alg:dp-sgd} the pseudocode of DP-SGD ~\cite{Abadi_2016} algorithm used in our methodology to train ML models with DP protection.

\begin{algorithm}[t]
\caption{Differentially private SGD,~\cite{Song_Chaudhuri_Sarwate_2013} }\label{alg:dp-sgd}
\begin{algorithmic}[1]

\Require Examples $\{x_1$, \ldots, $x_N\}$, loss function $\mathcal{L}(\theta) = \frac{1}{N} \sum_{i} \mathcal{L}(\theta, x_i)$. Parameters: learning rate $\eta_t$, noise scale $\sigma$, group size L, gradient norm bound $B$, number of steps $E$.

\State  \textbf{Initialize} $\theta_0$ randomly.

\For{$e\in [E]$}

\State Take a random sample $L_e$ with sampling probability L/N

\State \textbf{Compute gradient}

\State For each $i \in L_e$, compute $g_t(x_i) \leftarrow$ 
$\nabla_{\theta_e} \mathcal{L}(\theta_e, x_i)$

\State \textbf{Clip gradient} 

\State $\bar{g_e}(x_i) \leftarrow g_e(x_i) / max(1, \frac{||g_e(x_i)||_2}{B})$

\State \textbf{Add noise}

\State $\tilde{g_e} \leftarrow \frac{1}{L}(\sum_i \bar{g_e} (x_i) + \mathcal{N}(0, \sigma^2B^2I))$

\State \textbf{Descent} 

\State $\theta_{e+1} \leftarrow \theta_e - \eta_e \tilde{g_e}$
\EndFor

\State \Return $\theta_E$.
\end{algorithmic}
\end{algorithm}

\section{Model Architecture}
\label{sec:appendix:model_architecture}

We report in Table \ref{tab:models} the architecture of the models used in our experiments. 

\begin{table*}[!t]
    \centering
    \captionsetup{justification=justified}
    {\footnotesize
        \begin{tabular}{|c|c|}
        \hline
        Layer & Description \\ \hline 
        Conv2D with Relu & (3, 8, 3, 1) \\
        \hline
        Max Pooling & (2, 2) \\
        \hline
        Conv2D with Relu & (8, 16, 3, 1)\\
        \hline
        Max Pooling & (2, 2) \\
        \hline
        Conv2D with Relu & (16, 32, 3, 1)\\
        \hline
        Max Pooling & (2, 2) \\
        \hline
        Fully Connected with Relu & $(8 \times  8 \times   32)$\\
        \hline
    \end{tabular}\hspace{1cm}
    \begin{tabular}{|c|c|}
        \hline
        Layer & Description \\ \hline 
        Fully Connected Layer & (input size, 2) \\
        \hline
    \end{tabular}
    }
    \caption{Architecture of the neural network used in our experiments.  On the left is the one used with CelebA and on the right is the one used with the Dutch dataset and ACS Income datasets.}
     \label{tab:models}
\end{table*}

\section{Data Distribution}
\label{sec:appendix:split}

For the experiments presented in the paper, we used three different datasets: CelebA~\cite{liu2015faceattributes}, Dutch~\cite{dutch} and ACS Income~\cite{ding2022retiring}. The latter contains data from all 50 states of the USA and Puerto Rico. Therefore, we had a natural split of it into 51 clients and we exploited it to train the federated model.
For the other two datasets, we had to use an algorithm to artificially split them into $K$ clients to simulate an FL scenario. The approach used to split the dataset is described in Algorithm ~\ref{alg:data_split}. In this case, we assume a scenario in which, for each sample in the dataset, we have both a binary sensitive value $Z$ and a binary label $Y$. 

\begin{figure}
\captionsetup{justification=justified}
\centering
\includegraphics[width=\linewidth]{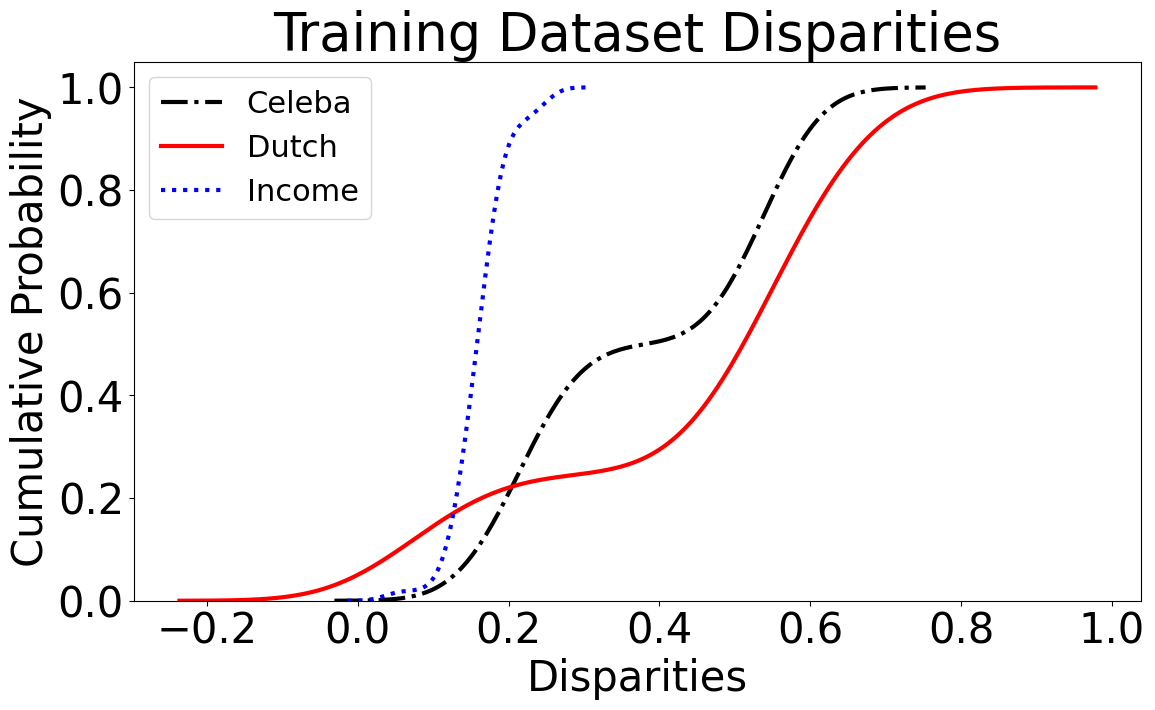}
\caption{A comparison between the Cumulative Distribution Function of the training dataset disparities for the different datasets used in the experiments}
\label{fig:cdf_training}
\vspace{0.5cm}
\end{figure}

First of all, the dataset $D$ is split into equal parts among the $K$ clients so that all the $D_k$ datasets have the same distribution as the original dataset $D$ (Line ~\ref{line:divide} of Algorithm ~\ref{alg:data_split}). Then, what we want to have is a group of fair clients $F$ and a group of unfair clients $U$. We fix this ratio with a parameter $u$ that represents the percentage of unfair clients. To make the group $U$ unfair, we fix a combination of $\langle Z=z$ $Y=y \rangle$ for which we want to decrease the number of samples in the dataset by a certain percentage $\zeta$. Consequently, we will increase the number of samples for the combination $\langle Z \neq z$, $Y=y \rangle$. 

Given the group $U$ of unfair clients, we remove from each of the datasets $D_k$ a percentage $\zeta$ of samples with $\langle Z = z$, $Y=y \rangle$ (Line ~\ref{line:remove} of the Algorithm ~\ref{alg:data_split}). Then, we divide the amount of total removed samples by the number $F$ of fair clients $\beta = \frac{\sum_{k=1}^K removed_{k}}{F}$ (Line ~\ref{line:beta} of the Algorithm ~\ref{alg:data_split}). For each client $k$ in the fair group $F$, we remove $\beta$ samples from the group $\langle Z \neq z$, $Y=y \rangle$ and we add the same amount of samples taken from $removed_{k}$ (Line ~\ref{line:fair} of the Algorithm ~\ref{alg:data_split}). The samples that we remove from the group of fair clients will be added in the same way to the group of unfair clients (Line ~\ref{line:unfair} of the Algorithm ~\ref{alg:data_split}). In the end, we will have a dataset split into $K$ equal parts of which $F$ clients will have a low demographic disparity while $U$ will have a higher disparity.

\begin{algorithm}[t]
\caption{Dataset Distribution}\label{alg:data_split}
\begin{algorithmic}[1]
\Require $D$ Dataset to be split among the clients. $K$ Number of clients. $U$ group of unfair clients. $u$ percentage of unfair clients. $F$ = $K$ - $U$ group of fair clients. $\langle Z=z, Y=y \rangle $ group of samples with sensitive value $Z=z$ and label $Y=y$ that we want to reduce in the unfair clients. $\langle Z\neq z, Y=y \rangle $ group of samples with sensitive value $Z \neq z$ and label $Y=y$ that we want to increase in the unfair clients. 
\State Split $D$ into $K$ parts keeping the original dataset distribution in each $D_{k}$ \label{line:divide}
\State counter = 0, removed = []
\For{k in $U$}
\State Remove from $D_K$ the samples with $Z=z$ and $Y=y$ and add them to $removed$ \label{line:remove}
\State $counter$ += len(removed samples)
\EndFor
\State $\beta = \frac{\sum_{k=1}^K removed_{k}}{F}$ \label{line:beta}
\State add = []
\For{k in $F$}
\State Add to $D_K$ $\beta$ samples taken from $removed$ \label{line:fair}
\State Remove $\beta$ samples with $Z \neq z$ and $Y = y$ and append to $add$
\EndFor
\For{k in $U$}
\State Add $\beta$ samples taken from $add$ to $D_K$ \label{line:unfair} 
\EndFor
\State 
\end{algorithmic}
\end{algorithm}

Besides considering this basic data distribution, we also wanted to consider a more real-life scenario in which some of the clients lack one of the possible sensitive values $Z$ or one of the combinations $\langle Z=z, Y=y\rangle$. The latter case is simple, we just needed to completely remove the samples with $\langle Z=z, Y=y\rangle$ from the group $U$ of unfair clients. We used this approach with the experiments with the Dutch Dataset.In particular, $50\%$ of the clients were unfair and lack samples with $\langle Z=1 \rangle$, i.e., female samples.

In the other case, when we only want to remove samples with $\langle Z=z, Y=y\rangle$ from some of the clients the algorithm described above is slightly changed. In particular, at the end of the algorithm, what we do is remove all the samples with $\langle Z=z, Y=y\rangle$ from the unfair group $U$ leaving these clients without these specific combinations. We used this approach with CelebA, in this case, $50\%$ of the clients were considered unfair and lacked samples with $\langle Z=1, Y=0\rangle$, i.e., samples with a not smiling man. In these cases, the size of the datasets $d_k$ was different between the fair group $F$ and the unfair group $U$.

Figure ~\ref{fig:cdf_training} shows the Cumulative Distribution Function of the disparities of the local training datasets $d_{k}$ for the three datasets used for our experiments.

\section{Results with other datasets}
\label{sec:appendix:results}

In Section ~\ref{sec:results}, we only presented the results we got using a specific combination of $\langle sensitive\ value,\ target \rangle$ using the CelebA dataset. In this section, we will present all the other results that we obtained using the other two datasets (Dutch and ACS Income) and the other combinations of $\langle sensitive\ value,\ target \rangle$ for CelebA.

\subsection{CelebA}

\begin{figure*}[t]
    \centering
    \captionsetup{justification=justified}
    \begin{subfigure}[b]{0.45\textwidth}
         \centering
         \includegraphics[width=\textwidth]{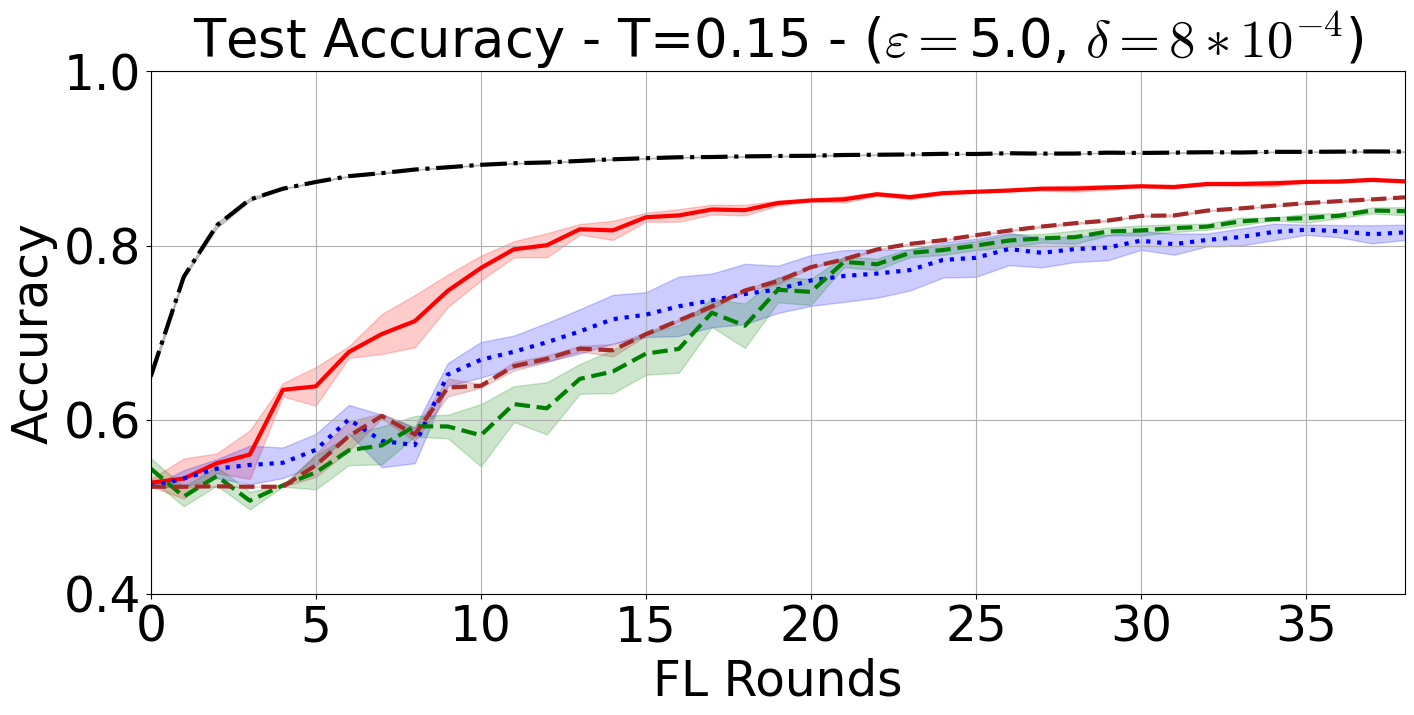}
         \caption{}
         \label{fig:accuracy_4}
     \end{subfigure}
    \hfill
     \begin{subfigure}[b]{0.45\textwidth}
         \centering
         \includegraphics[width=\textwidth]{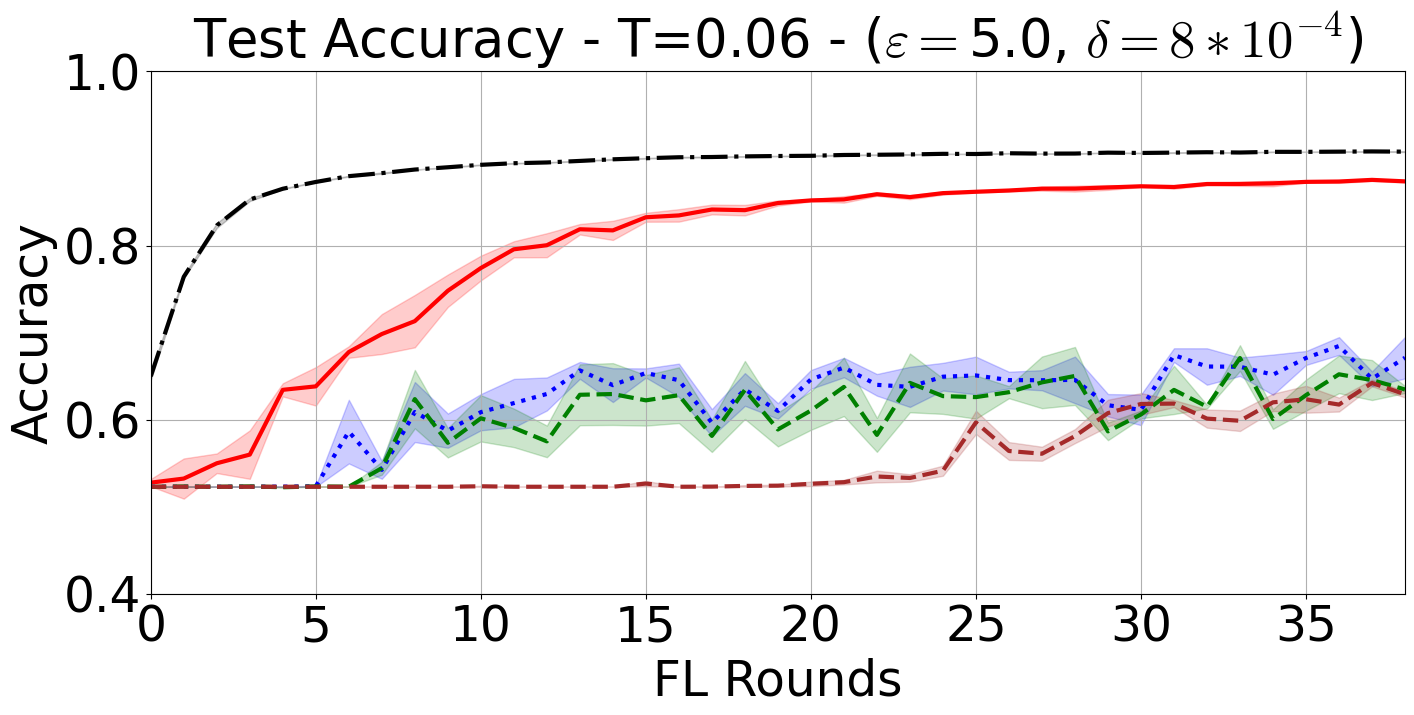}
         \caption{}
         \label{fig:accuracy_5}
     \end{subfigure}\\
     \vspace{0.5cm}
     \includegraphics[width=0.50\linewidth]{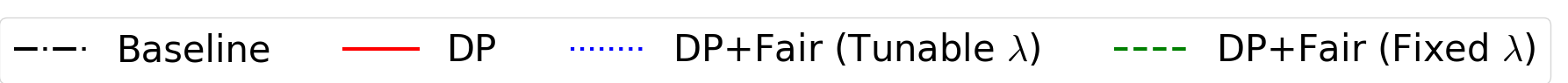}\\

    \begin{subfigure}[b]{0.45\textwidth}
         \centering
         \includegraphics[width=\textwidth]{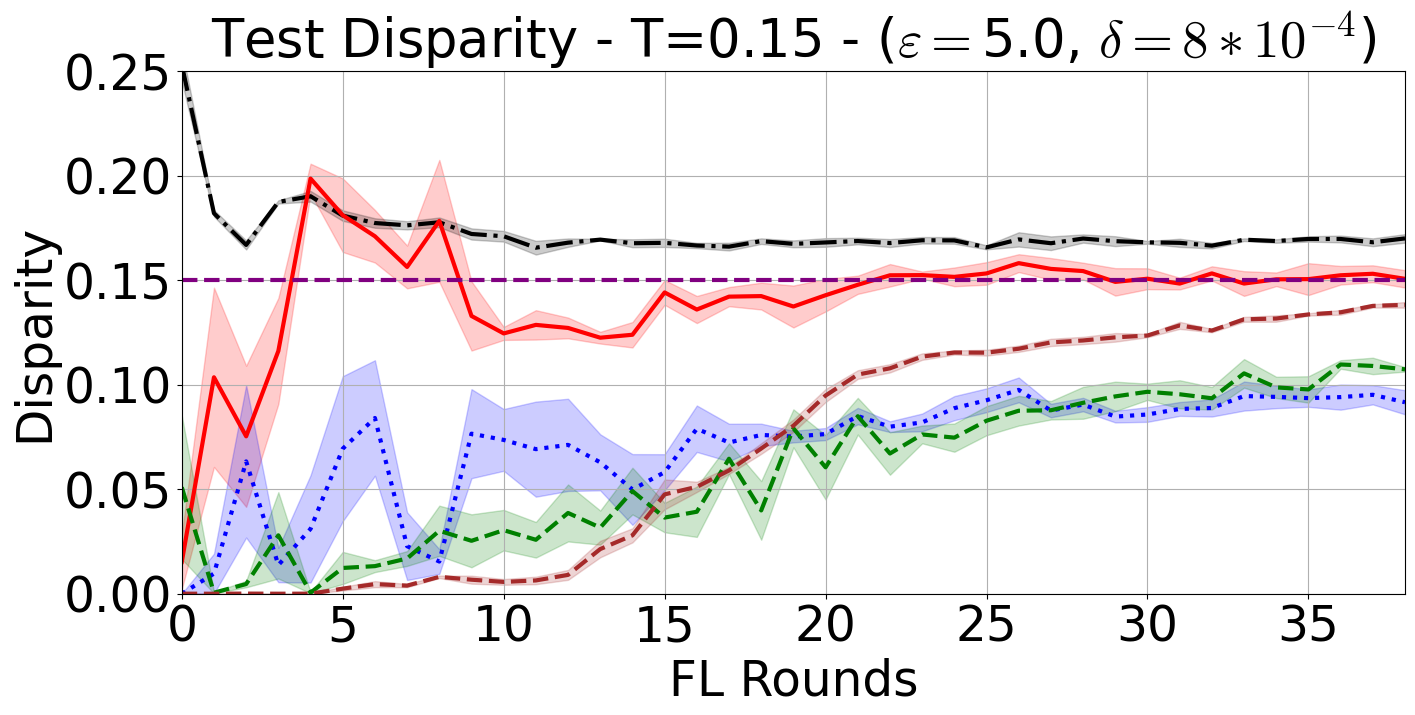}
         \caption{}
         \label{fig:disparity_4}
    \end{subfigure}
     \hfill
     \begin{subfigure}[b]{0.45\textwidth}
         \centering
         \includegraphics[width=\textwidth]{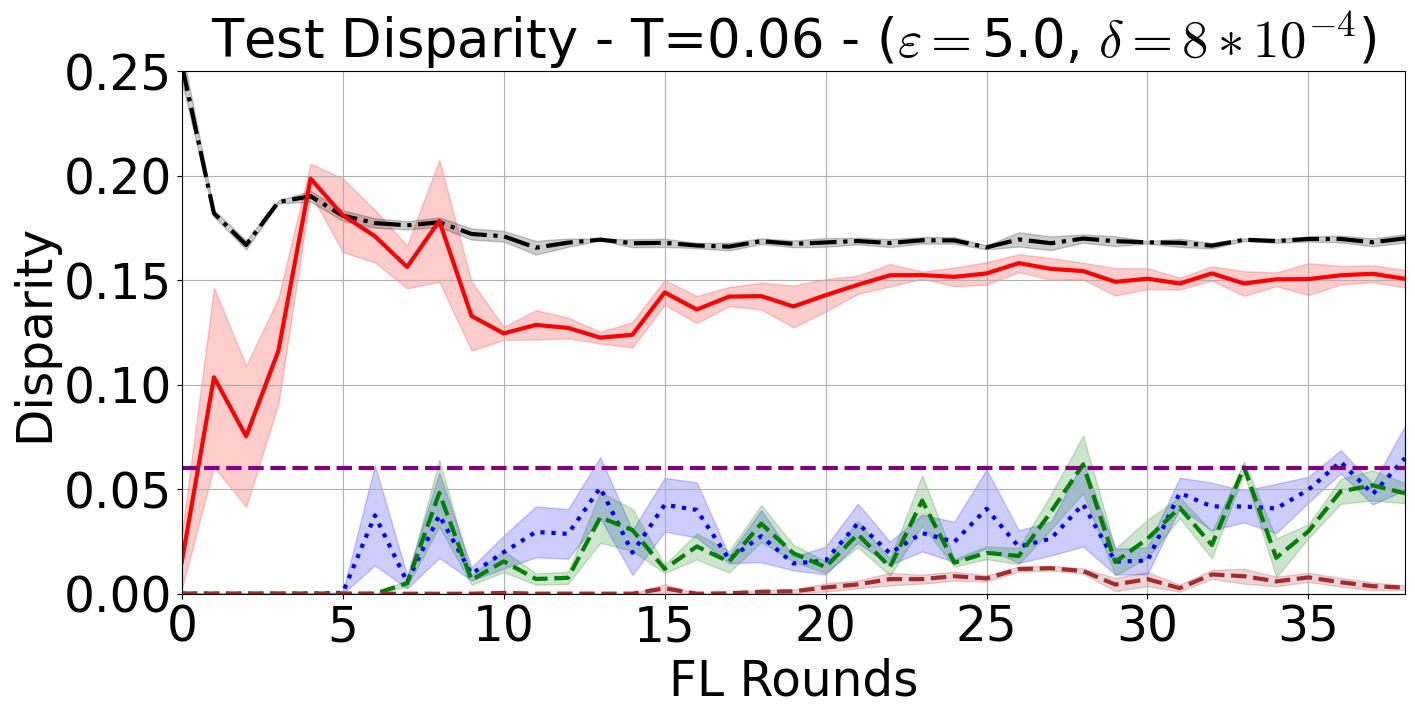}
         \caption{}
         \label{fig:disparity_5}
     \end{subfigure}\\
     \vspace{0.5cm}
     \includegraphics[width=0.60\linewidth]{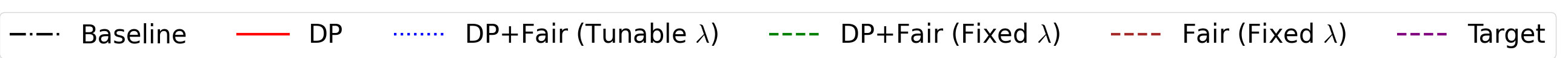}\\

        \caption{Figures~\ref{fig:accuracy_4} and~\ref{fig:accuracy_5} show the Test Accuracy. Figures~\ref{fig:disparity_4} and~\ref{fig:disparity_5} show the test disparity of the experiments conducted with a privacy parameter ($\varepsilon=5, \delta=8 \times  10^{-4}$) and different levels of target demographic disparity $T$ ($0.15$ and $0.06$). These target parameters correspond to a reduction of 10\% and 75\% with respect to the disparity of the Baseline. The more we decrease the target disparity $T$, the more we degrade the utility of the trained model both using the Tunable and the Fixed $\lambda$.}
        \label{fig:celeba_5_others}
\end{figure*}

\begin{figure*}
     \centering
     \captionsetup{justification=justified}
     \begin{subfigure}[b]{0.45\textwidth}
         \centering
         \includegraphics[width=\textwidth]{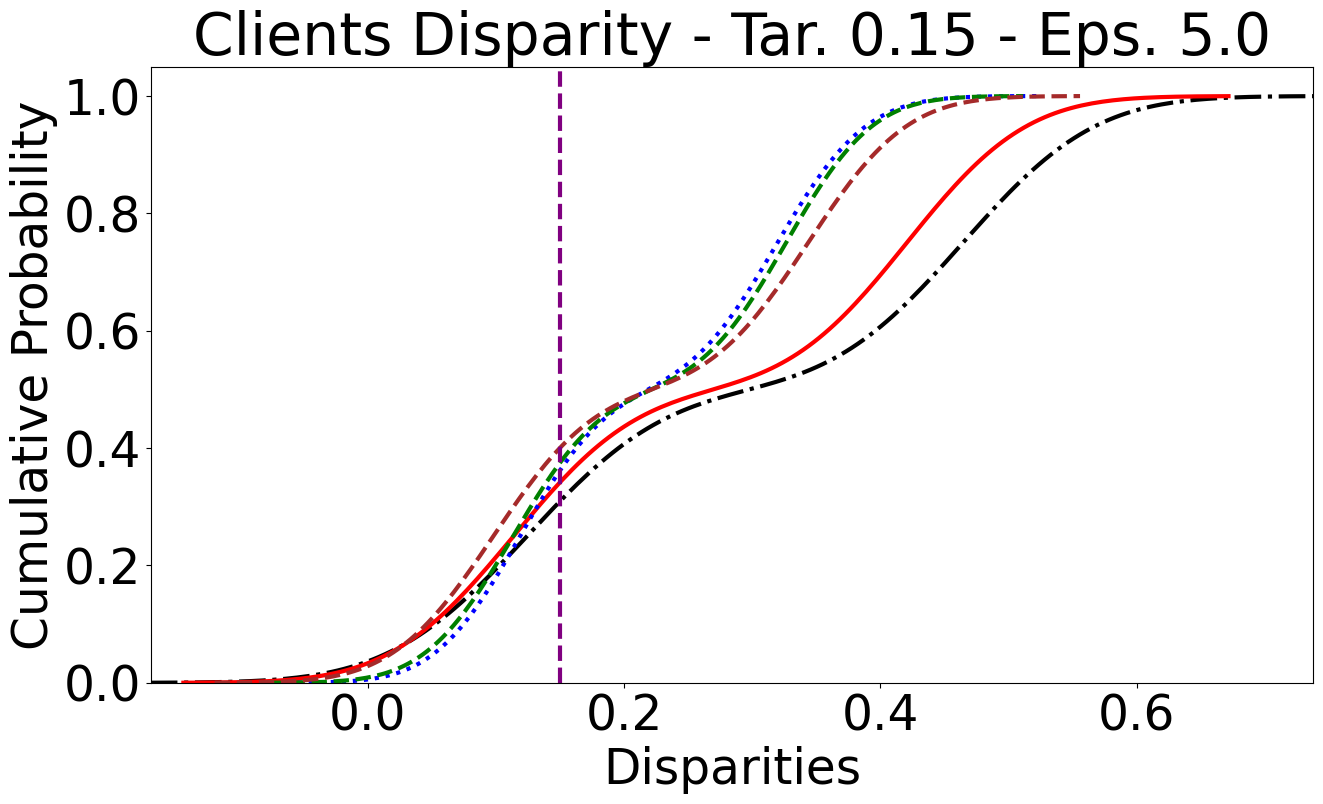}
         \caption{}
         \label{fig:cdf_4}
     \end{subfigure}
     \hfill
     \begin{subfigure}[b]{0.45\textwidth}
         \centering
         \includegraphics[width=\textwidth]{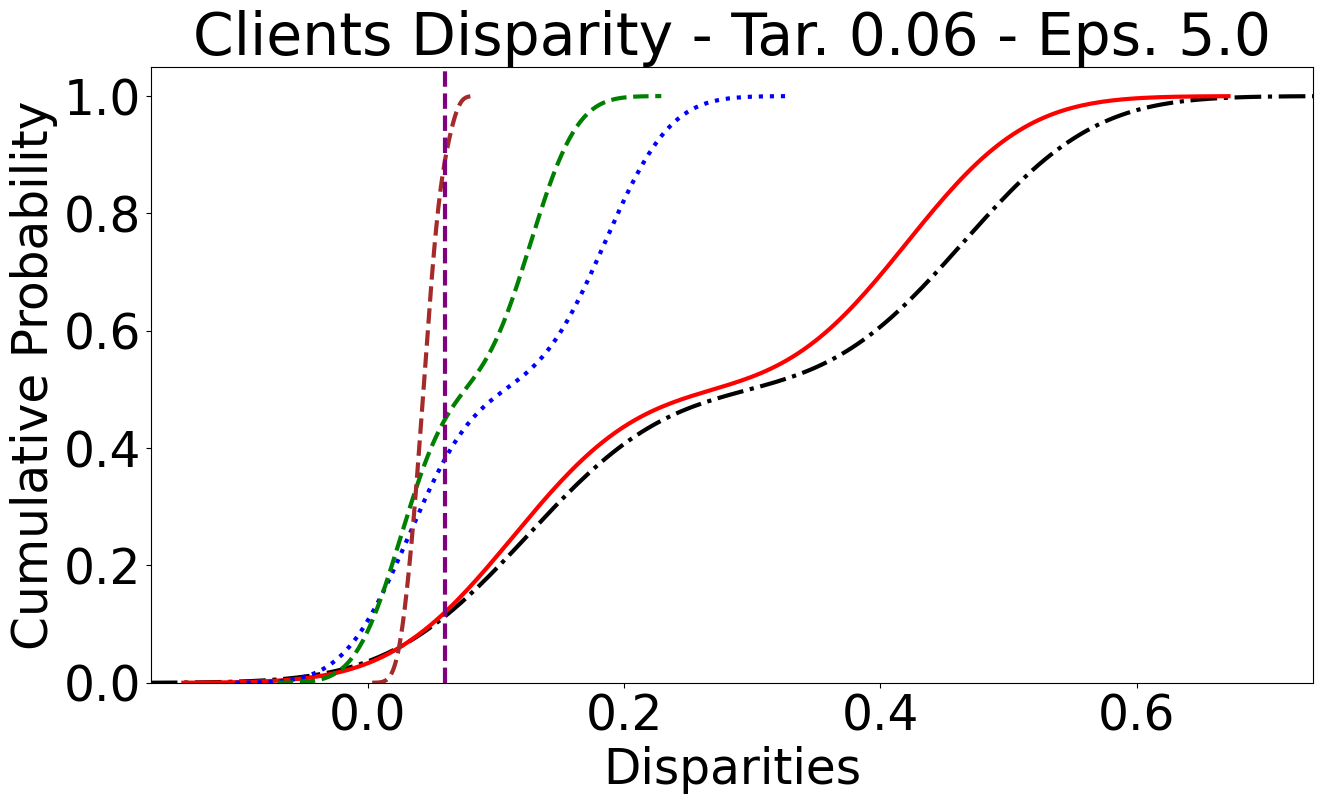}
         \caption{}
         \label{fig:cdf_5}
     \end{subfigure}\\
     \vspace{0.5cm}
     \includegraphics[width=0.60\linewidth]{images/legend_disparity.png}\\

        \caption{Cumulative Distribution Functions (CDF) of the local clients' disparity with three disparity targets $T$ ($0.15$ and $0.06$) and privacy parameter ($\varepsilon=5, \delta=8 \times  10^{-4}$)}
        \label{fig:celeba_local_others}
\end{figure*}

Figure ~\ref{fig:celeba_5_others} shows the results obtained with CelebA with the following combinations of privacy and disparity targets: privacy target $(\epsilon=5, \gamma=8 \times  10^{-4})$ and target disparity $T=0.06$ and $T=0.15$. These two targets correspond to a reduction of 65\% and 10\% with respect to the baseline disparity. We also report the CDF of the local clients' disparity in Figure \ref{fig:celeba_local_others}.

\begin{table*}[!t]
    \centering
    \captionsetup{justification=justified}
    \begin{tabular}{|c|c|c|c|c|}\hline
         & \textbf{($\varepsilon$, $\delta$)} & \textbf{$T$}& \textbf{Accuracy} & \textbf{Disparity}\\\hline
        Baseline & - & - & 0.908+-0.0& 0.17+-0.002\\\hline\hline
        
        DP & (5.0, $\mathrm{8*10^{-4}}$)) & - & 0.874+-0.001& 0.151+-0.004\\\hline
        DP & (8.0, $\mathrm{8*10^{-4}}$)) & - & 0.879+-0.002& 0.151+-0.004\\\hline\hline
        
        Fair & - & 0.04 & 0.714+-0.009& 0.026+-0.001\\\hline
        Fair & - & 0.06 & 0.629+-0.004& 0.003+-0.001\\\hline
        Fair & - & 0.09 & 0.803+-0.0& 0.091+-0.002\\\hline
        Fair & - & 0.12 & 0.83+-0.007& 0.075+-0.004\\\hline
        Fair & - & 0.15 & 0.855+-0.0& 0.138+-0.001\\\hline\hline
        
        PUFFLE (Fixed $\lambda$) & (5.0, $\mathrm{8*10^{-4}}$)) & 0.04 & 0.775+-0.015& 0.04+-0.004\\\hline
        PUFFLE (Tunable $\lambda$) & (5.0, $\mathrm{8*10^{-4}}$)) & 0.04 & 0.683+-0.01& 0.04+-0.003\\\hline
        PUFFLE (Fixed $\lambda$) & (8.0, $\mathrm{8*10^{-4}}$)) & 0.04 & 0.636+-0.022& 0.021+-0.003\\\hline
        PUFFLE (Tunable $\lambda$) & (8.0, $\mathrm{8*10^{-4}}$)) & 0.04 & 0.66+-0.006& 0.047+-0.007\\\hline\hline
        
        PUFFLE (Fixed $\lambda$) & (5.0, $\mathrm{8*10^{-4}}$)) & 0.06 & 0.635+-0.004& 0.048+-0.005\\\hline
        PUFFLE (Tunable $\lambda$) & (5.0, $\mathrm{8*10^{-4}}$)) & 0.06 & 0.671+-0.024& 0.065+-0.015\\\hline
        PUFFLE (Fixed $\lambda$) & (8.0, $\mathrm{8*10^{-4}}$)) & 0.06 & 0.634+-0.004& 0.008+-0.002\\\hline
        PUFFLE (Tunable $\lambda$) & (8.0, $\mathrm{8*10^{-4}}$)) & 0.06 & 0.686+-0.013& 0.056+-0.009\\\hline\hline

        PUFFLE (Fixed $\lambda$) & (5.0, $\mathrm{8*10^{-4}}$)) & 0.09 & 0.774+-0.011& 0.102+-0.008\\\hline
        PUFFLE (Tunable $\lambda$) & (5.0, $\mathrm{8*10^{-4}}$)) & 0.09 & 0.807+-0.005& 0.085+-0.007\\\hline
        PUFFLE (Fixed $\lambda$) & (8.0, $\mathrm{8*10^{-4}}$)) & 0.09 & 0.733+-0.002& 0.107+-0.003\\\hline
        PUFFLE (Tunable $\lambda$) & (8.0, $\mathrm{8*10^{-4}}$)) & 0.09 & 0.739+-0.055& 0.068+-0.018\\\hline\hline
        
        PUFFLE (Fixed $\lambda$) & (5.0, $\mathrm{8*10^{-4}}$)) & 0.12 & 0.83+-0.007& 0.118+-0.005\\\hline
        PUFFLE (Tunable $\lambda$) & (5.0, $\mathrm{8*10^{-4}}$)) & 0.12 & 0.815+-0.004& 0.103+-0.008\\\hline
        PUFFLE (Fixed $\lambda$) & (8.0, $\mathrm{8*10^{-4}}$)) & 0.12 & 0.686+-0.002& 0.089+-0.001\\\hline
        PUFFLE (Tunable $\lambda$) & (8.0, $\mathrm{8*10^{-4}}$)) & 0.12 & 0.792+-0.003& 0.073+-0.005\\\hline\hline
        
        PUFFLE (Fixed $\lambda$) & (5.0, $\mathrm{8*10^{-4}}$)) & 0.15 & 0.839+-0.005& 0.107+-0.001\\\hline
        PUFFLE (Tunable $\lambda$) & (5.0, $\mathrm{8*10^{-4}}$)) & 0.15 & 0.815+-0.009& 0.092+-0.006\\\hline
        PUFFLE (Fixed $\lambda$) & (8.0, $\mathrm{8*10^{-4}}$)) & 0.15 & 0.872+-0.002& 0.138+-0.007\\\hline
        PUFFLE (Tunable $\lambda$) & (8.0, $\mathrm{8*10^{-4}}$)) & 0.15 & 0.779+-0.061& 0.128+-0.014\\\hline

    \end{tabular}
    \caption{Celeba dataset: A comparison of the final test accuracy and the final test disparity across all the possible combinations. Specifically, the analysis encompasses two possible privacy parameters, namely ($\epsilon=5$, $8 \times  10^{-4}$) and ($\epsilon=8$, $8 \times  10^{-4}$) alongside three target fairness disparity, denoted as $T=0.04$, $T=0.06$, $T=0.09$, $T=0.12$, and $T=0.15$ corresponding to a reduction of 10\%, 25\%, 50\%, 65\%, 75\% with respect to the Baseline disparity.}
    \label{tab:results_celeba}
\end{table*}

\begin{figure*}
     \centering
     \captionsetup{justification=justified}
     \begin{subfigure}[b]{0.18\textwidth}
         \centering
         \includegraphics[width=\textwidth]{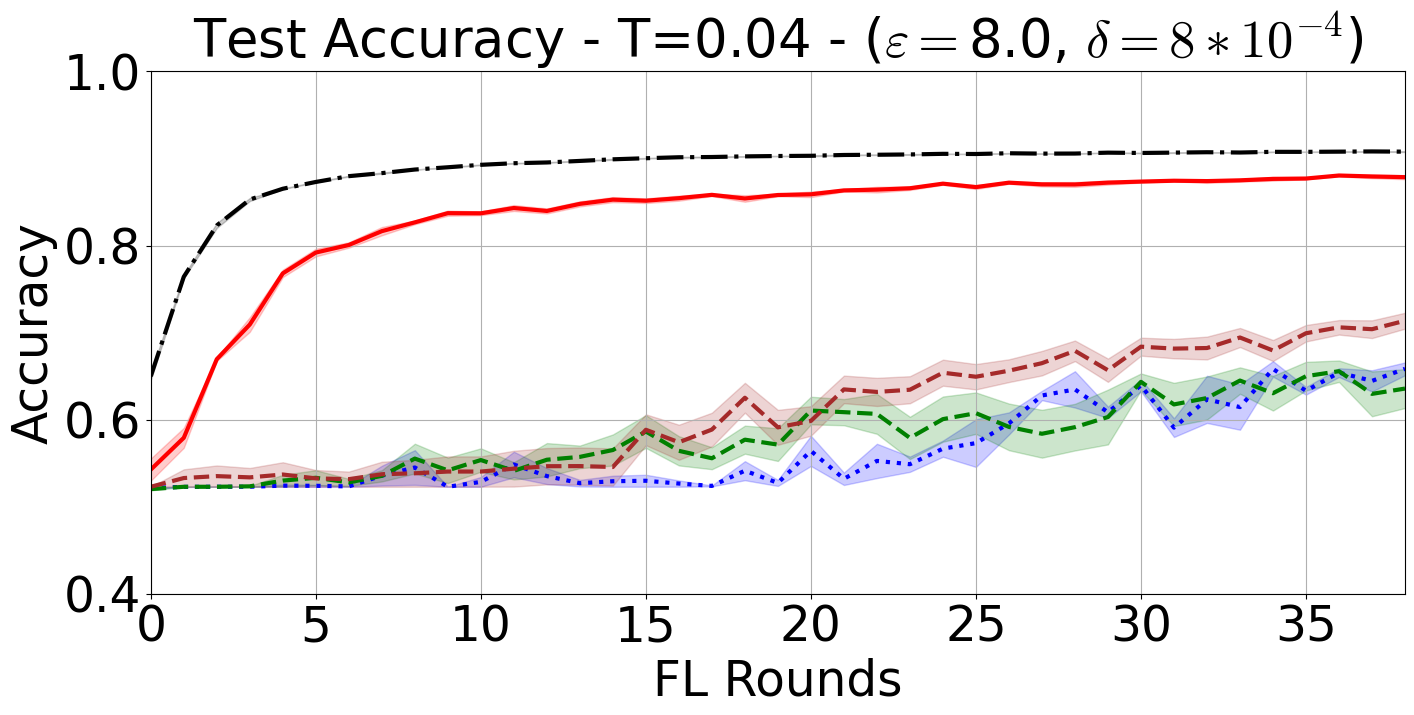}
         \caption{}
         \label{fig:celeba_8:accuracy_1}
     \end{subfigure}
     \hfill
     \begin{subfigure}[b]{0.18\textwidth}
         \centering
         \includegraphics[width=\textwidth]{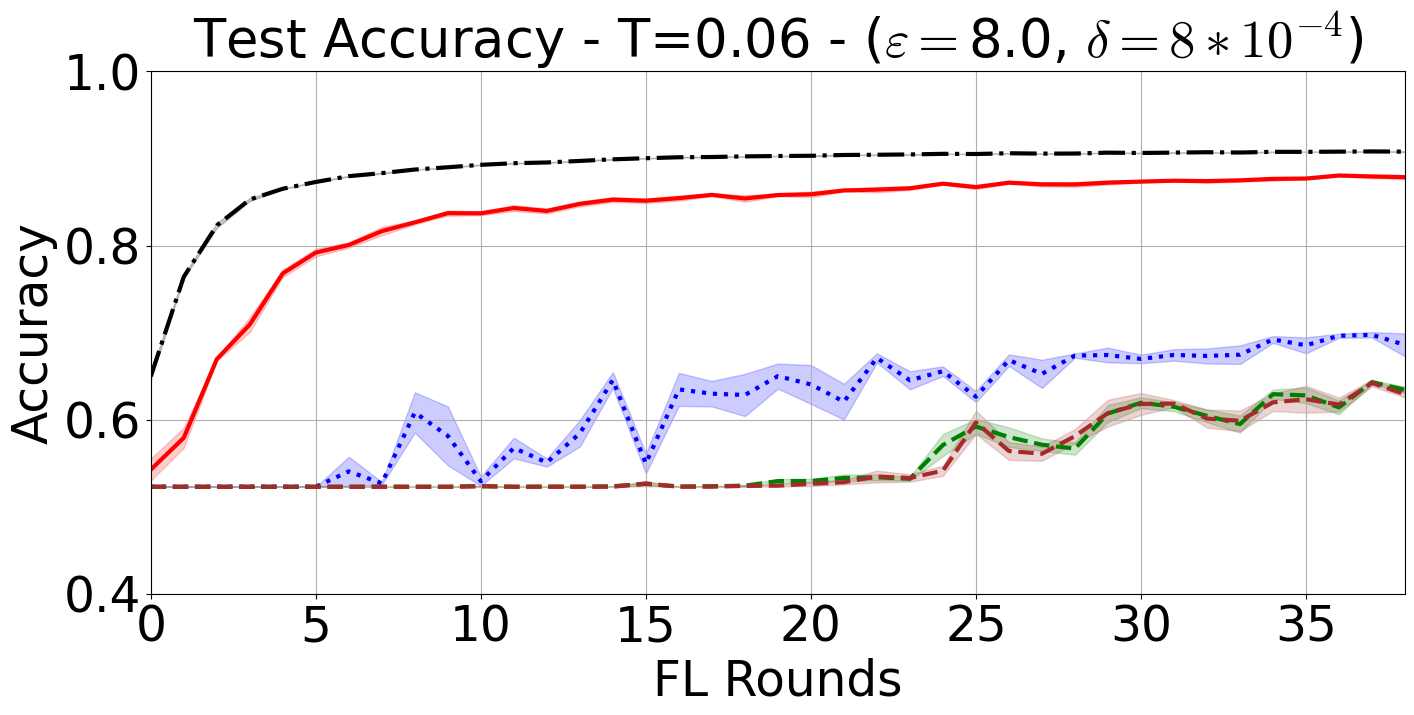}
         \caption{}
         \label{fig:celeba_8:accuracy_2}
     \end{subfigure}
     \hfill
     \begin{subfigure}[b]{0.18\textwidth}
         \centering
         \includegraphics[width=\textwidth]{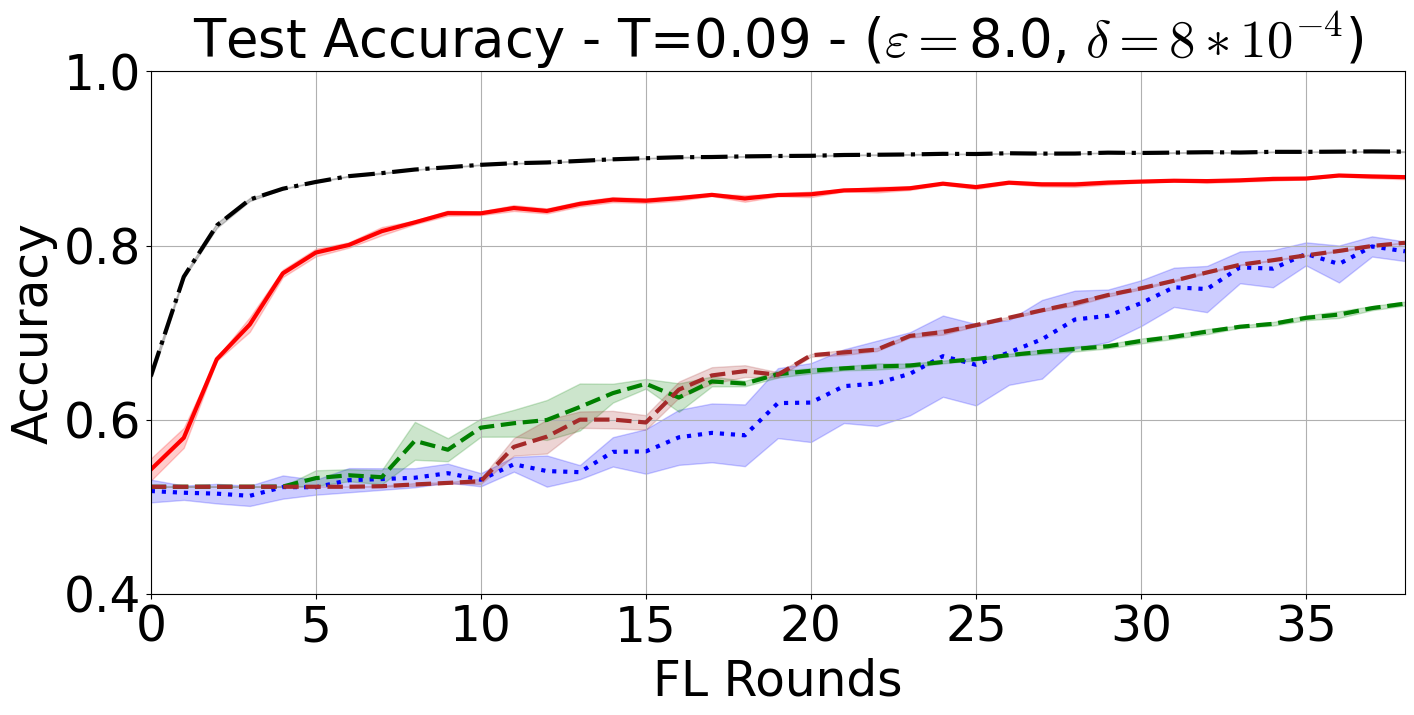}
         \caption{}
         \label{fig:celeba_8:accuracy_3}
     \end{subfigure}
     \hfill
     \begin{subfigure}[b]{0.18\textwidth}
         \centering
         \includegraphics[width=\textwidth]{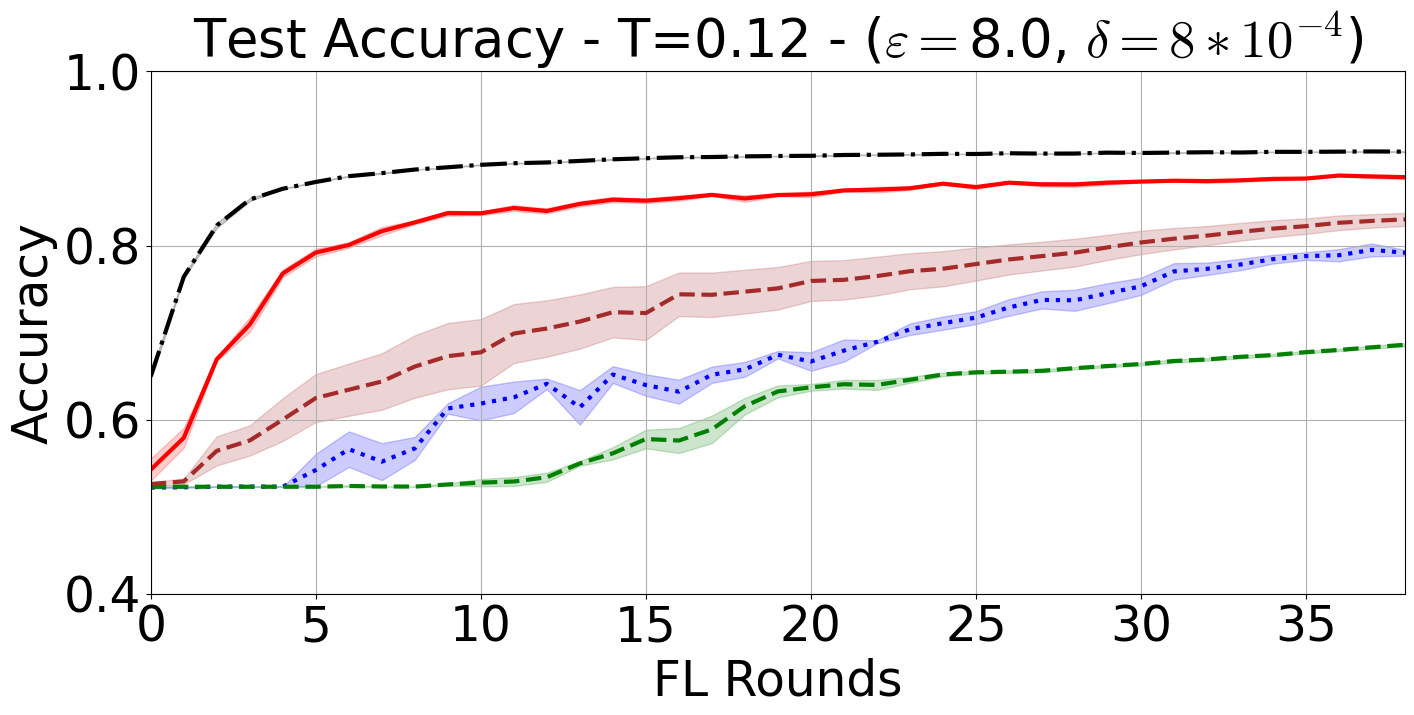}
         \caption{}
         \label{fig:celeba_8:accuracy_4}
     \end{subfigure}
     \hfill
     \begin{subfigure}[b]{0.18\textwidth}
         \centering
         \includegraphics[width=\textwidth]{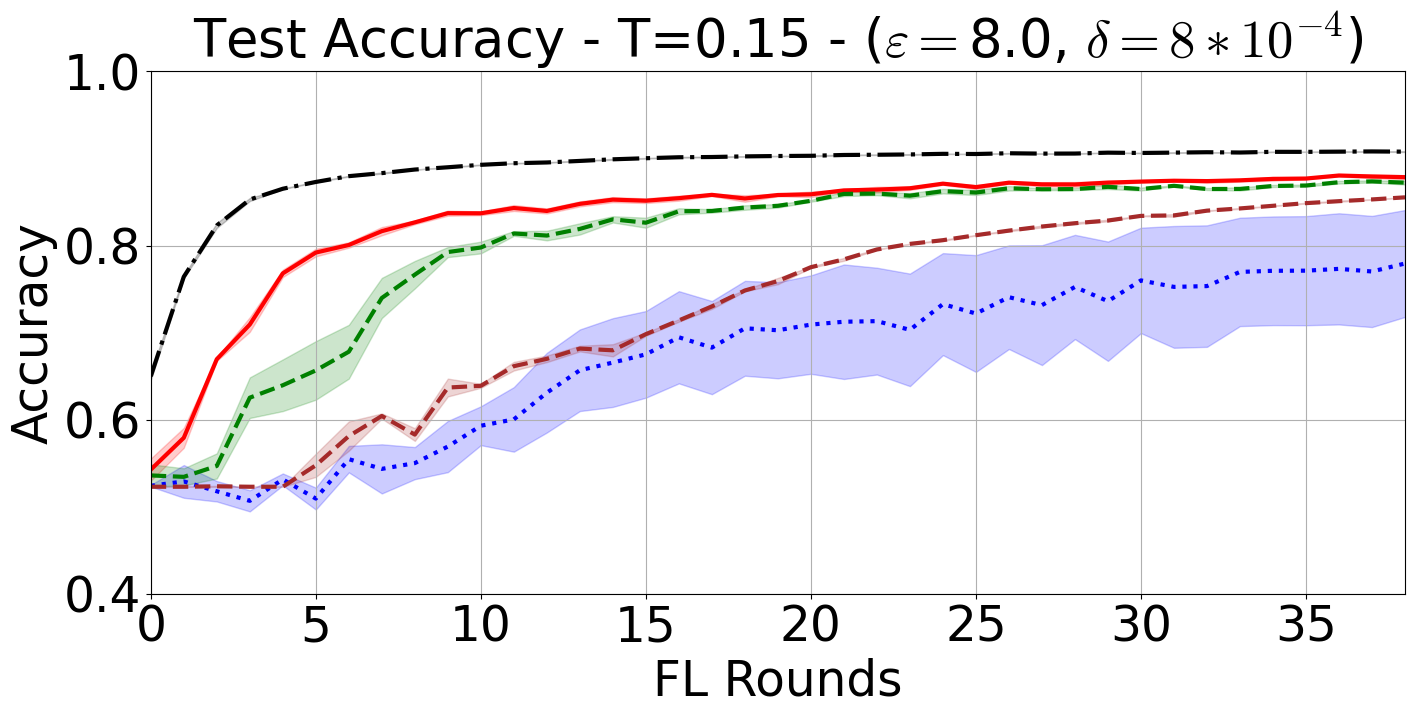}
         \caption{}
         \label{fig:celeba_8:accuracy_5}
     \end{subfigure}\\
    \vspace{0.30cm}
    \begin{subfigure}[b]{0.18\textwidth}
         \centering
         \includegraphics[width=\textwidth]{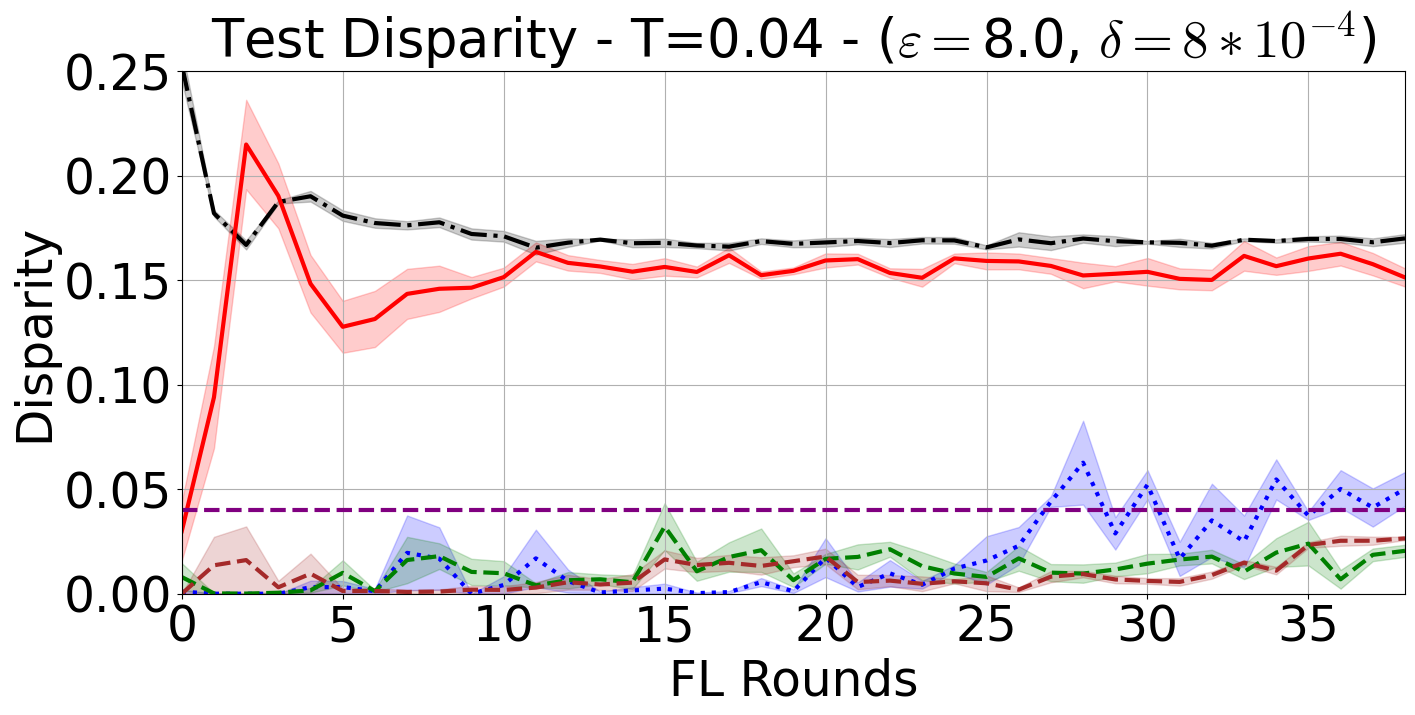}
         \caption{}
         \label{fig:celeba_8:disparity_1}
     \end{subfigure}
     \hfill
     \begin{subfigure}[b]{0.18\textwidth}
         \centering
         \includegraphics[width=\textwidth]{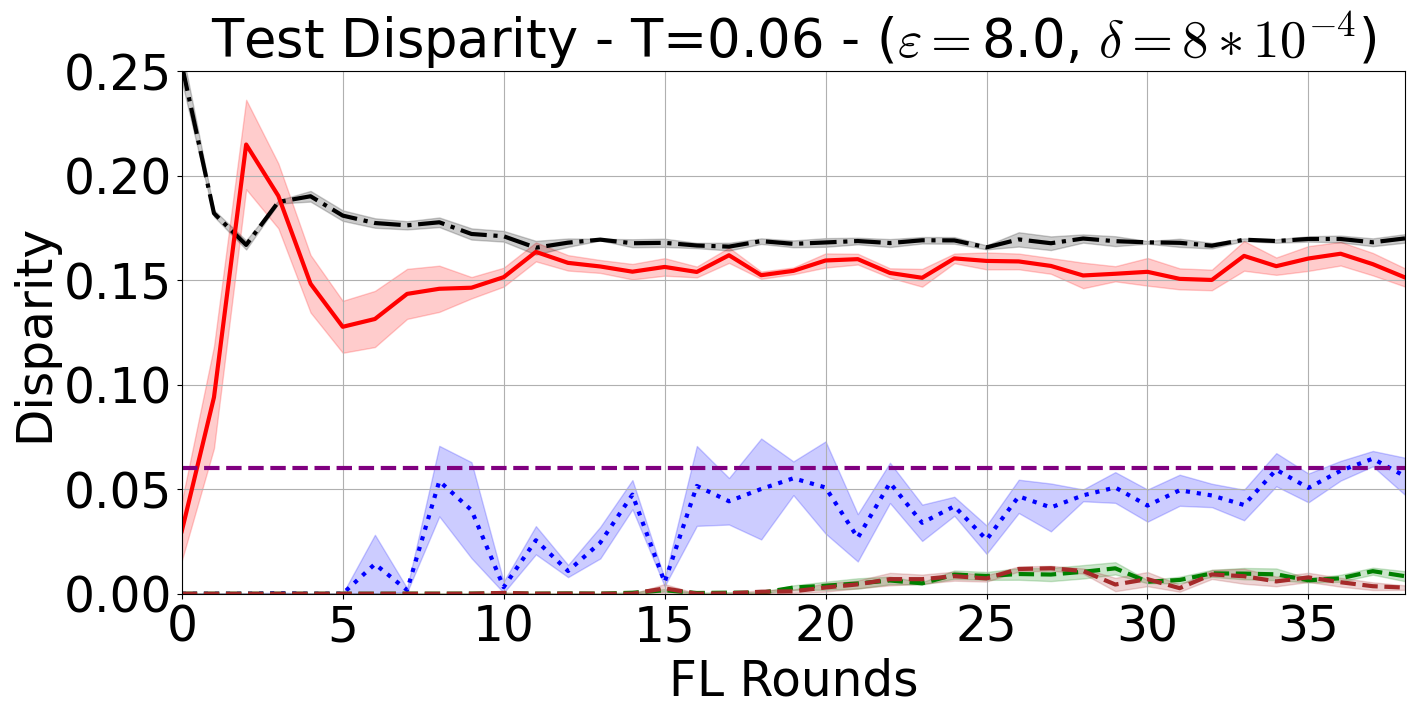}
         \caption{}
         \label{fig:celeba_8:disparity_2}
     \end{subfigure}
     \hfill
     \begin{subfigure}[b]{0.18\textwidth}
         \centering
         \includegraphics[width=\textwidth]{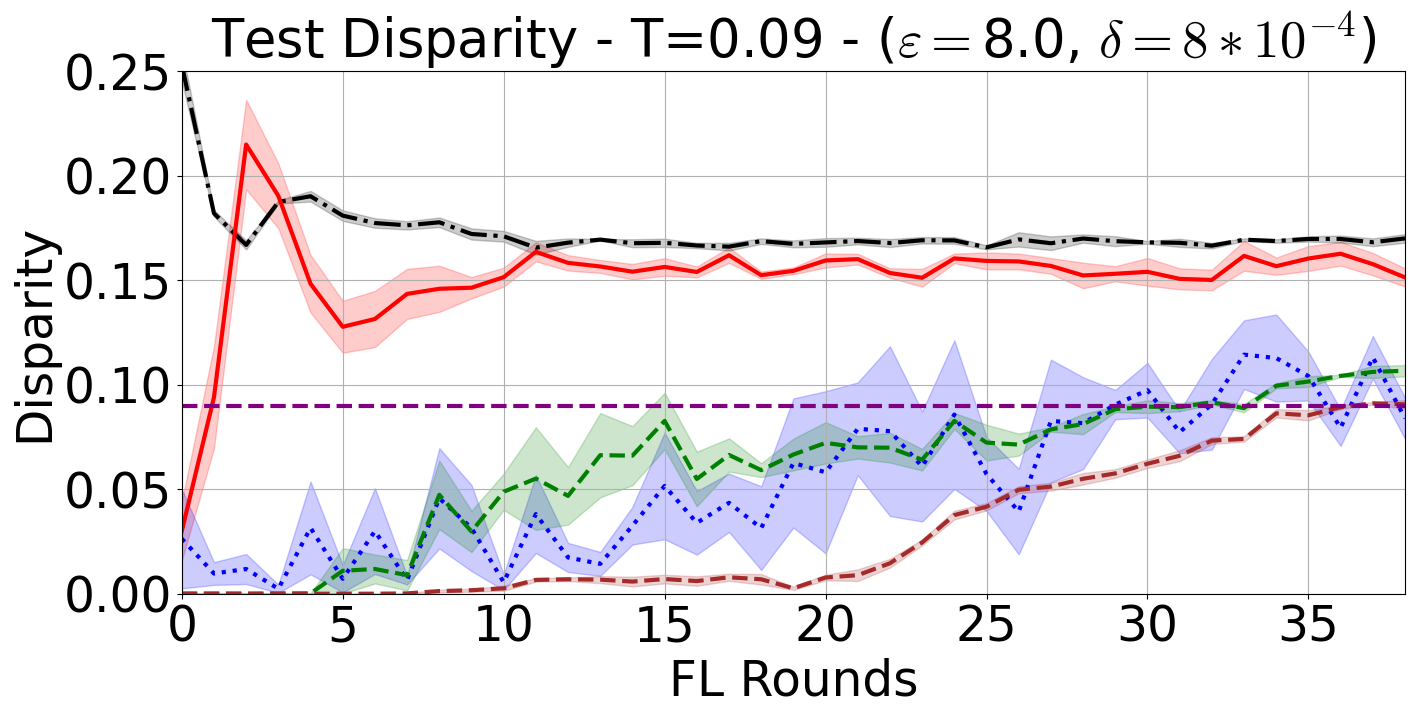}
         \caption{}
         \label{fig:celeba_8:disparity_3}
     \end{subfigure}
     \hfill
     \begin{subfigure}[b]{0.18\textwidth}
         \centering
         \includegraphics[width=\textwidth]{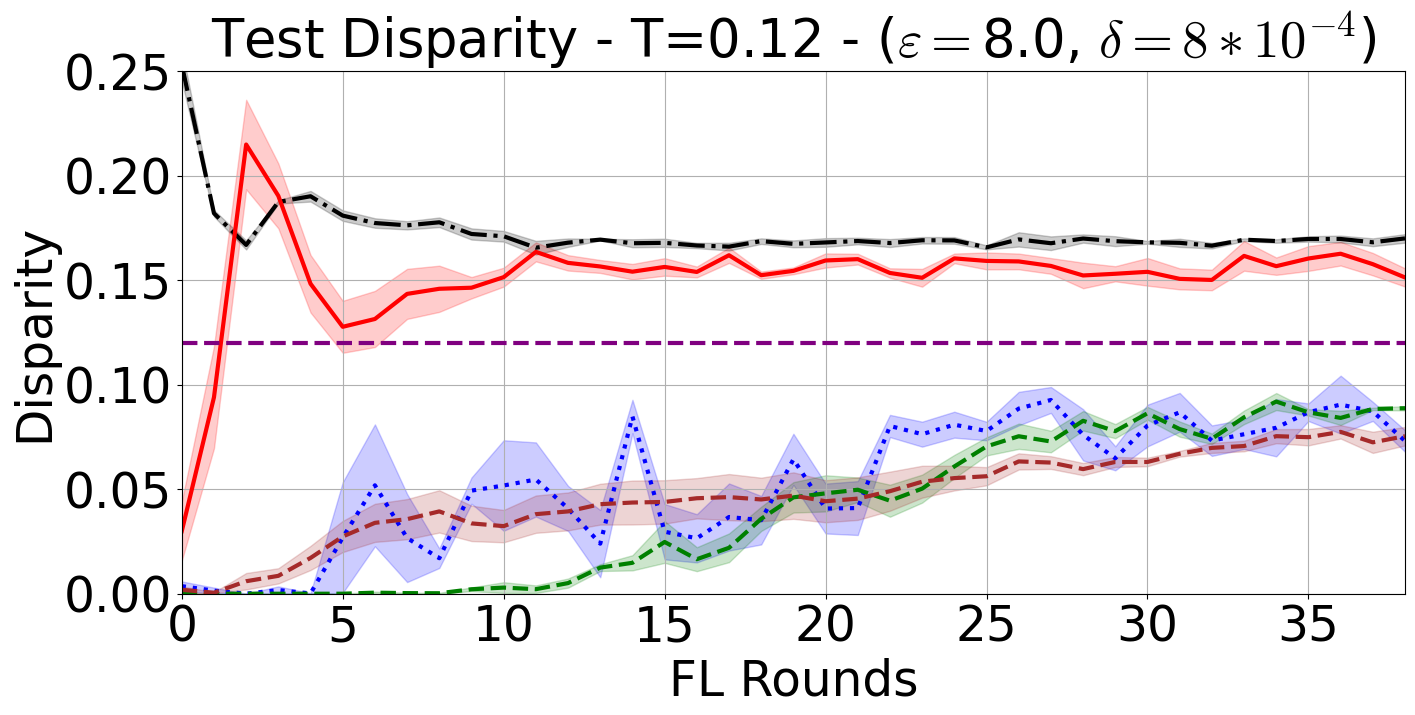}
         \caption{}
         \label{fig:celeba_8:disparity_4}
     \end{subfigure}
     \hfill
     \begin{subfigure}[b]{0.18\textwidth}
         \centering
         \includegraphics[width=\textwidth]{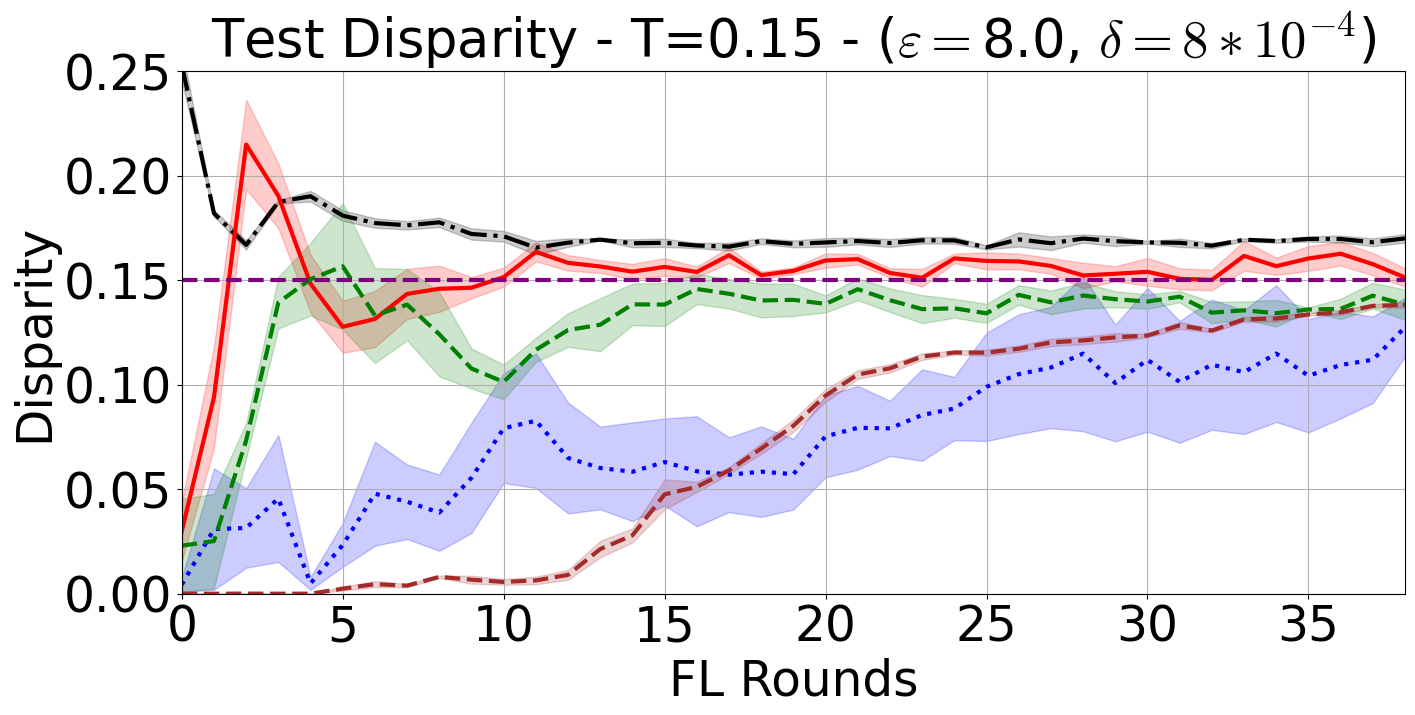}
         \caption{}
         \label{fig:celeba_8:disparity_5}
     \end{subfigure}
     \hfill
     \\
     \vspace{0.30cm}
     \includegraphics[width=0.60\linewidth]{images/legend_disparity.png}\\
    \vspace{0.15cm}
        \caption{Experiment with CelebA. Fairness parameters $T=0.04$, $T=0.06$ and $T=0.09$, privacy parameters ($\varepsilon=8, \delta=8 \times  10^{-4}$). Figures ~\ref{fig:celeba_8:accuracy_1}, ~\ref{fig:celeba_8:accuracy_2}, ~\ref{fig:celeba_8:accuracy_3},~\ref{fig:celeba_8:accuracy_4} and ~\ref{fig:celeba_8:accuracy_5} show the test accuracy of training model while Figures ~\ref{fig:celeba_8:disparity_1}, ~\ref{fig:celeba_8:disparity_2},~\ref{fig:celeba_8:disparity_3}, ~\ref{fig:celeba_8:disparity_4} and ~\ref{fig:celeba_8:disparity_5} show the model disparity.}
        \label{fig:celeba_8}
\end{figure*}

Figure ~\ref{fig:celeba_8} shows the results obtained with CelebA using the combinations of privacy target $(\epsilon=8, \gamma=8 \times  10^{-4})$ with target disparity $T=0.04$, $T=0.06$, $T=0.09$, $T=0.012$ and $T=0.15$. The results are similar to the ones we have shown in Section ~\ref{sec:results}. We report in Table ~\ref{tab:results_celeba} a comparison of the different use of regularization and DP. In particular, here we compare the following setups: Baseline, Only DP, Only Fairness, Fairness and DP both with Tunable and Fixed $\lambda$.

\begin{figure*}
     \centering
     \captionsetup{justification=justified}
     \begin{subfigure}[b]{0.18\textwidth}
         \centering
         \includegraphics[width=\textwidth]{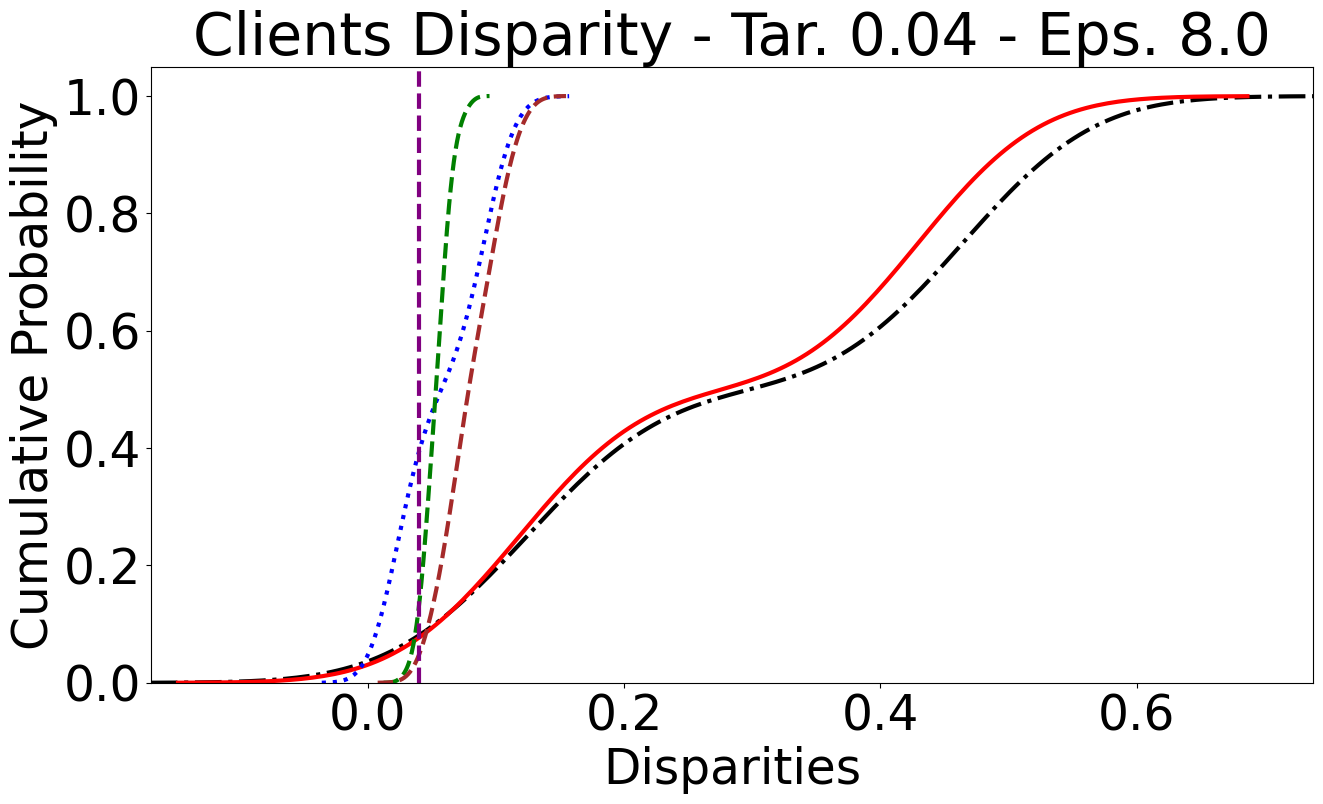}
         \caption{}
         \label{fig:celeba_local_8:cdf_1}
     \end{subfigure}
     \hfill
     \begin{subfigure}[b]{0.18\textwidth}
         \centering
         \includegraphics[width=\textwidth]{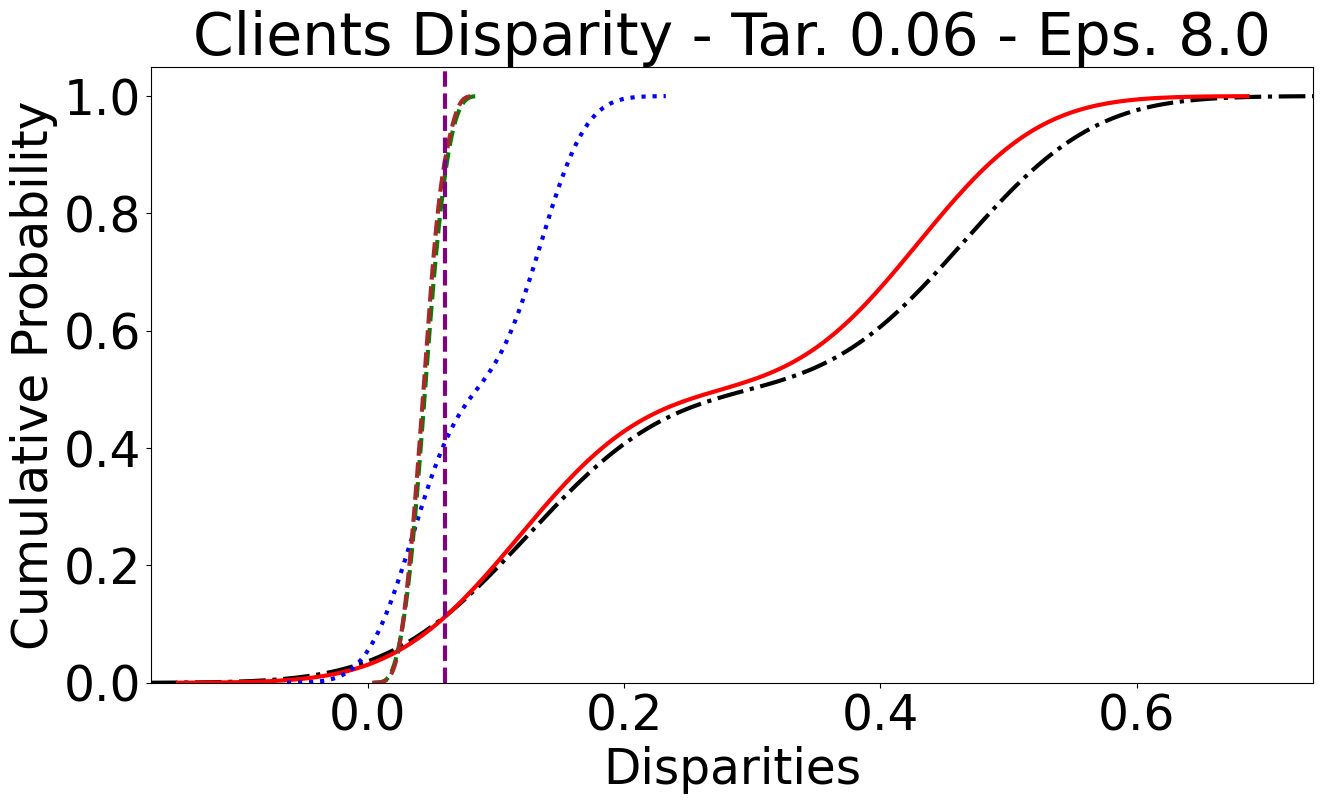}
         \caption{}
         \label{fig:celeba_local_8:cdf_2}
     \end{subfigure}
     \hfill
     \begin{subfigure}[b]{0.18\textwidth}
         \centering
         \includegraphics[width=\textwidth]{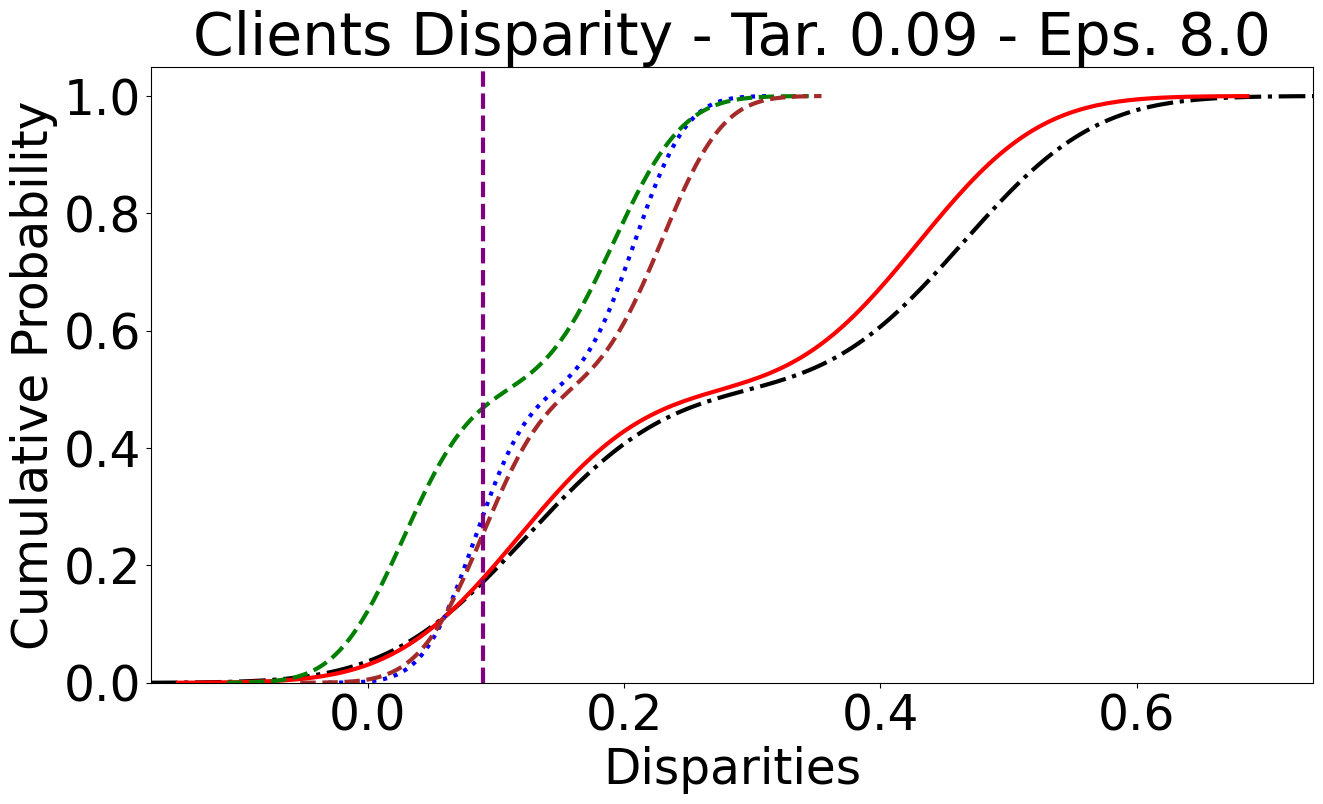}
         \caption{}
         \label{fig:celeba_local_8:cdf_3}
     \end{subfigure}
     \begin{subfigure}[b]{0.18\textwidth}
         \centering
         \includegraphics[width=\textwidth]{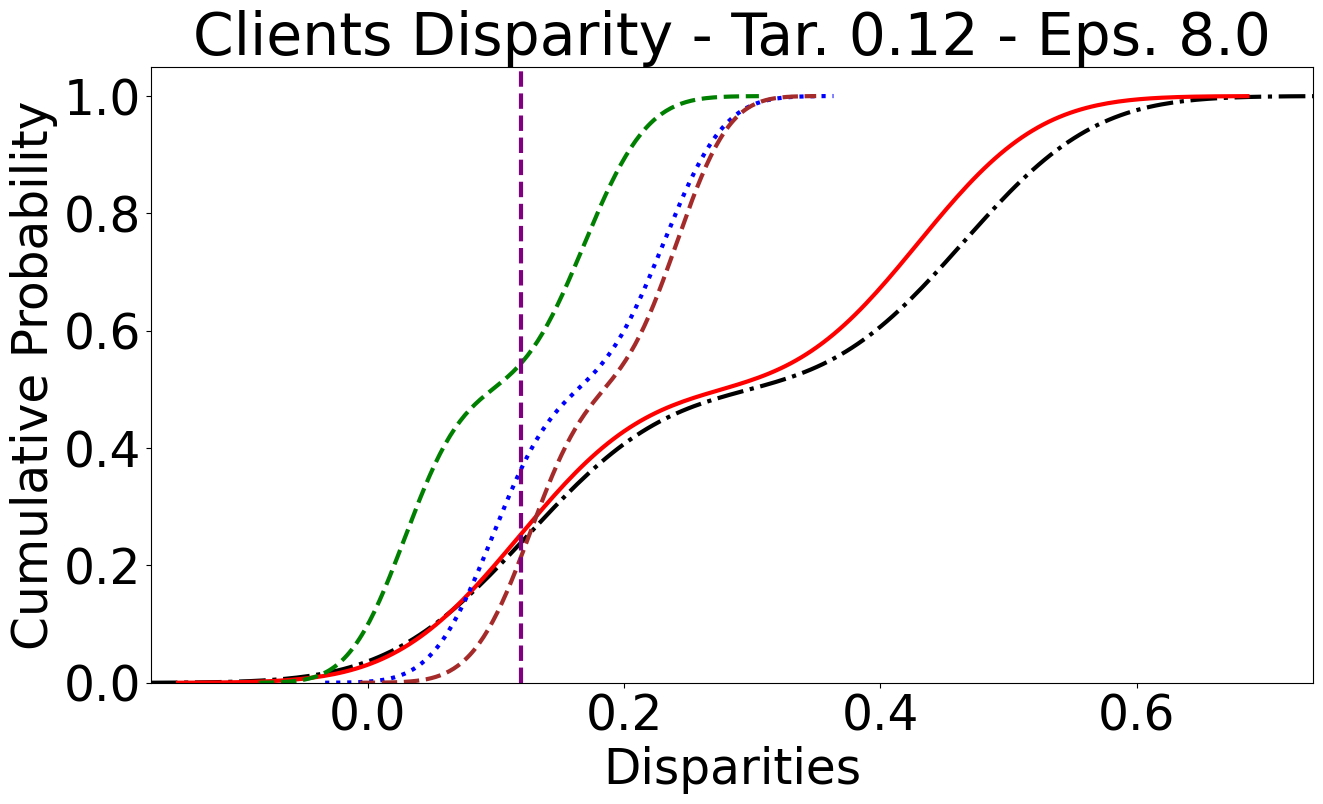}
         \caption{}
         \label{fig:celeba_local_8:cdf_4}
     \end{subfigure}
     \hfill
     \begin{subfigure}[b]{0.18\textwidth}
         \centering
         \includegraphics[width=\textwidth]{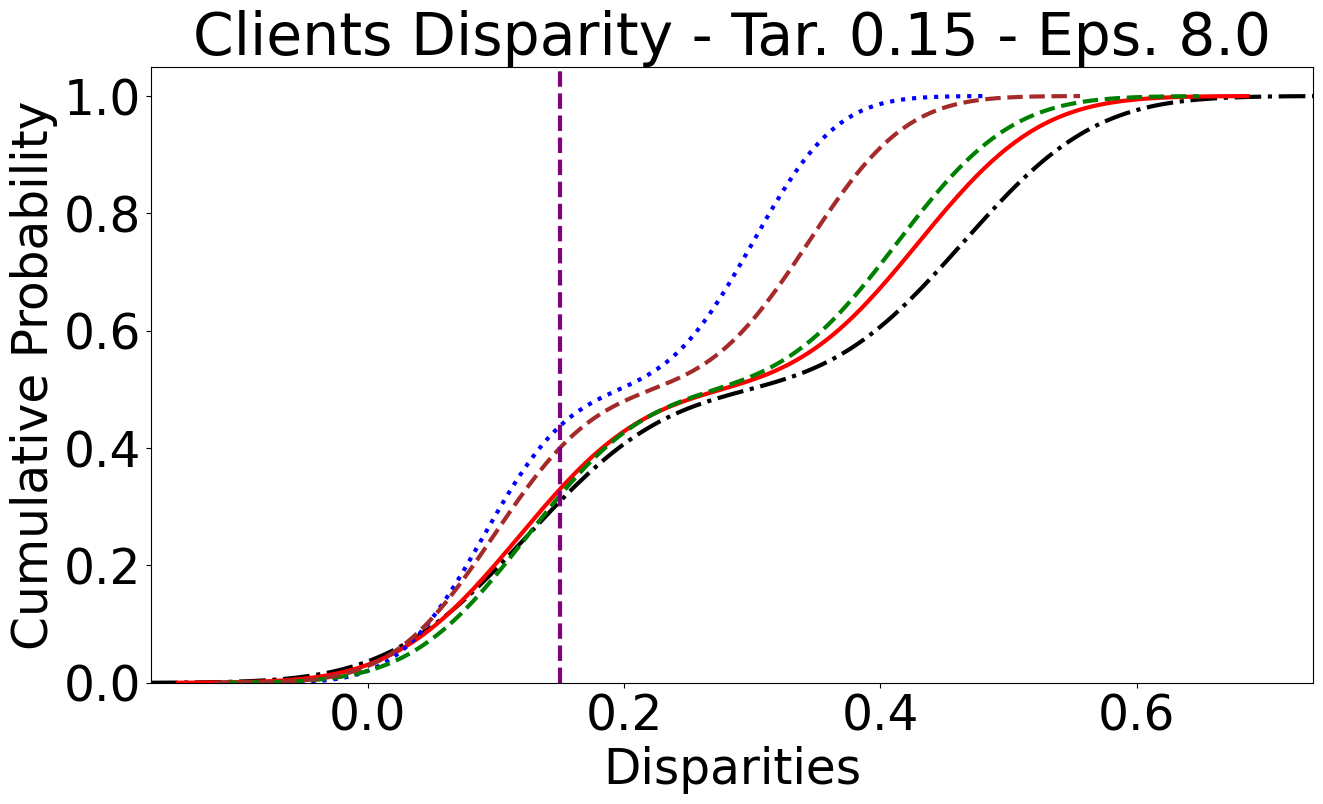}
         \caption{}
         \label{fig:celeba_local_8:cdf_5}
     \end{subfigure}\\
     \vspace{0.30cm}
     \includegraphics[width=0.60\linewidth]{images/legend_disparity.png}\\
        \vspace{0.15cm}
        \caption{Experiment with CelebA. Fairness parameters $T=0.04$, $T=0.06$, $T=0.09$, $T=0.12$ and $T=0.15$ privacy parameters ($\varepsilon=8, \delta=8 \times  10^{-4}$). The Figures show the cumulative distribution function of the local disparities of the clients.}
        \label{fig:celeba_local_8}
\end{figure*}

\subsection{Dutch}

\begin{figure*}
     \centering
     \captionsetup{justification=justified}
     \begin{subfigure}[b]{0.18\textwidth}
         \centering
         \includegraphics[width=\textwidth]{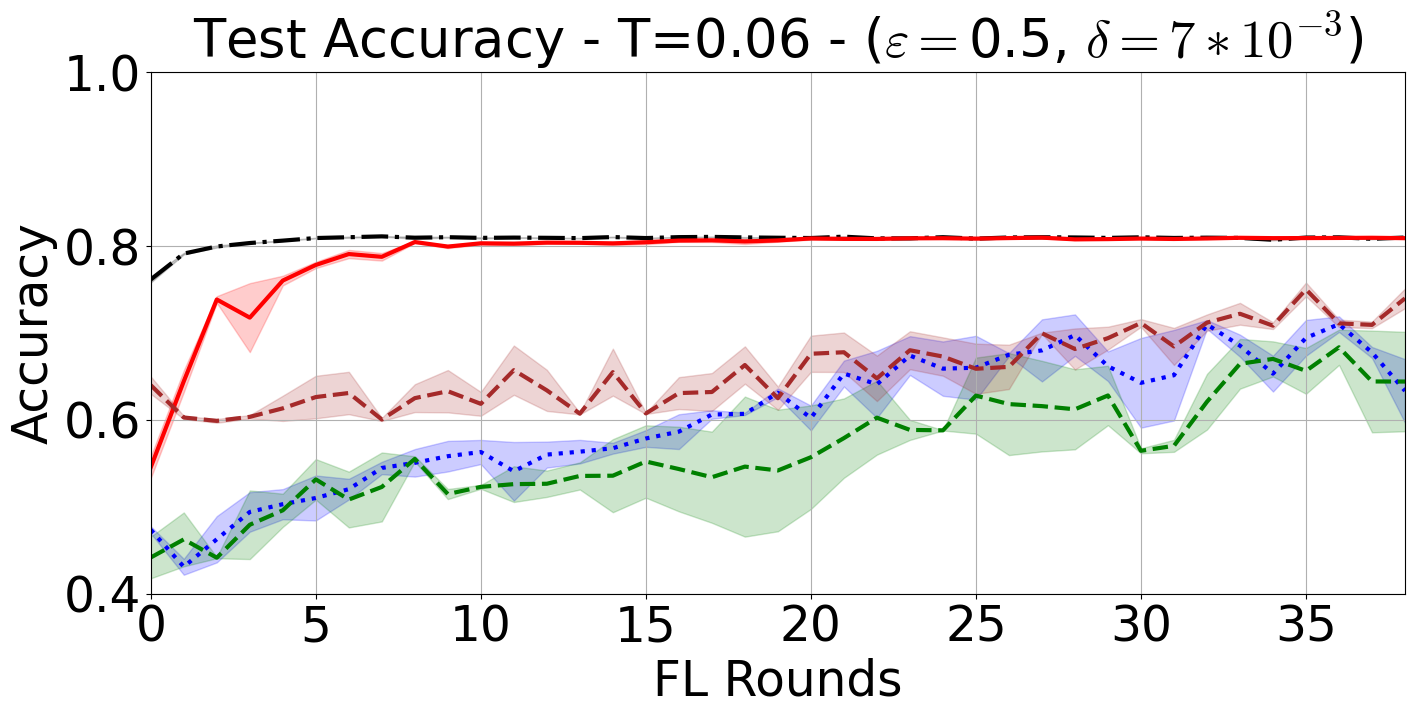}
         \caption{}
         \label{fig:dutch_05:accuracy_1}
     \end{subfigure}
     \hfill
     \begin{subfigure}[b]{0.18\textwidth}
         \centering
         \includegraphics[width=\textwidth]{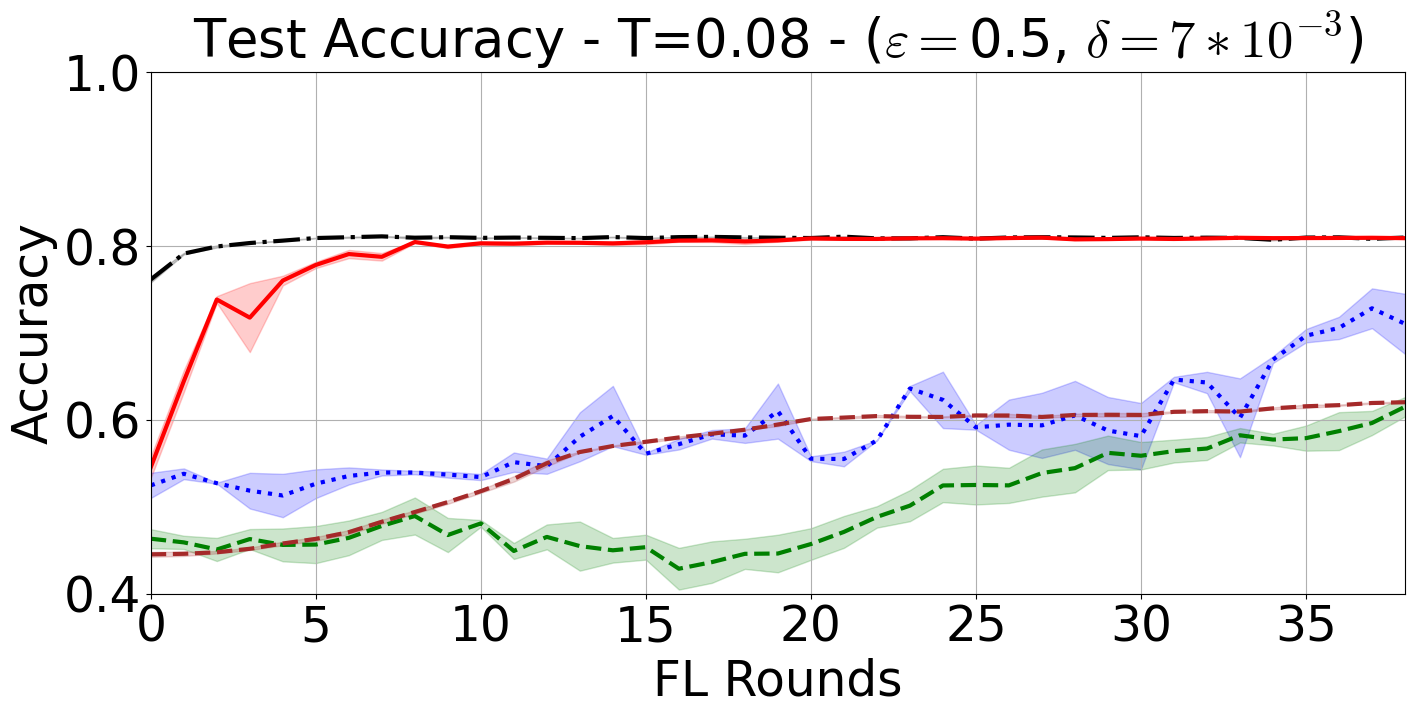}
         \caption{}
         \label{fig:dutch_05:accuracy_2}
     \end{subfigure}
     \hfill
     \begin{subfigure}[b]{0.18\textwidth}
         \centering
         \includegraphics[width=\textwidth]{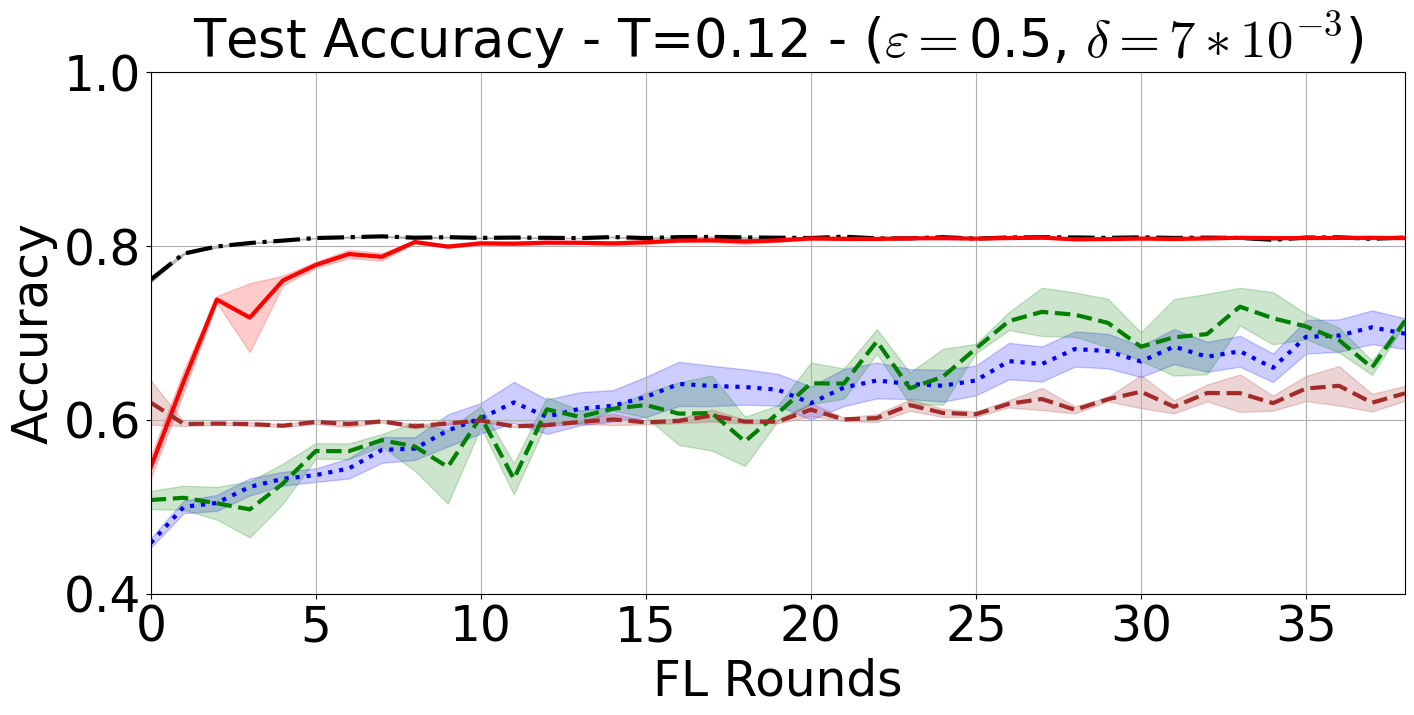}
         \caption{}
         \label{fig:dutch_05:accuracy_3}
     \end{subfigure}
     \hfill
     \begin{subfigure}[b]{0.18\textwidth}
         \centering
         \includegraphics[width=\textwidth]{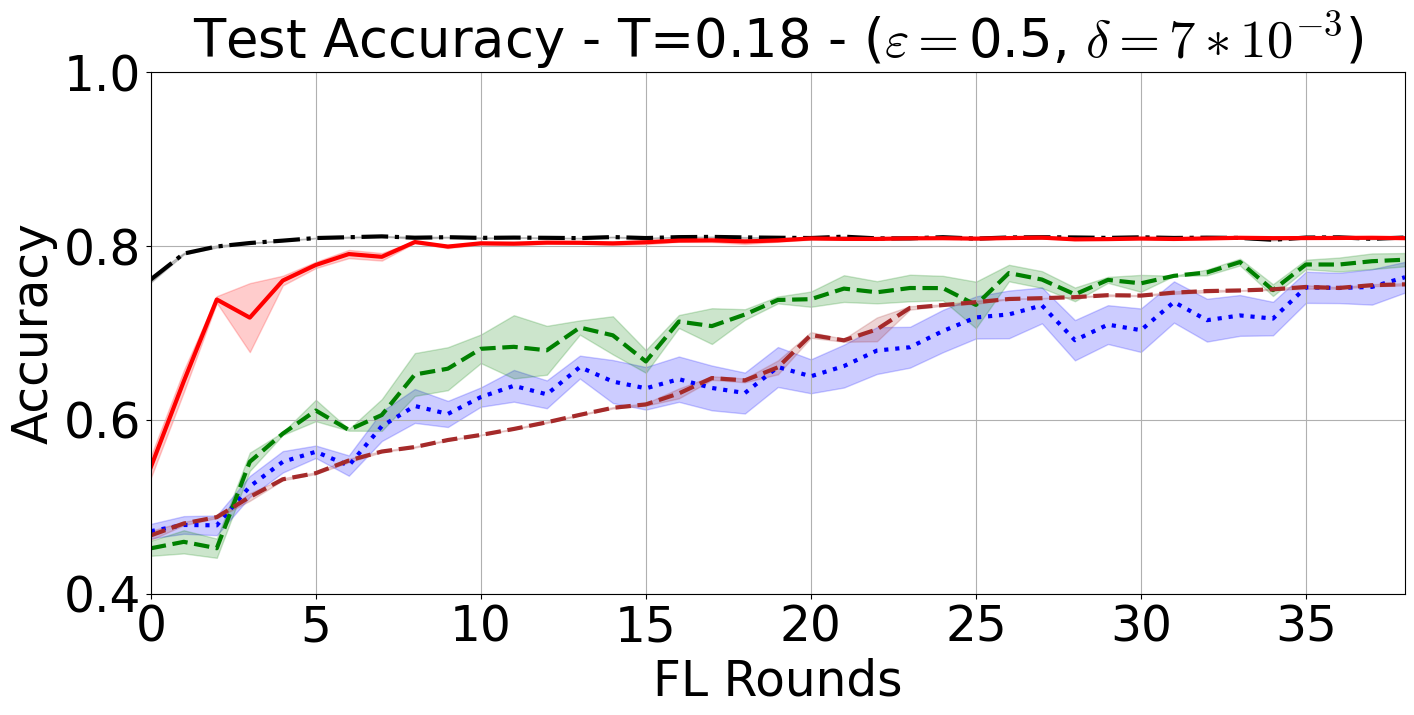}
         \caption{}
         \label{fig:dutch_05:accuracy_4}
     \end{subfigure}
     \hfill
     \begin{subfigure}[b]{0.18\textwidth}
         \centering
         \includegraphics[width=\textwidth]{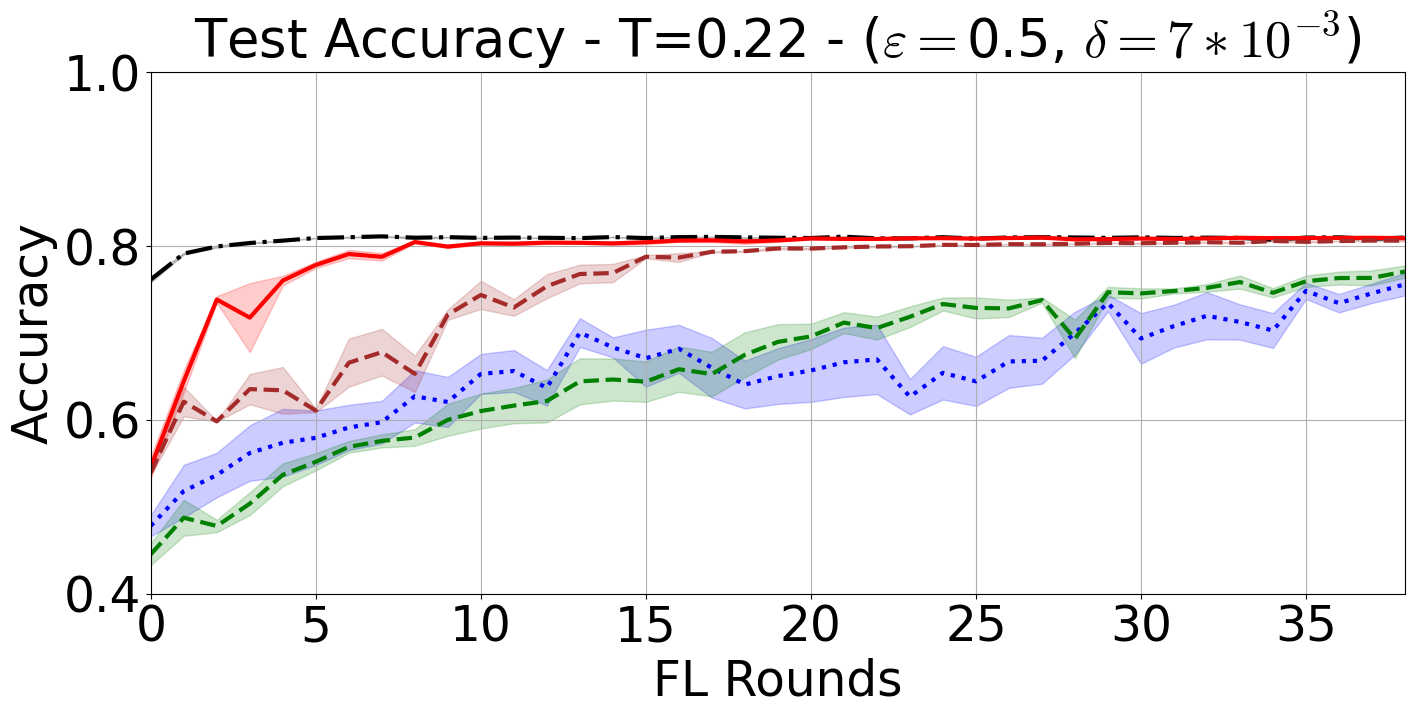}
         \caption{}
         \label{fig:dutch_05:accuracy_5}
     \end{subfigure}\\
     \vspace{0.30cm}

    \begin{subfigure}[b]{0.18\textwidth}
         \centering
         \includegraphics[width=\textwidth]{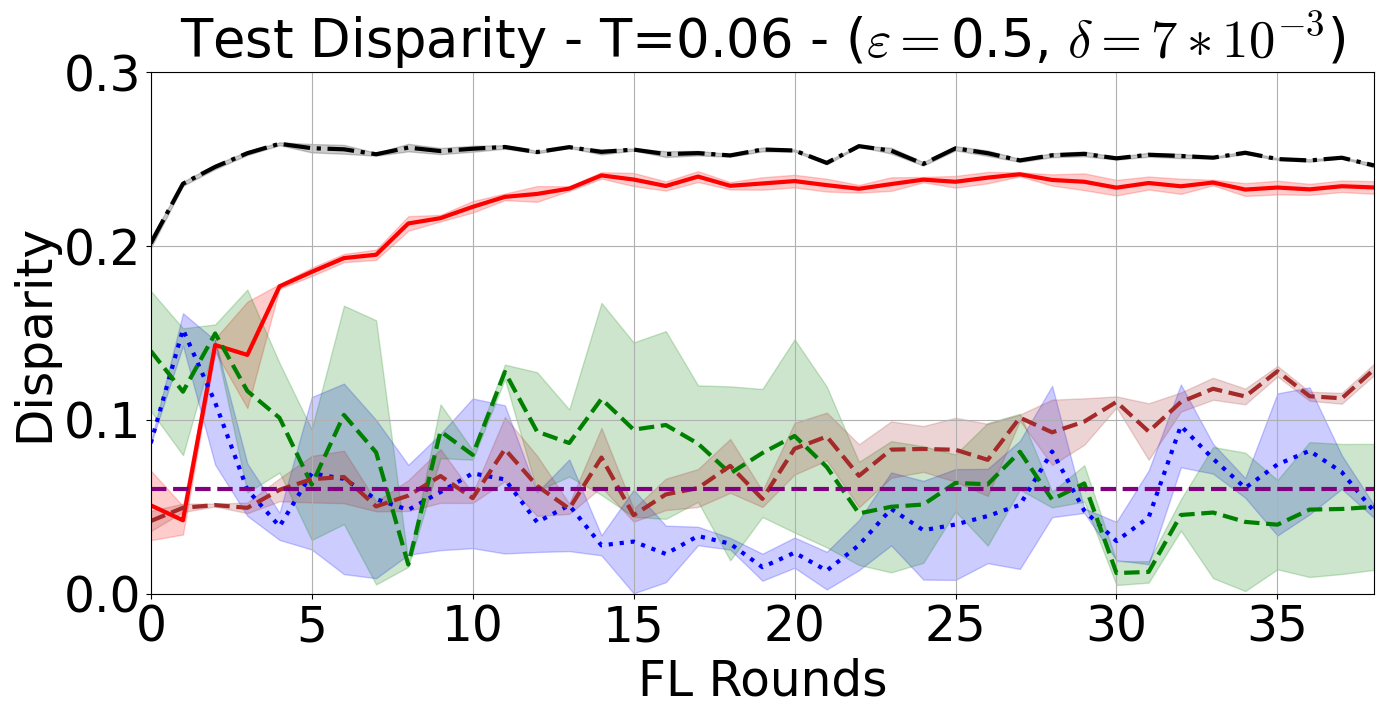}
         \caption{}
         \label{fig:dutch_05:disparity_1}
     \end{subfigure}
     \hfill
     \begin{subfigure}[b]{0.18\textwidth}
         \centering
         \includegraphics[width=\textwidth]{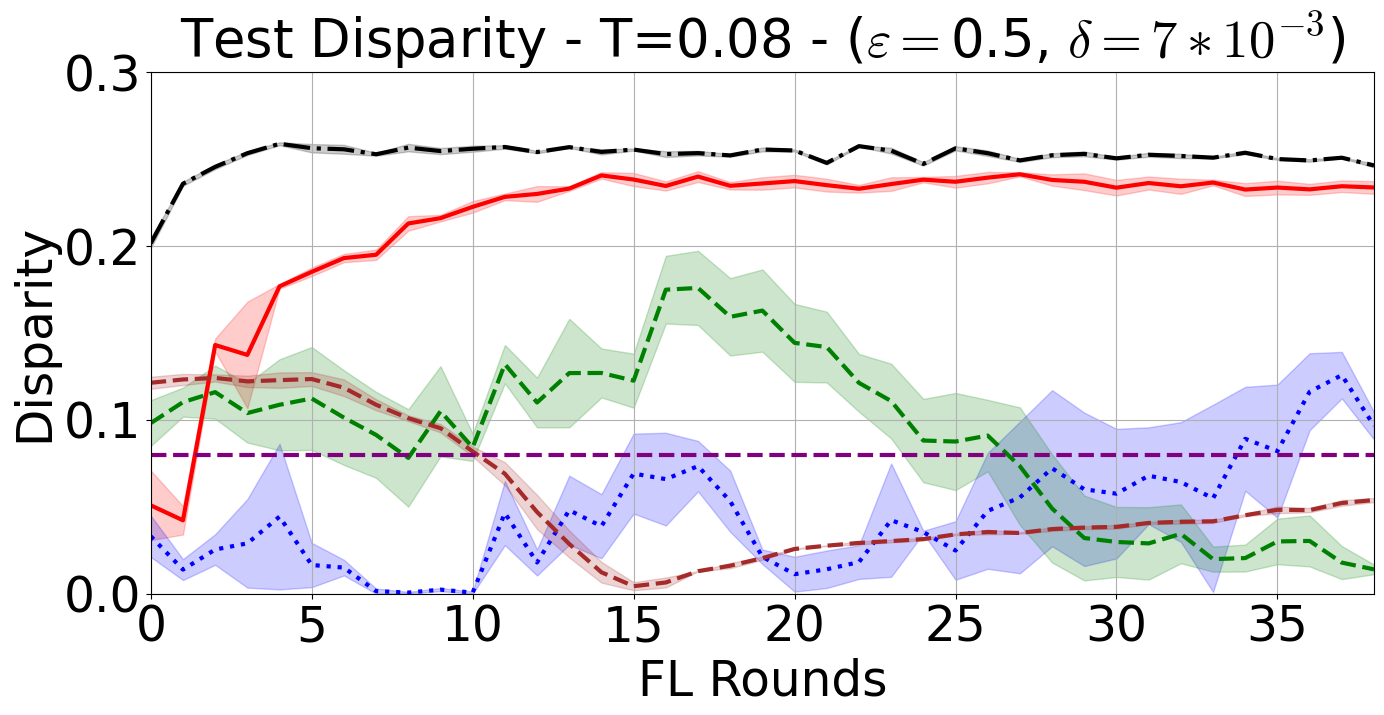}
         \caption{}
         \label{fig:dutch_05:disparity_2}
     \end{subfigure}
     \hfill
     \begin{subfigure}[b]{0.18\textwidth}
         \centering
         \includegraphics[width=\textwidth]{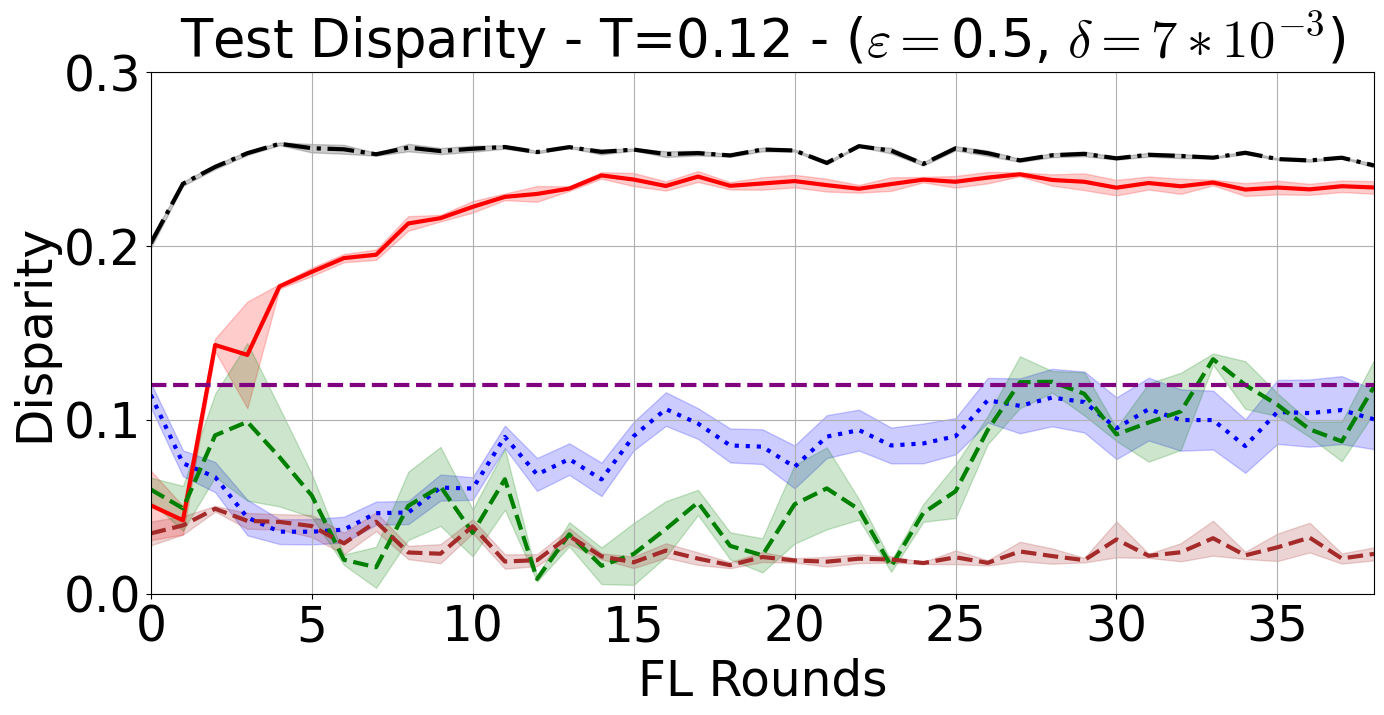}
         \caption{}
         \label{fig:dutch_05:disparity_3}
     \end{subfigure}
     \hfill
     \begin{subfigure}[b]{0.18\textwidth}
         \centering
         \includegraphics[width=\textwidth]{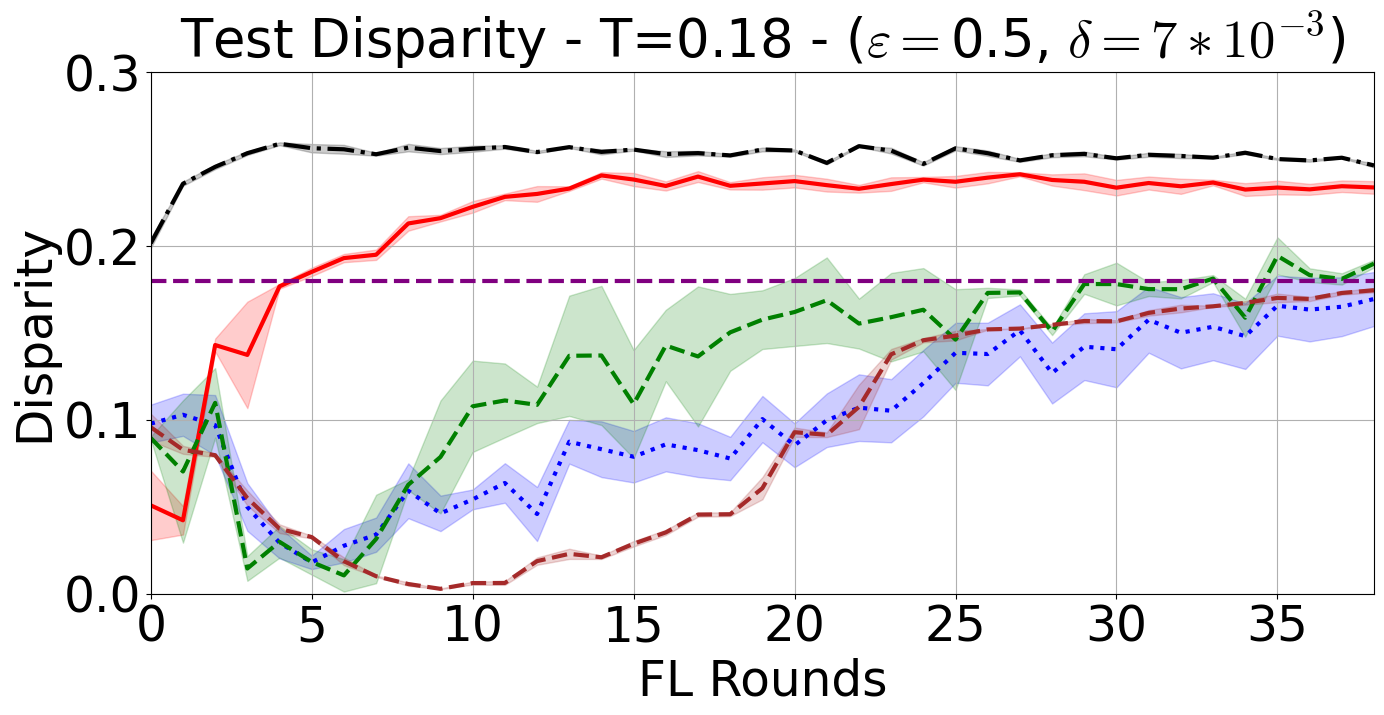}
         \caption{}
         \label{fig:dutch_05:disparity_4}
     \end{subfigure}
     \hfill
     \begin{subfigure}[b]{0.18\textwidth}
         \centering
         \includegraphics[width=\textwidth]{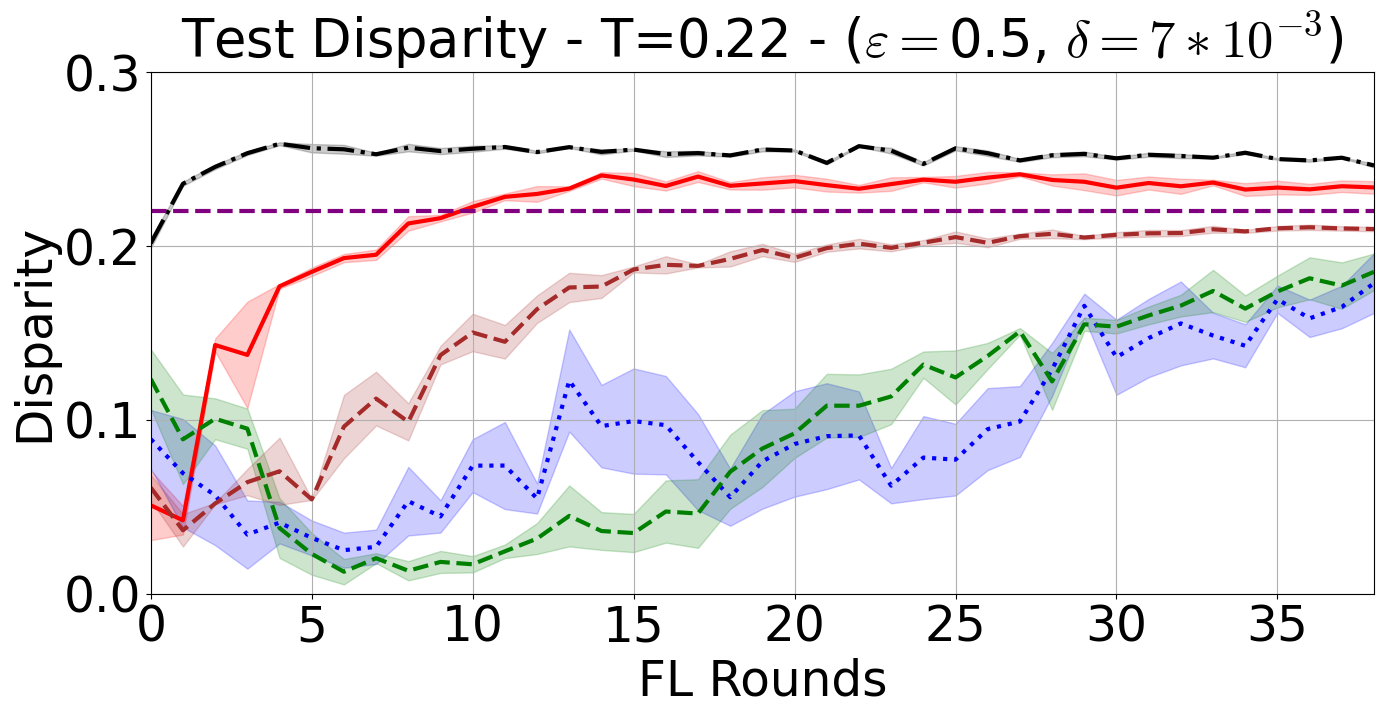}
         \caption{}
         \label{fig:dutch_05:disparity_5}
     \end{subfigure}\\
     \vspace{0.3cm}
     \includegraphics[width=0.60\linewidth]{images/legend_disparity.png}\\
        \vspace{0.15cm}
        \caption{Experiment with Dutch. Fairness parameters $T=0.04$, $T=0.06$, $T=0.09$, $T=0.12$ and $T=0.15$, privacy parameters ($\epsilon=0.5$, $\delta=7 \times  10^{-3}$). Figures ~\ref{fig:dutch_05:accuracy_1}, ~\ref{fig:dutch_05:accuracy_2} and ~\ref{fig:dutch_05:accuracy_3} show the test accuracy of training model while Figures ~\ref{fig:dutch_05:disparity_1}, ~\ref{fig:dutch_05:disparity_2} and ~\ref{fig:dutch_05:disparity_3} show the model disparity.}
        \label{fig:dutch_05}
\end{figure*}

\begin{figure*}
     \centering
     \captionsetup{justification=justified}
     \begin{subfigure}[b]{0.18\textwidth}
         \centering
         \includegraphics[width=\textwidth]{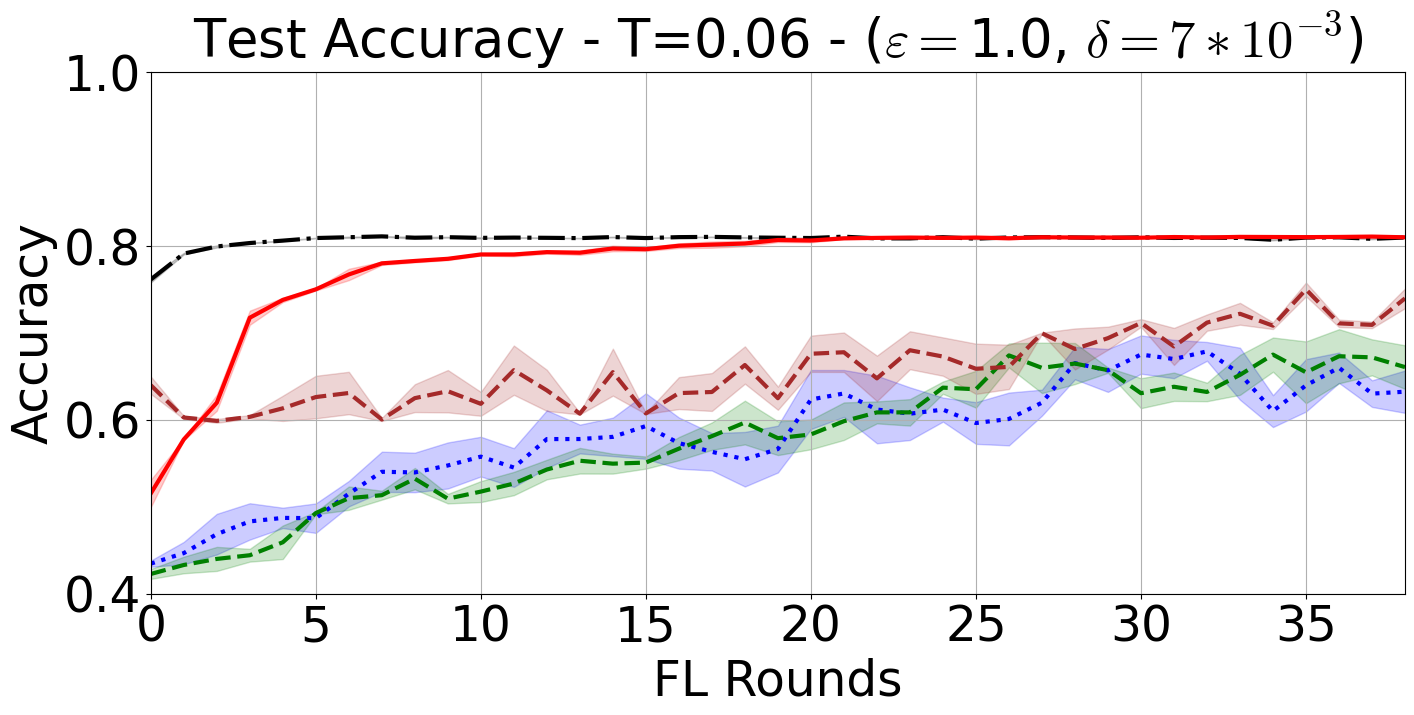}
         \caption{}
         \label{fig:dutch_1:accuracy_1}
     \end{subfigure}
     \hfill
     \begin{subfigure}[b]{0.18\textwidth}
         \centering
         \includegraphics[width=\textwidth]{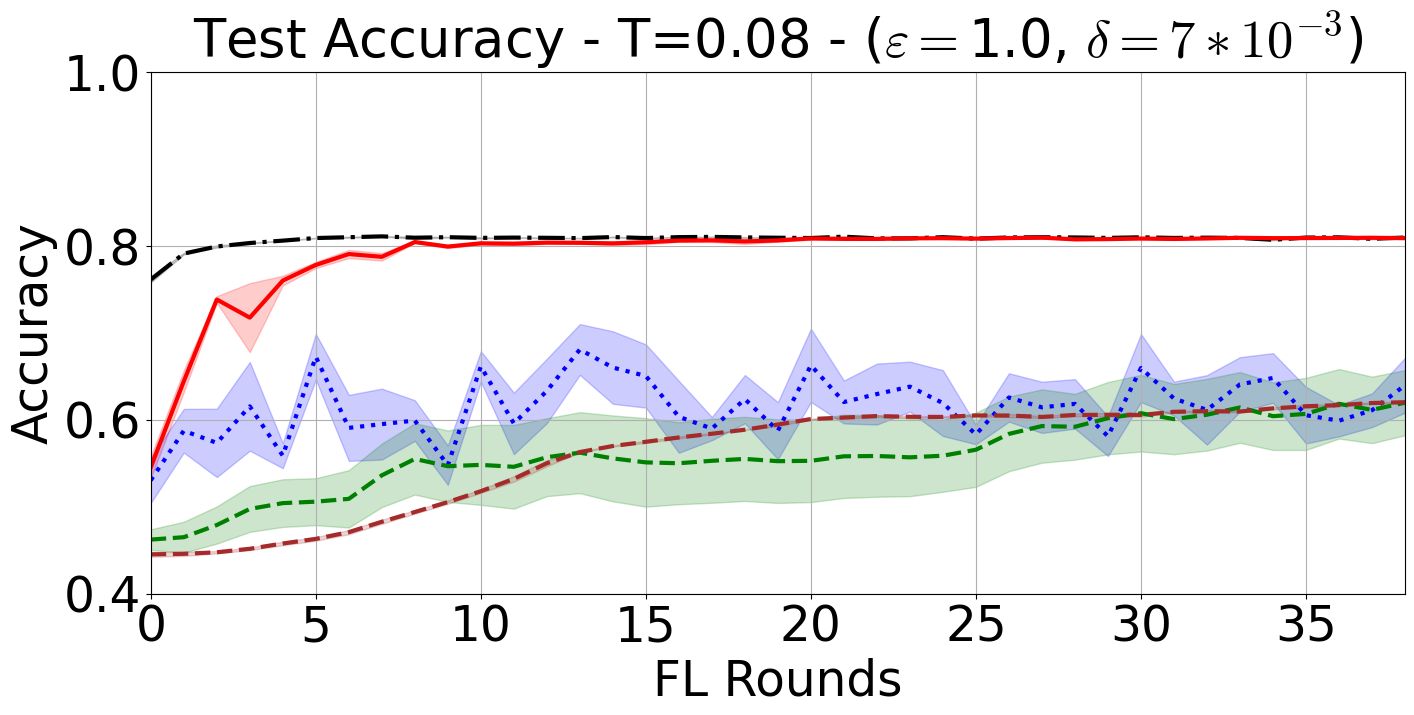}
         \caption{}
         \label{fig:dutch_1:accuracy_2}
     \end{subfigure}
     \hfill
     \begin{subfigure}[b]{0.18\textwidth}
         \centering
         \includegraphics[width=\textwidth]{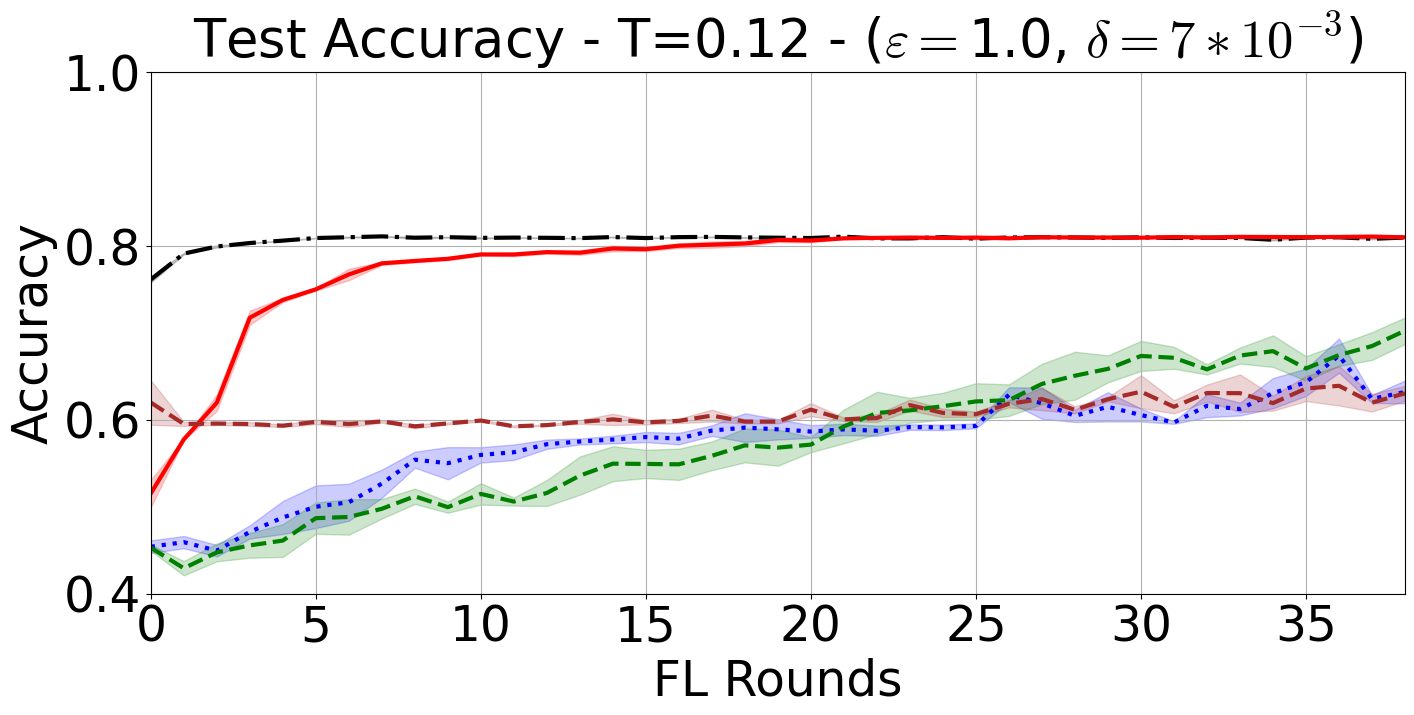}
         \caption{}
         \label{fig:dutch_1:accuracy_3}
     \end{subfigure}
     \hfill
     \begin{subfigure}[b]{0.18\textwidth}
         \centering
         \includegraphics[width=\textwidth]{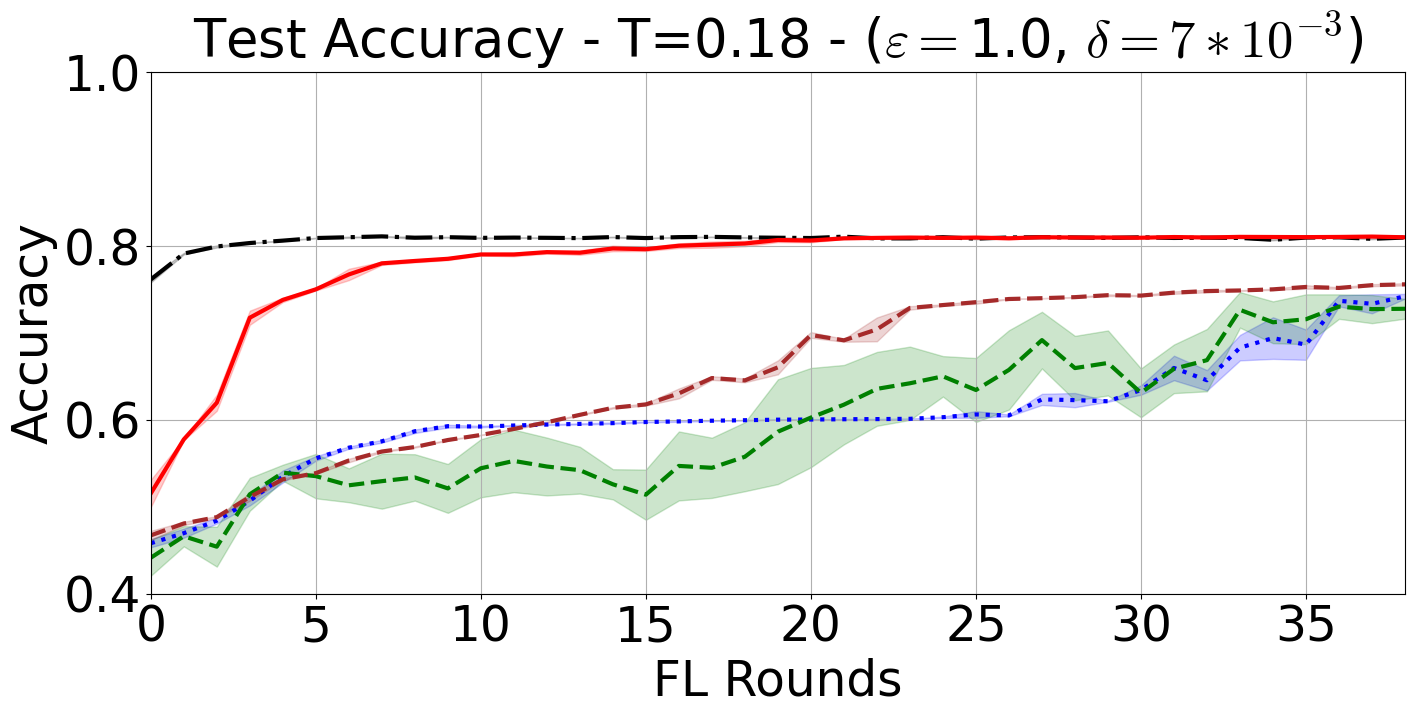}
         \caption{}
         \label{fig:dutch_1:accuracy_4}
     \end{subfigure}
     \hfill
     \begin{subfigure}[b]{0.18\textwidth}
         \centering
         \includegraphics[width=\textwidth]{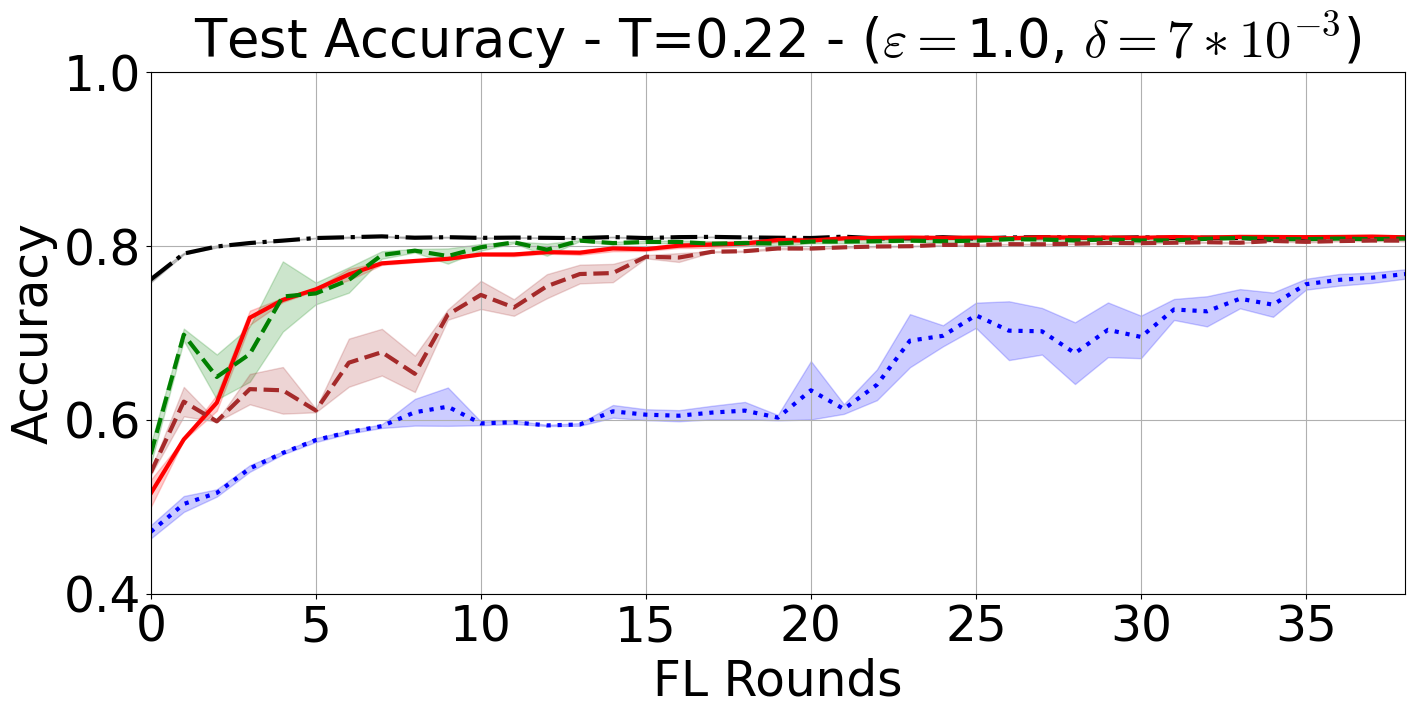}
         \caption{}
         \label{fig:dutch_1:accuracy_5}
     \end{subfigure}\\
    \vspace{0.30cm}

    \begin{subfigure}[b]{0.18\textwidth}
         \centering
         \includegraphics[width=\textwidth]{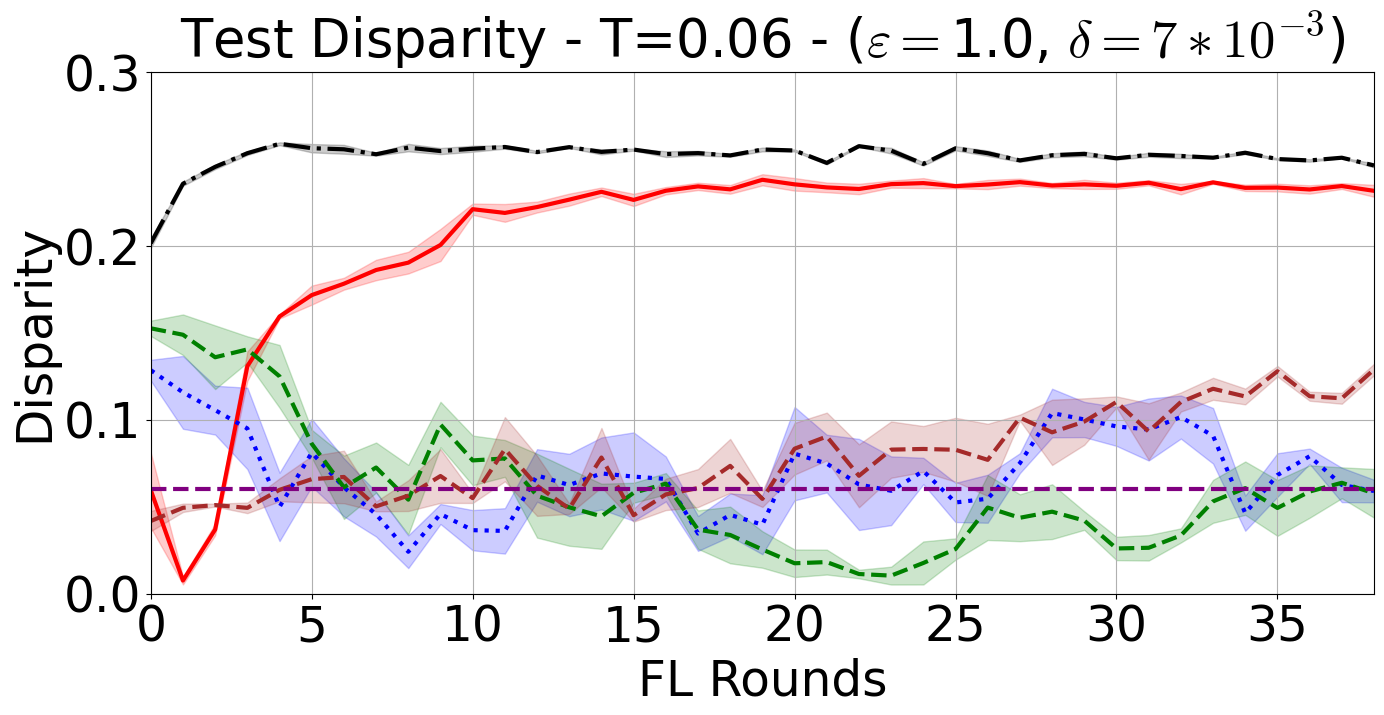}
         \caption{}
         \label{fig:dutch_1:disparity_1}
     \end{subfigure}
     \hfill
     \begin{subfigure}[b]{0.18\textwidth}
         \centering
         \includegraphics[width=\textwidth]{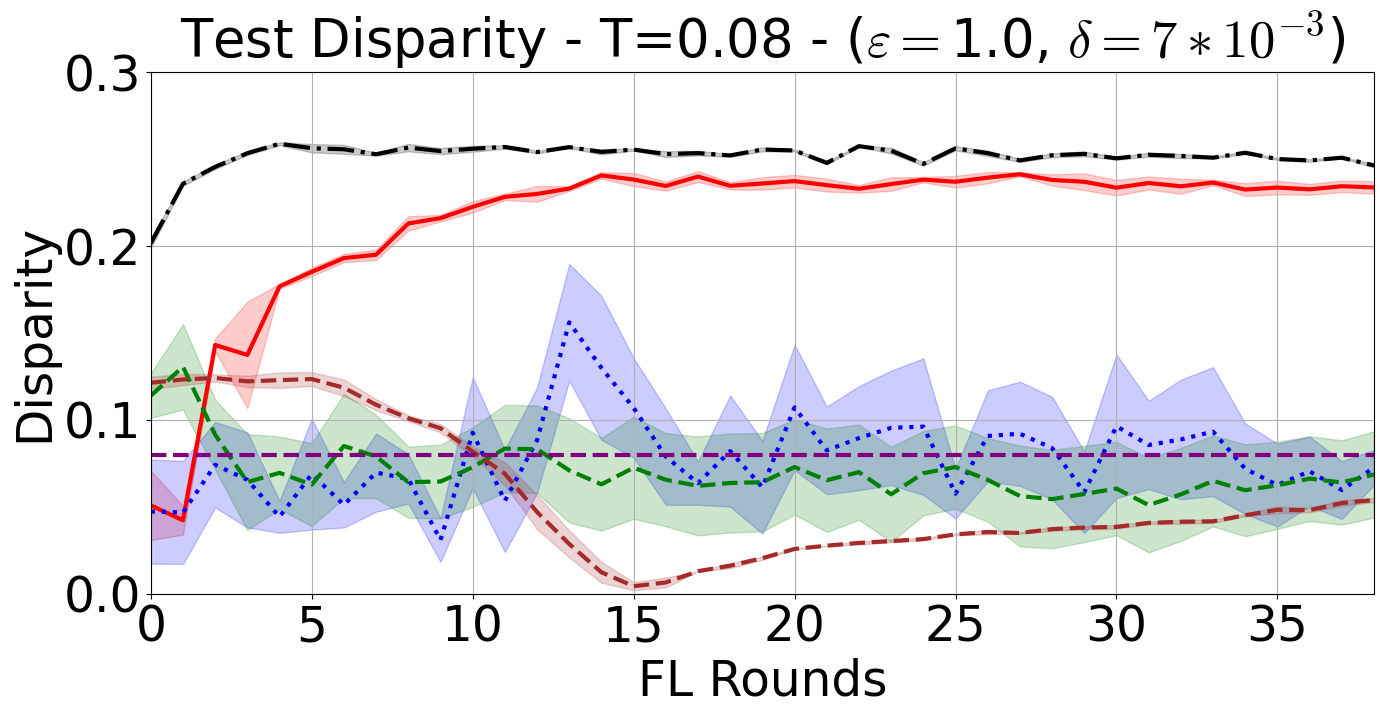}
         \caption{}
         \label{fig:dutch_1:disparity_2}
     \end{subfigure}
     \hfill
     \begin{subfigure}[b]{0.18\textwidth}
         \centering
         \includegraphics[width=\textwidth]{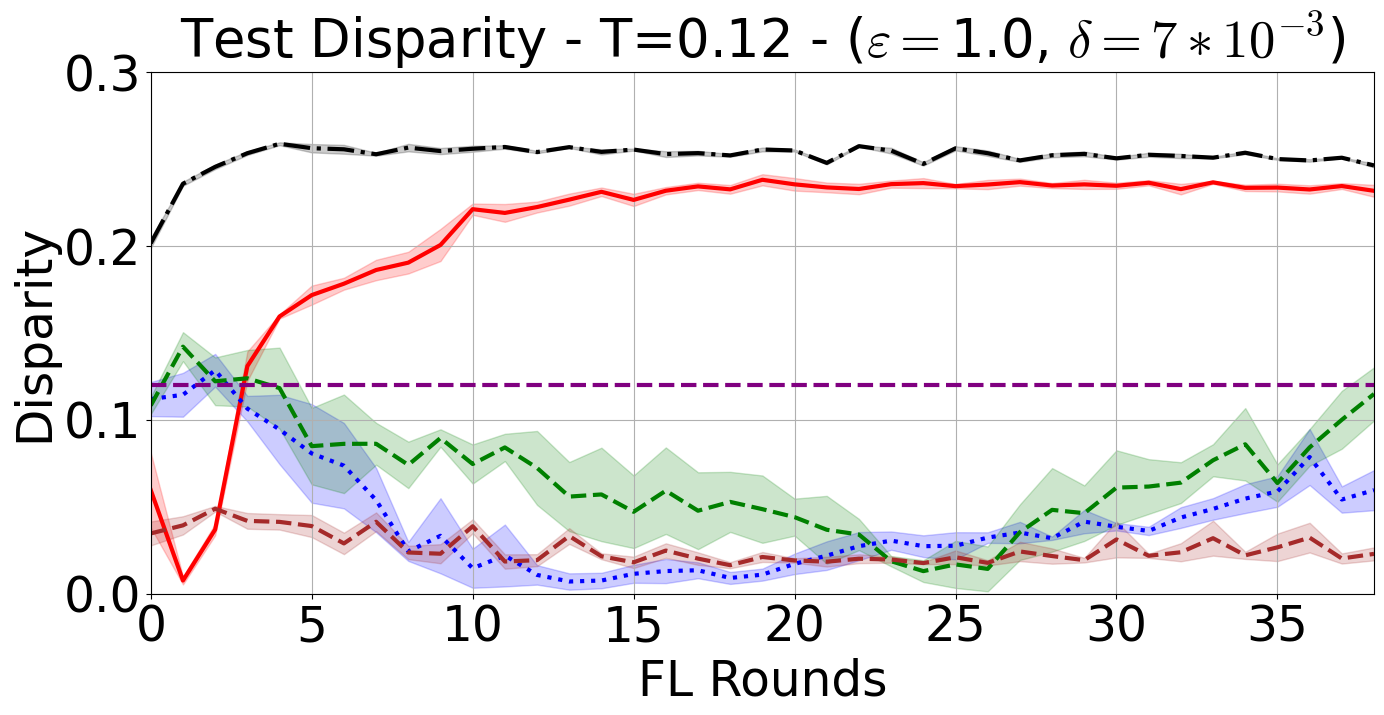}
         \caption{}
         \label{fig:dutch_1:disparity_3}
     \end{subfigure}
     \hfill
     \begin{subfigure}[b]{0.18\textwidth}
         \centering
         \includegraphics[width=\textwidth]{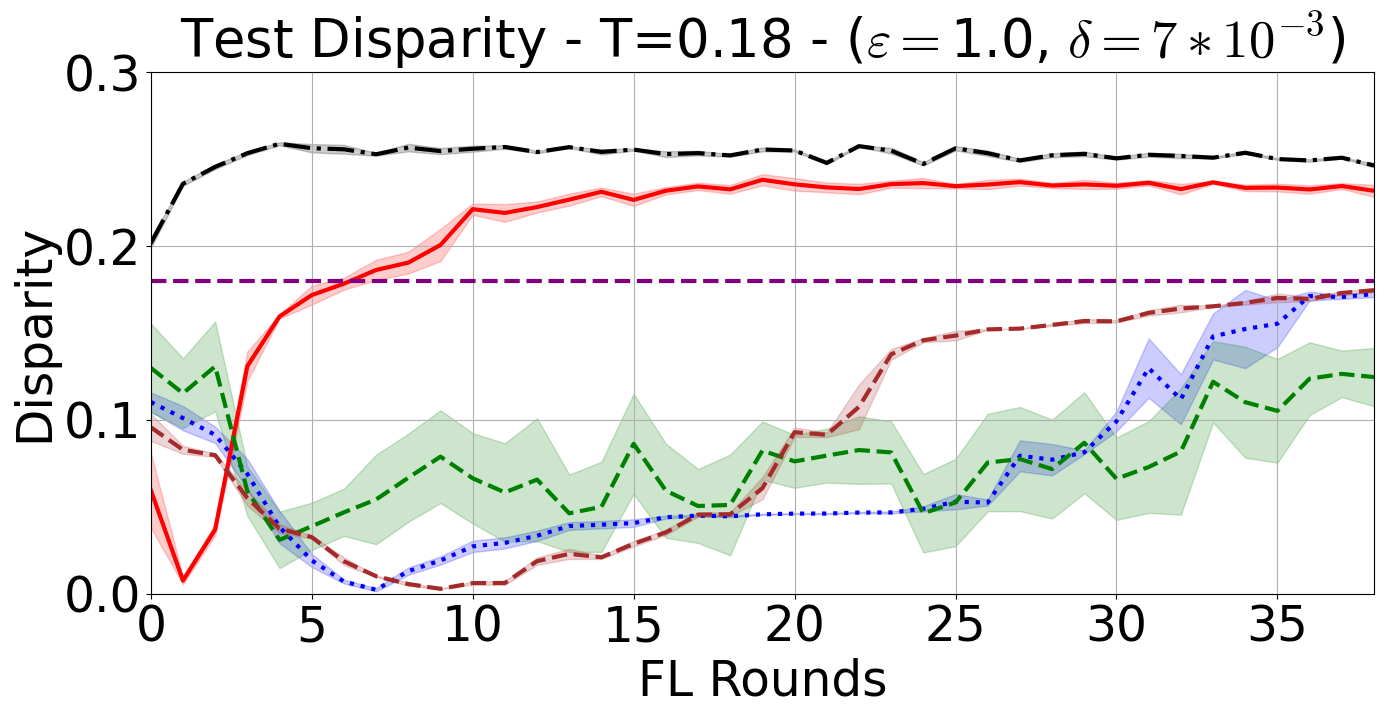}
         \caption{}
         \label{fig:dutch_1:disparity_4}
     \end{subfigure}
     \hfill
     \begin{subfigure}[b]{0.18\textwidth}
         \centering
         \includegraphics[width=\textwidth]{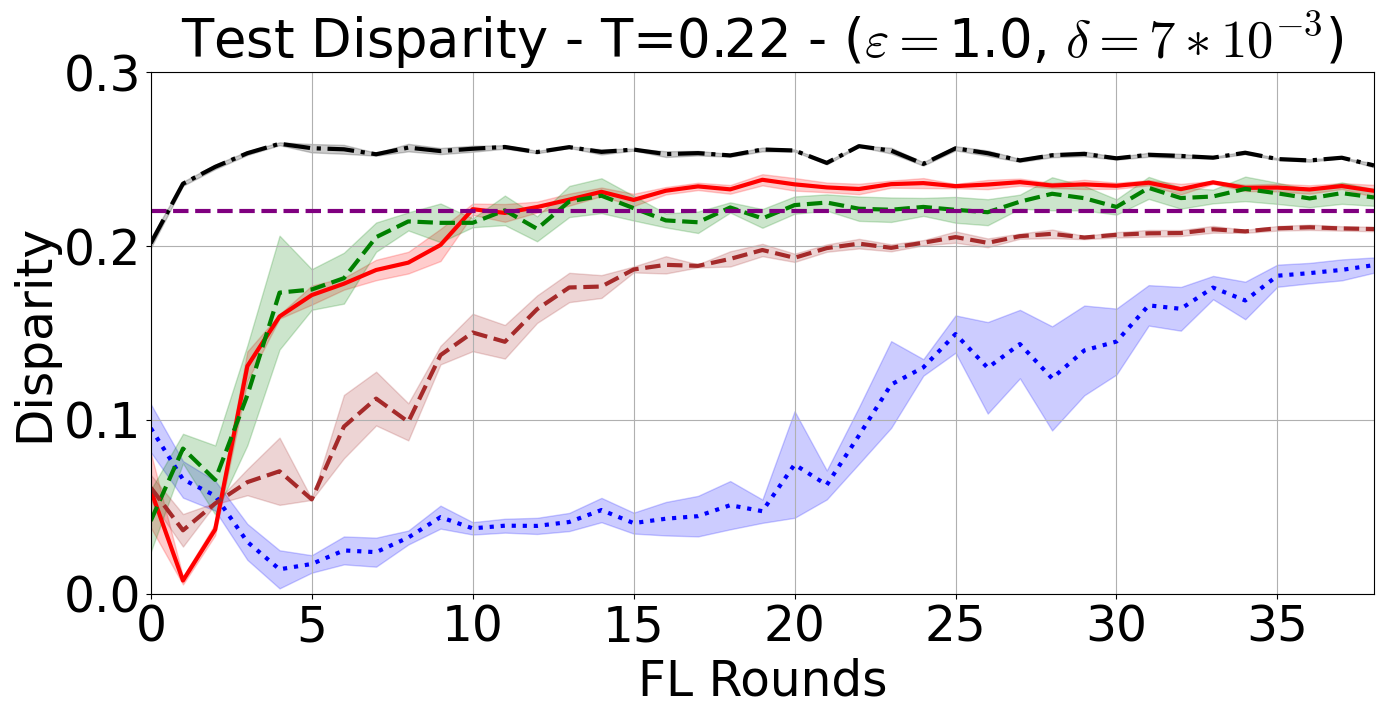}
         \caption{}
         \label{fig:dutch_1:disparity_5}
     \end{subfigure}\\
     \vspace{0.30cm}
     \includegraphics[width=0.60\linewidth]{images/legend_disparity.png}\\
    \vspace{0.15cm}
        \caption{Experiment with Dutch. Fairness parameters $T=0.04$, $T=0.06$, $T=0.09$, $T=0.12$ and $T=0.15$, privacy parameters ($\epsilon=0.5$, $\delta=7 \times  10^{-3}$). Figures ~\ref{fig:dutch_1:accuracy_1}, ~\ref{fig:dutch_1:accuracy_2}, ~\ref{fig:dutch_1:accuracy_3},~\ref{fig:dutch_1:accuracy_4},   and ~\ref{fig:dutch_1:accuracy_5} show the test accuracy of training model while Figures ~\ref{fig:dutch_1:disparity_1}, ~\ref{fig:dutch_1:disparity_2}, ~\ref{fig:dutch_1:disparity_3}, ~\ref{fig:dutch_1:disparity_4},  and ~\ref{fig:dutch_1:disparity_5} show the model disparity.}
        \label{fig:dutch_1}
\end{figure*}

Figures ~\ref{fig:dutch_05} and ~\ref{fig:dutch_1} show the results we obtained testing our proposed approach on the Dutch Dataset. 
We report in Table ~\ref{tab:results_dutch} a comparison of the different use of regularization and DP. In particular, here we compare the following experiments: Baseline, Only DP, Only Fairness, Fairness and DP both with Tunable and Fixed $\lambda$.

\begin{table*}[!t]
    \centering
    \captionsetup{justification=justified}
    \begin{tabular}{|c|c|c|c|c|}\hline
         & \textbf{($\varepsilon$, $\delta$)} & \textbf{$T$}& \textbf{Accuracy} & \textbf{Disparity}\\\hline
        Baseline & - & - & 0.81+-0.0& 0.246+-0.001\\\hline
        
        DP & (0.5, $\mathrm{7*10^{-3}}$) & - & 0.809+-0.001& 0.234+-0.004\\\hline
        DP & (1.0, $\mathrm{7*10^{-3}}$) & - & 0.809+-0.001& 0.234+-0.004\\\hline\hline
        
        Fair & - & 0.06 & 0.74+-0.011& 0.129+-0.003\\\hline
        Fair & - & 0.08 & 0.62+-0.001& 0.054+-0.001\\\hline
        Fair & - & 0.12 & 0.631+-0.009& 0.023+-0.004\\\hline
        Fair & - & 0.18 & 0.756+-0.001& 0.175+-0.001\\\hline
        Fair & - & 0.22 & 0.806+-0.001& 0.21+-0.001\\\hline\hline

        PUFFLE (Fixed $\lambda$) & (0.5, $\mathrm{7*10^{-3}}$) & 0.06 & 0.644+-0.057& 0.05+-0.036\\\hline
        PUFFLE (Tunable $\lambda$) & (0.5, $\mathrm{7*10^{-3}}$) & 0.06 & 0.633+-0.037& 0.047+-0.004\\\hline
        PUFFLE (Fixed $\lambda$) & (1.0, $\mathrm{7*10^{-3}}$) & 0.06 & 0.661+-0.025& 0.058+-0.014\\\hline
        PUFFLE (Tunable $\lambda$) & (1.0, $\mathrm{7*10^{-3}}$) & 0.06 & 0.632+-0.024& 0.059+-0.007\\\hline\hline
        
        PUFFLE (Fixed $\lambda$) & (0.5, $\mathrm{7*10^{-3}}$) & 0.08 & 0.615+-0.011& 0.014+-0.003\\\hline
        PUFFLE (Tunable $\lambda$) & (0.5, $\mathrm{7*10^{-3}}$) & 0.08 & 0.711+-0.035& 0.097+-0.008\\\hline
        PUFFLE (Fixed $\lambda$) & (1.0, $\mathrm{7*10^{-3}}$) & 0.08 & 0.62+-0.038& 0.069+-0.025\\\hline
        PUFFLE (Tunable $\lambda$) & (1.0, $\mathrm{7*10^{-3}}$) & 0.08 & 0.64+-0.032& 0.073+-0.01\\\hline\hline
        
        PUFFLE (Fixed $\lambda$) & (0.5, $\mathrm{7*10^{-3}}$) & 0.12 & 0.714+-0.003& 0.119+-0.015\\\hline
        PUFFLE (Tunable $\lambda$) & (0.5, $\mathrm{7*10^{-3}}$) & 0.12 & 0.703+-0.017& 0.104+-0.017\\\hline
        PUFFLE (Fixed $\lambda$) & (1.0, $\mathrm{7*10^{-3}}$) & 0.12 & 0.703+-0.015& 0.115+-0.015\\\hline
        PUFFLE (Tunable $\lambda$) & (1.0, $\mathrm{7*10^{-3}}$) & 0.12 & 0.632+-0.013& 0.06+-0.012\\\hline\hline
        
        PUFFLE (Fixed $\lambda$) & (0.5, $\mathrm{7*10^{-3}}$) & 0.18 & 0.784+-0.008& 0.19+-0.002\\\hline
        PUFFLE (Tunable $\lambda$) & (0.5, $\mathrm{7*10^{-3}}$) & 0.18 & 0.764+-0.018& 0.17+-0.016\\\hline
        PUFFLE (Fixed $\lambda$) & (1.0, $\mathrm{7*10^{-3}}$) & 0.18 & 0.728+-0.011& 0.125+-0.017\\\hline
        PUFFLE (Tunable $\lambda$) & (1.0, $\mathrm{7*10^{-3}}$) & 0.18 & 0.742+-0.003& 0.172+-0.002\\\hline\hline
        
        PUFFLE (Fixed $\lambda$) & (0.5, $\mathrm{7*10^{-3}}$) & 0.22 & 0.771+-0.007& 0.185+-0.011\\\hline
        PUFFLE (Tunable $\lambda$) & (0.5, $\mathrm{7*10^{-3}}$) & 0.22 & 0.756+-0.013& 0.179+-0.017\\\hline
        PUFFLE (Fixed $\lambda$) & (1.0, $\mathrm{7*10^{-3}}$) & 0.22 & 0.809+-0.0& 0.228+-0.005\\\hline
        PUFFLE (Tunable $\lambda$) & (1.0, $\mathrm{7*10^{-3}}$) & 0.22 & 0.768+-0.005& 0.189+-0.004\\\hline
    
    \end{tabular}
    \caption{Dutch dataset: A comparison of the final test accuracy and the final test disparity across all the possible combinations. Specifically, the analysis encompasses two possible privacy parameters, namely ($\epsilon=5$, $8 \times  10^{-4}$) and ($\epsilon=8$, $8 \times  10^{-4}$) alongside three target fairness disparity, denoted as $T=0.06$, $T=0.08$, $T=0.12$, $T=0.18$ and $T=0.22$  corresponding to a reduction of 10\%, 25\%, 50\%, 65\%, 75\% with respect to the Baseline disparity.}
    \label{tab:results_dutch}
\end{table*}

The results that in terms of unfairness mitigation are coherent with the ones that we have shown with the CelebA Dataset. Finding a trade-off between the different trustworthiness requirements is more difficult with this tabular dataset. 
The reduction in accuracy caused by the use of regularization and DP is more evident w.r.t. the one that we have noticed with CelebA. This degradation of the accuracy increases as we reduce the target disparity $T$ as shown in Figures ~\ref{fig:dutch_05:accuracy_1}, ~\ref{fig:dutch_05:accuracy_2}, ~\ref{fig:dutch_05:accuracy_3},~\ref{fig:dutch_05:accuracy_4}, ~\ref{fig:dutch_05:accuracy_5}, ~\ref{fig:dutch_1:accuracy_1}, ~\ref{fig:dutch_1:accuracy_2}, ~\ref{fig:dutch_1:accuracy_3}, ~\ref{fig:dutch_1:accuracy_4}  and ~\ref{fig:dutch_1:accuracy_5}.

\begin{figure*}
     \centering
     \captionsetup{justification=justified}
     \begin{subfigure}[b]{0.18\textwidth}
         \centering
         \includegraphics[width=\textwidth]{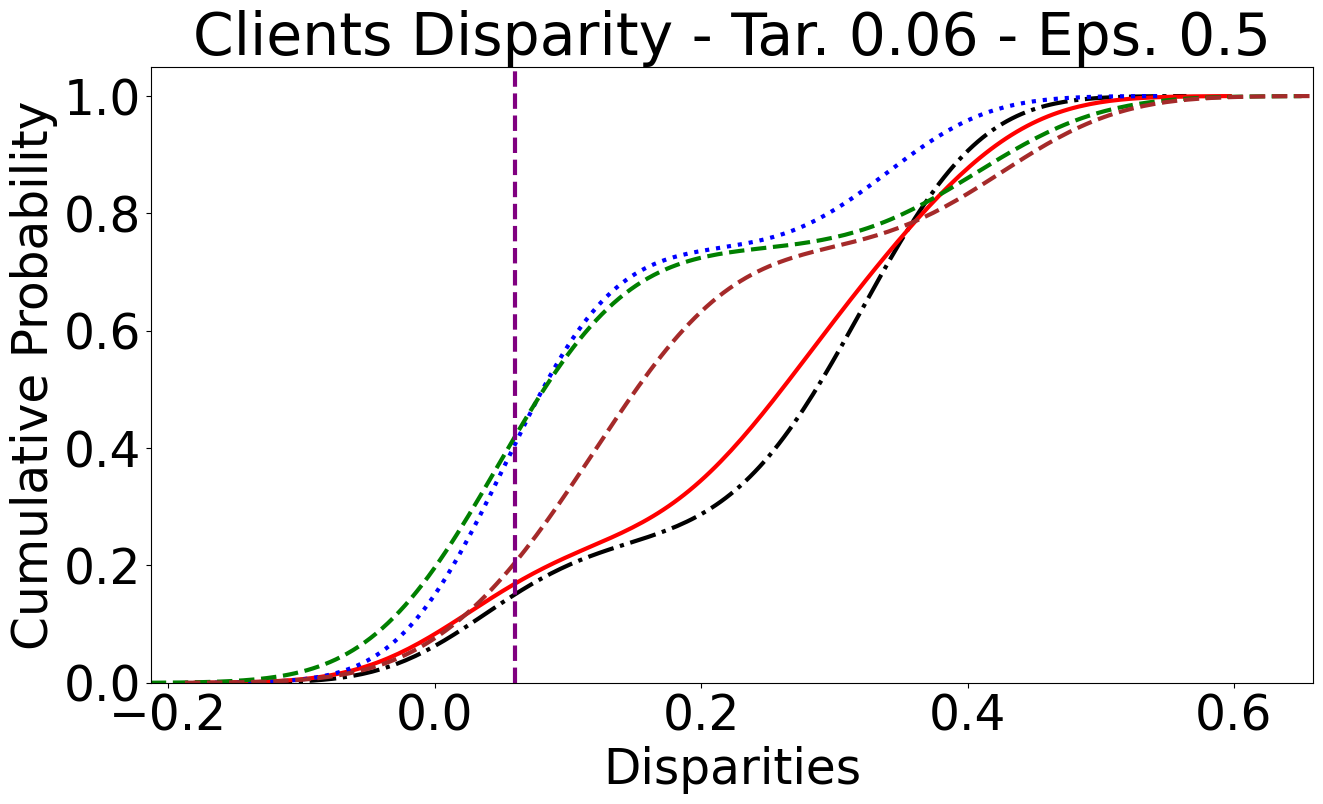}
         \caption{}
         \label{fig:dutch_local_05:cdf_1}
     \end{subfigure}
     \hfill
     \begin{subfigure}[b]{0.18\textwidth}
         \centering
         \includegraphics[width=\textwidth]{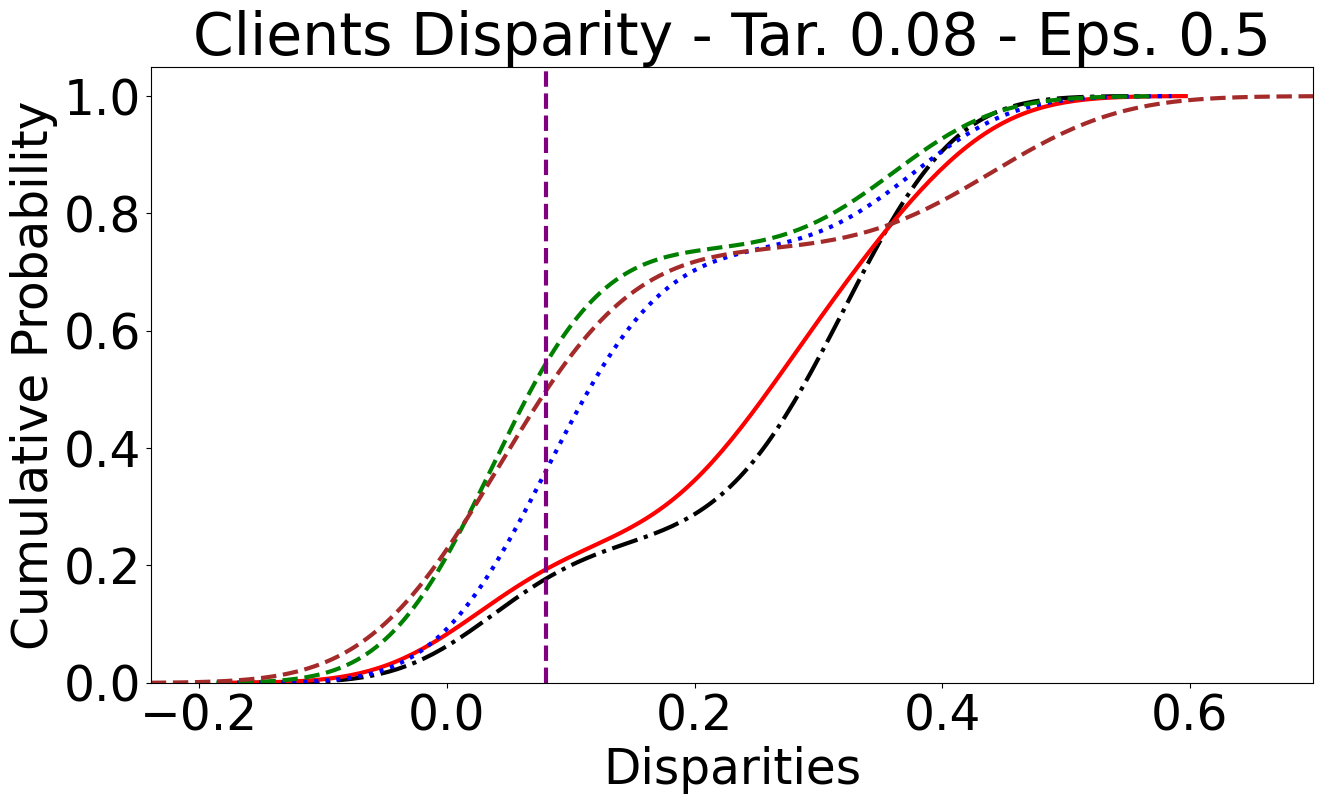}
         \caption{}
         \label{fig:dutch_local_05:cdf_2}
     \end{subfigure}
     \hfill
     \begin{subfigure}[b]{0.18\textwidth}
         \centering
         \includegraphics[width=\textwidth]{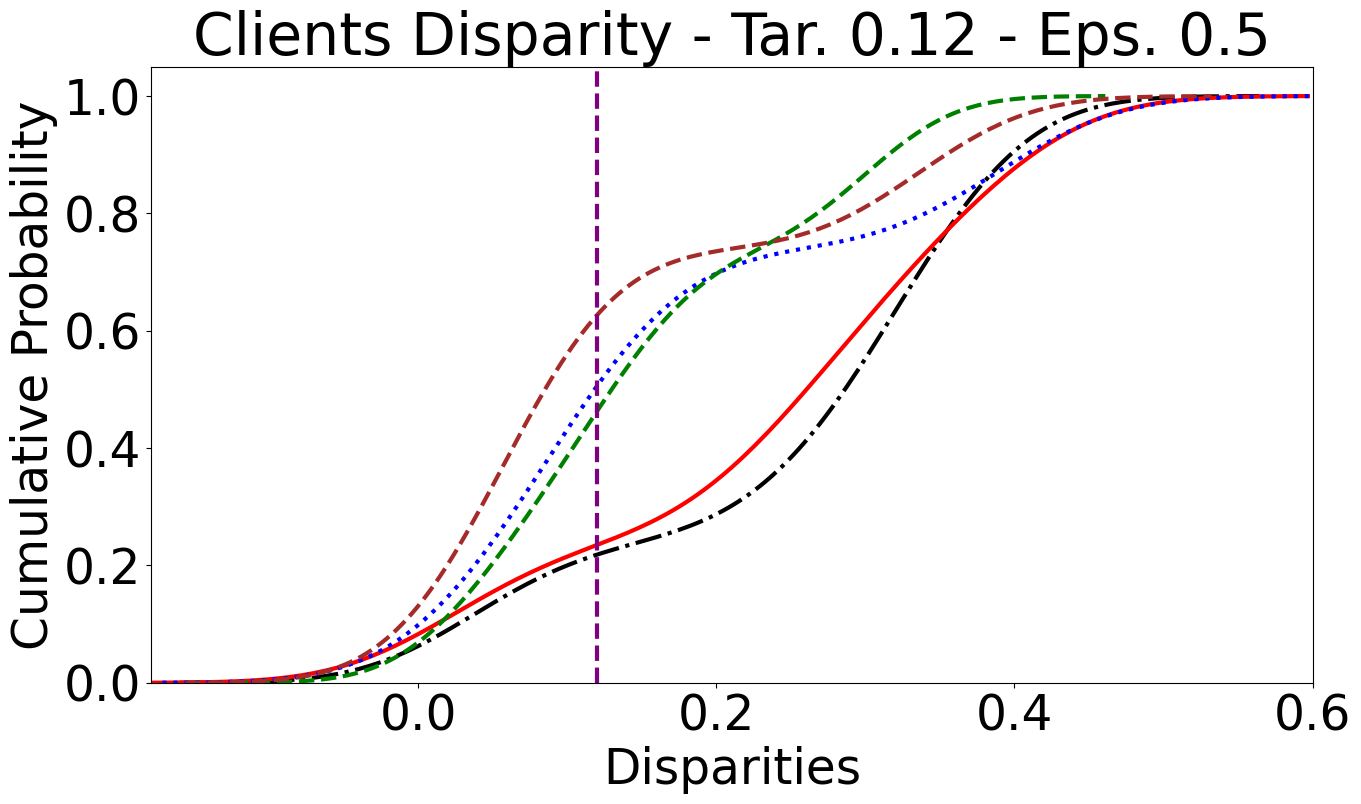}
         \caption{}
         \label{fig:dutch_local_05:cdf_3}
     \end{subfigure}
     \hfill
     \begin{subfigure}[b]{0.18\textwidth}
         \centering
         \includegraphics[width=\textwidth]{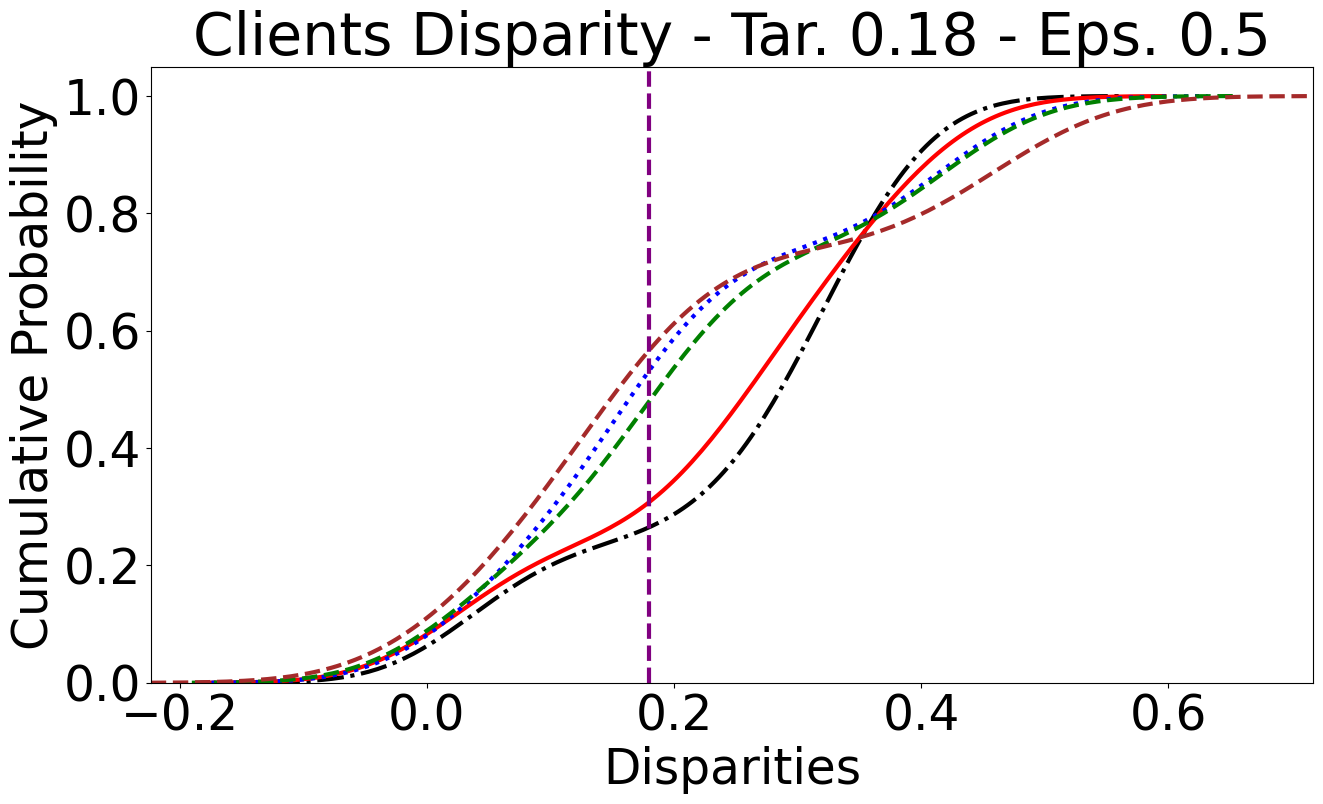}
         \caption{}
         \label{fig:dutch_local_05:cdf_4}
     \end{subfigure}
     \hfill
     \begin{subfigure}[b]{0.18\textwidth}
         \centering
         \includegraphics[width=\textwidth]{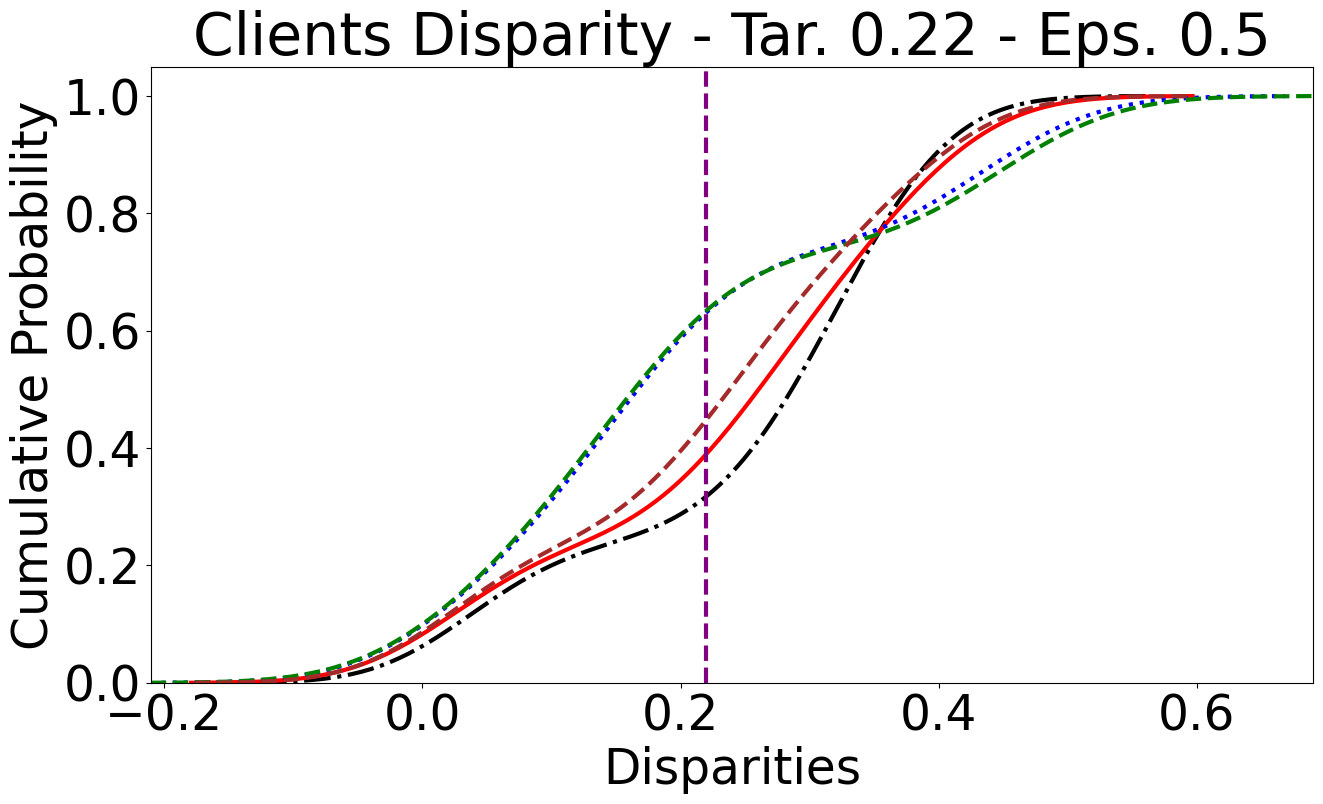}
         \caption{}
         \label{fig:dutch_local_05:cdf_5}
     \end{subfigure}\\
    \vspace{0.30cm}
     \includegraphics[width=0.60\linewidth]{images/legend_disparity.png}\\
    \vspace{0.15cm}
        \caption{Experiment with Dutch. Fairness parameters $T=0.04$, $T=0.06$, $T=0.09$, $T=0.12$ and $T=0.15$, privacy parameters ($\epsilon=0.5$, $\delta=7 \times  10^{-3}$). The Figures show the cumulative distribution function of the local disparities of the clients.}
        \label{fig:dutch_local_05}
\end{figure*}

\begin{figure*}
     \centering
     \captionsetup{justification=justified}
     \begin{subfigure}[b]{0.18\textwidth}
         \centering
         \includegraphics[width=\textwidth]{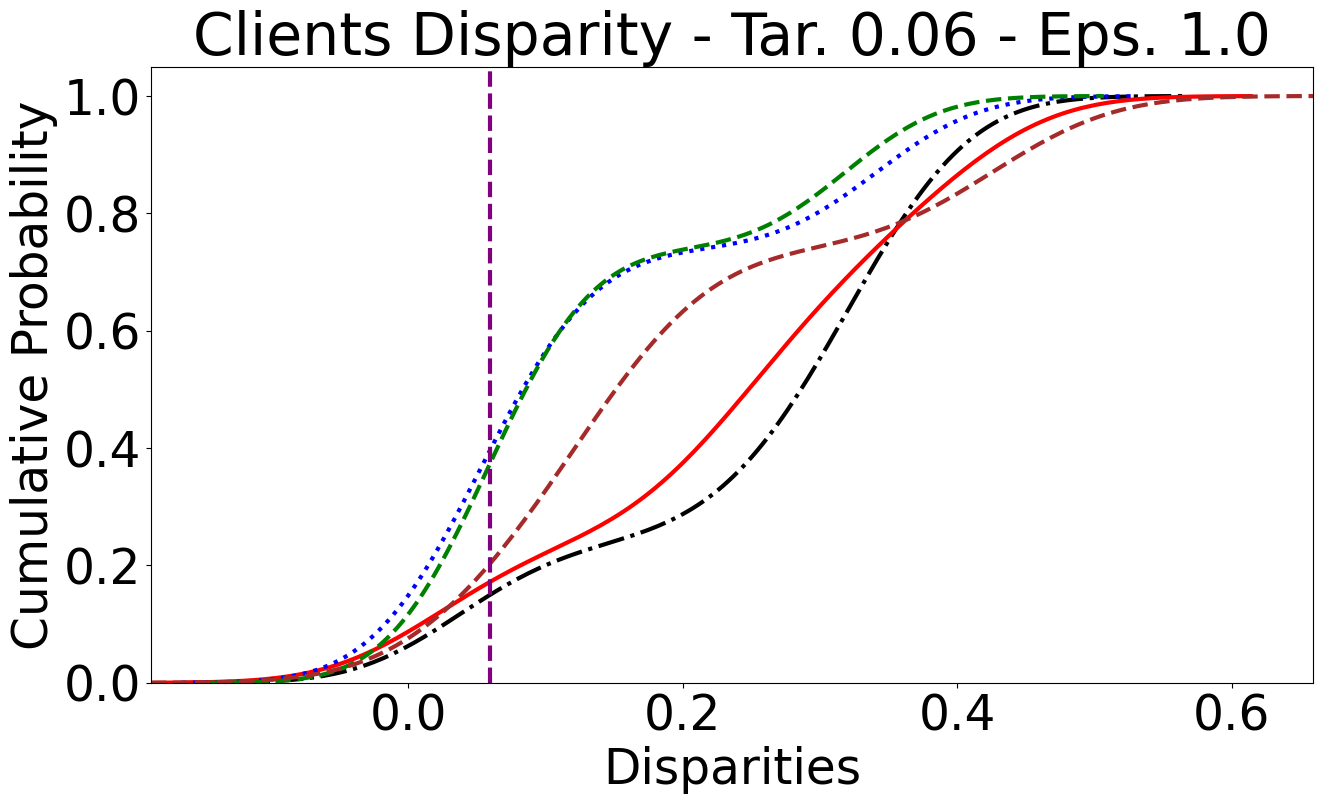}
         \caption{}
         \label{fig:dutch_local_1:cdf_1}
     \end{subfigure}
     \hfill
     \begin{subfigure}[b]{0.18\textwidth}
         \centering
         \includegraphics[width=\textwidth]{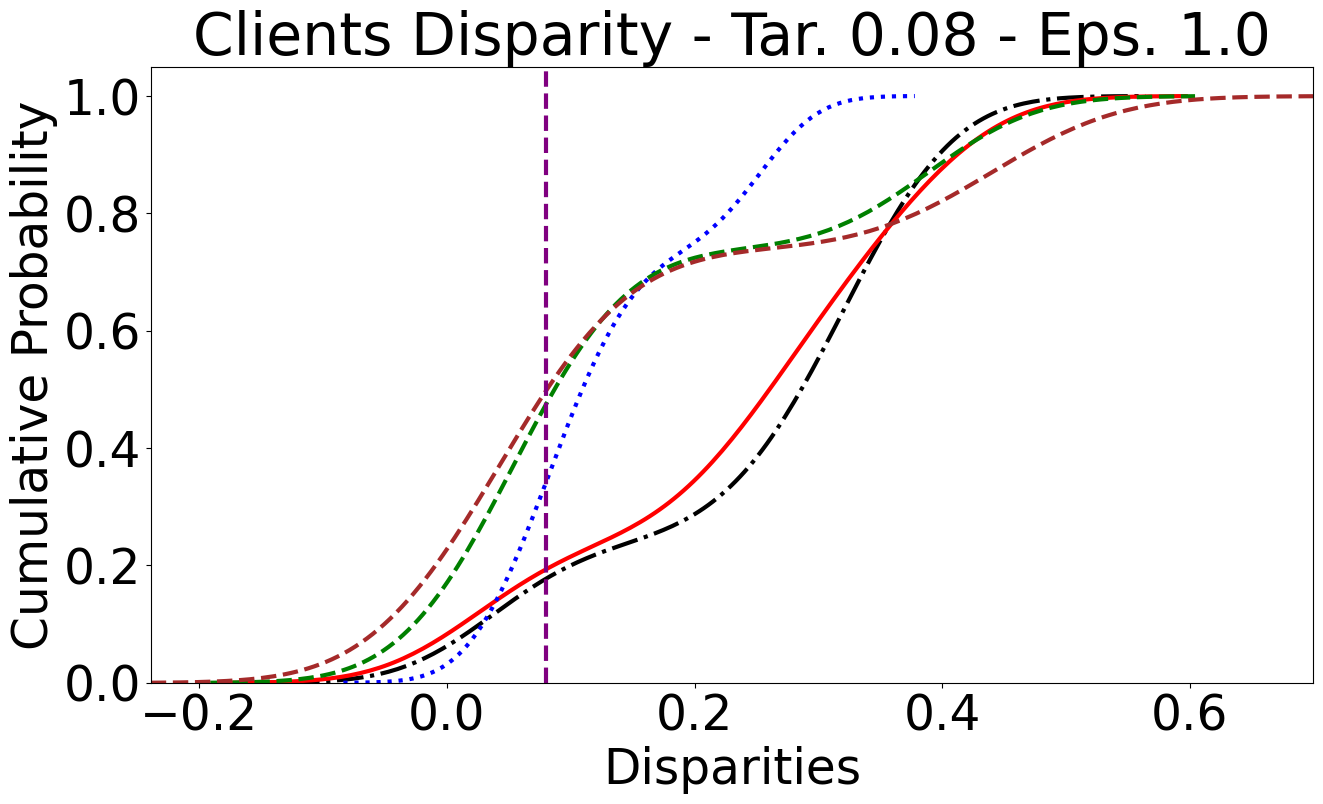}
         \caption{}
         \label{fig:dutch_local_1:cdf_2}
     \end{subfigure}
     \hfill
     \begin{subfigure}[b]{0.18\textwidth}
         \centering
         \includegraphics[width=\textwidth]{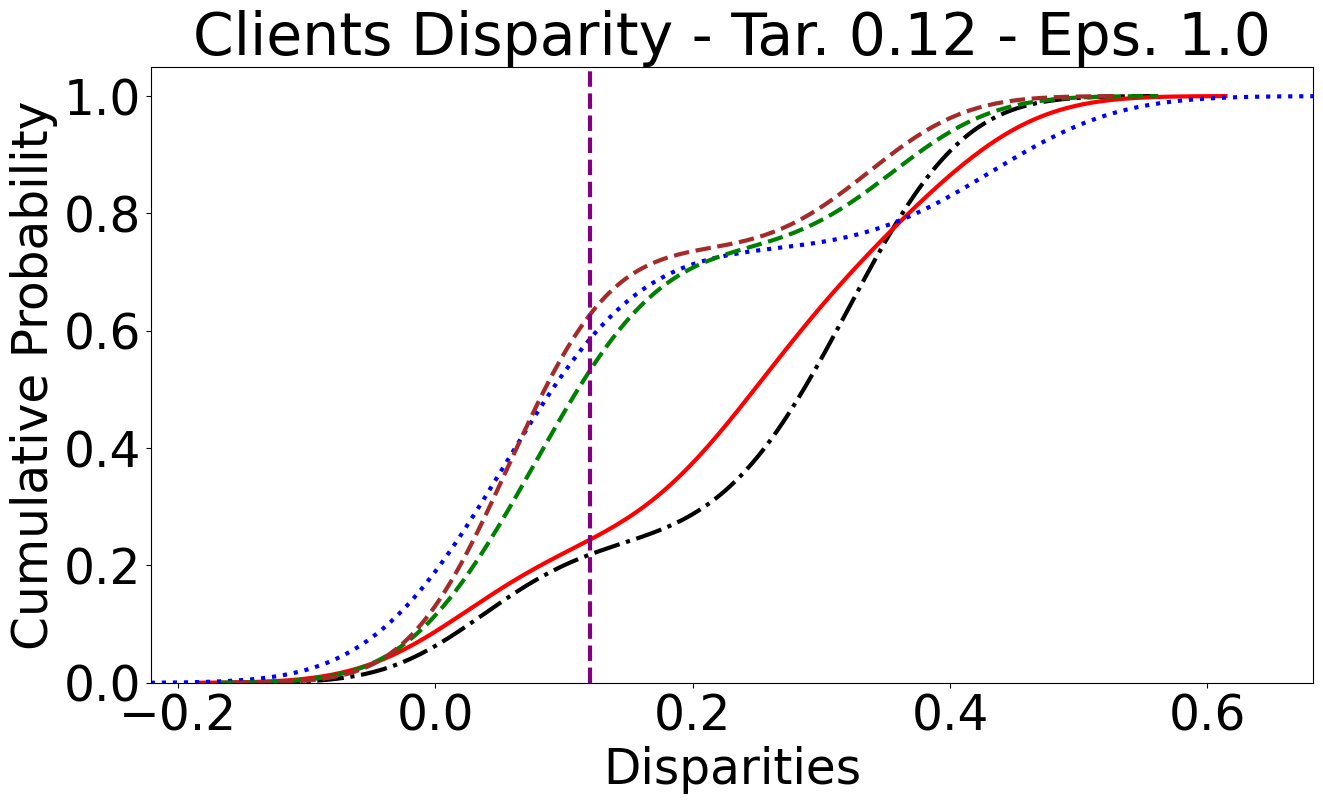}
         \caption{}
         \label{fig:dutch_local_1:cdf_3}
     \end{subfigure}
     \hfill
     \begin{subfigure}[b]{0.18\textwidth}
         \centering
         \includegraphics[width=\textwidth]{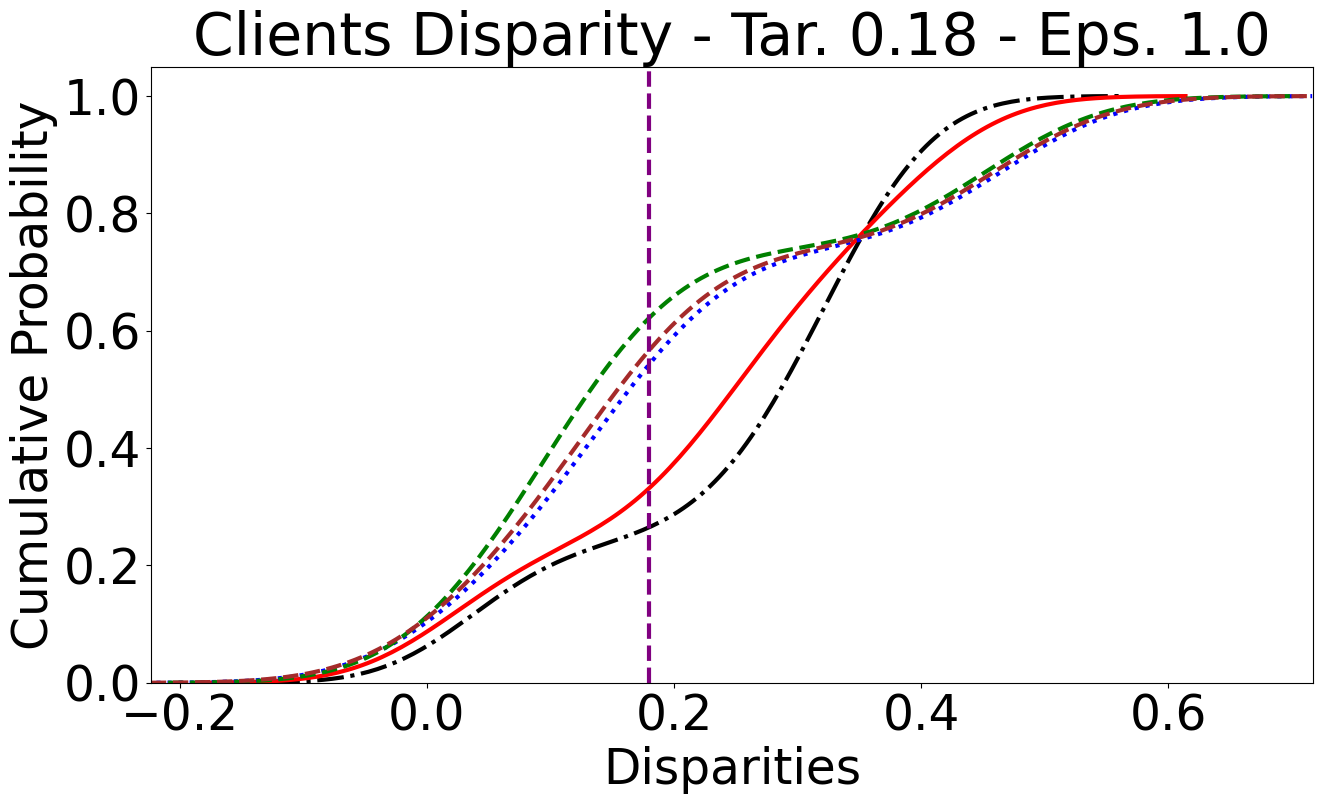}
         \caption{}
         \label{fig:dutch_local_1:cdf_4}
     \end{subfigure}
     \hfill
     \begin{subfigure}[b]{0.18\textwidth}
         \centering
         \includegraphics[width=\textwidth]{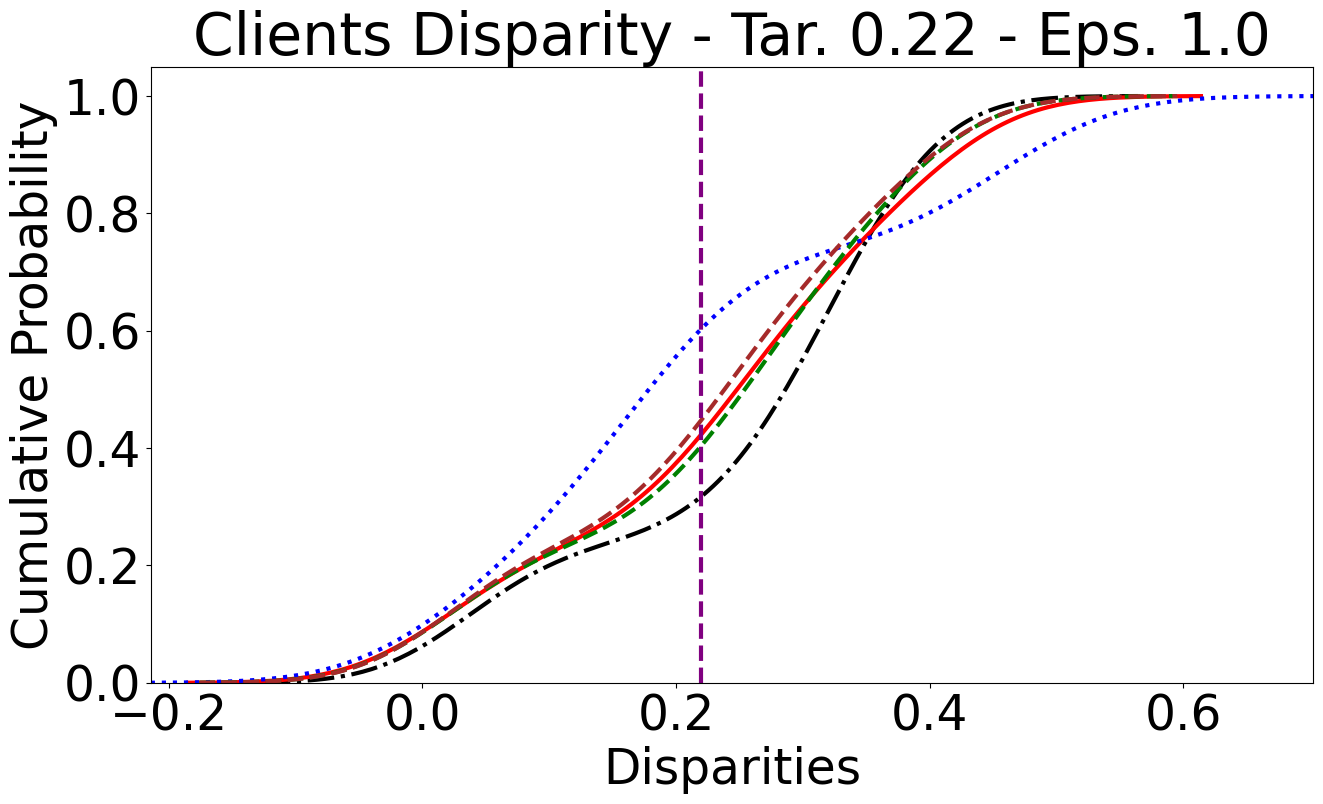}
         \caption{}
         \label{fig:dutch_local_1:cdf_5}
     \end{subfigure}\\
    \vspace{0.30cm}
     \includegraphics[width=0.60\linewidth]{images/legend_disparity.png}\\
        \vspace{0.15cm}
        \caption{Experiment with Dutch. Fairness parameters $T=0.04$, $T=0.06$, $T=0.09$, $T=0.12$ and $T=0.15$, privacy parameters ($\epsilon=0.5$, $\delta=7 \times  10^{-3}$). The Figures show the cumulative distribution function of the local disparities of the clients.}
        \label{fig:dutch_local_1}
\end{figure*}

As we did with CelebA, we also report the plots of the CDF in Figures ~\ref{fig:dutch_local_05} and ~\ref{fig:dutch_local_1}.

\subsection{ACS Income}

\begin{figure*}
     \centering
     \captionsetup{justification=justified}
     \begin{subfigure}[b]{0.18\textwidth}
         \centering
         \includegraphics[width=\textwidth]{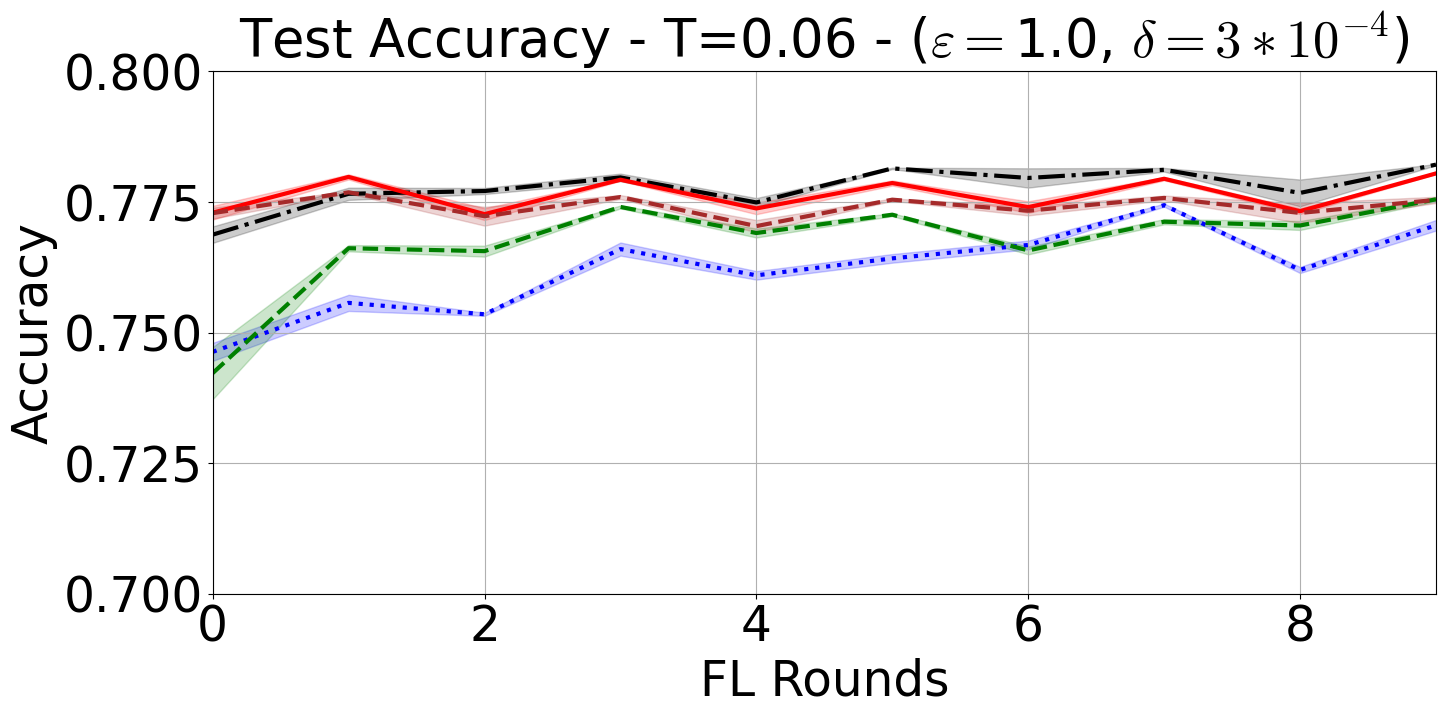}
         \caption{}
         \label{fig:income_1:accuracy_1}
     \end{subfigure}
     \hfill
     \begin{subfigure}[b]{0.18\textwidth}
         \centering
         \includegraphics[width=\textwidth]{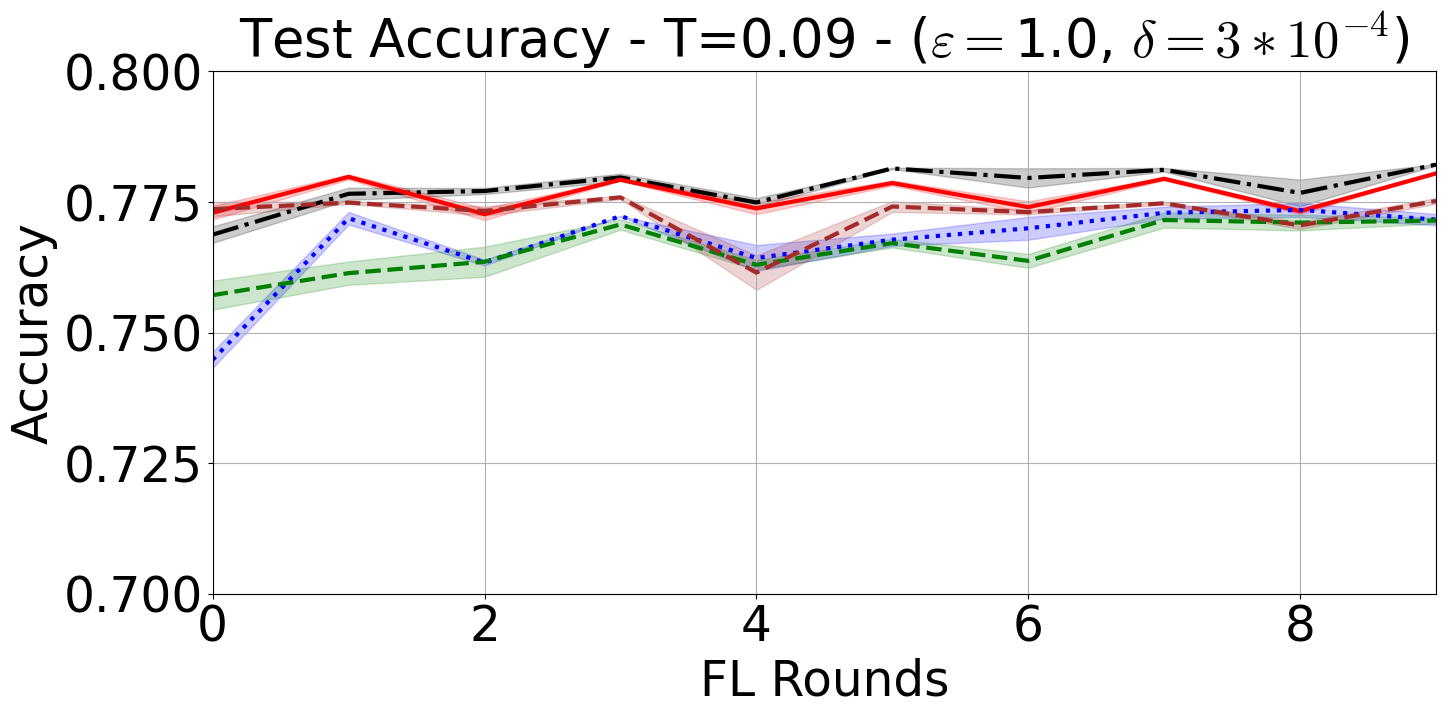}
         \caption{}
         \label{fig:income_1:accuracy_2}
     \end{subfigure}
     \hfill
     \begin{subfigure}[b]{0.18\textwidth}
         \centering
         \includegraphics[width=\textwidth]{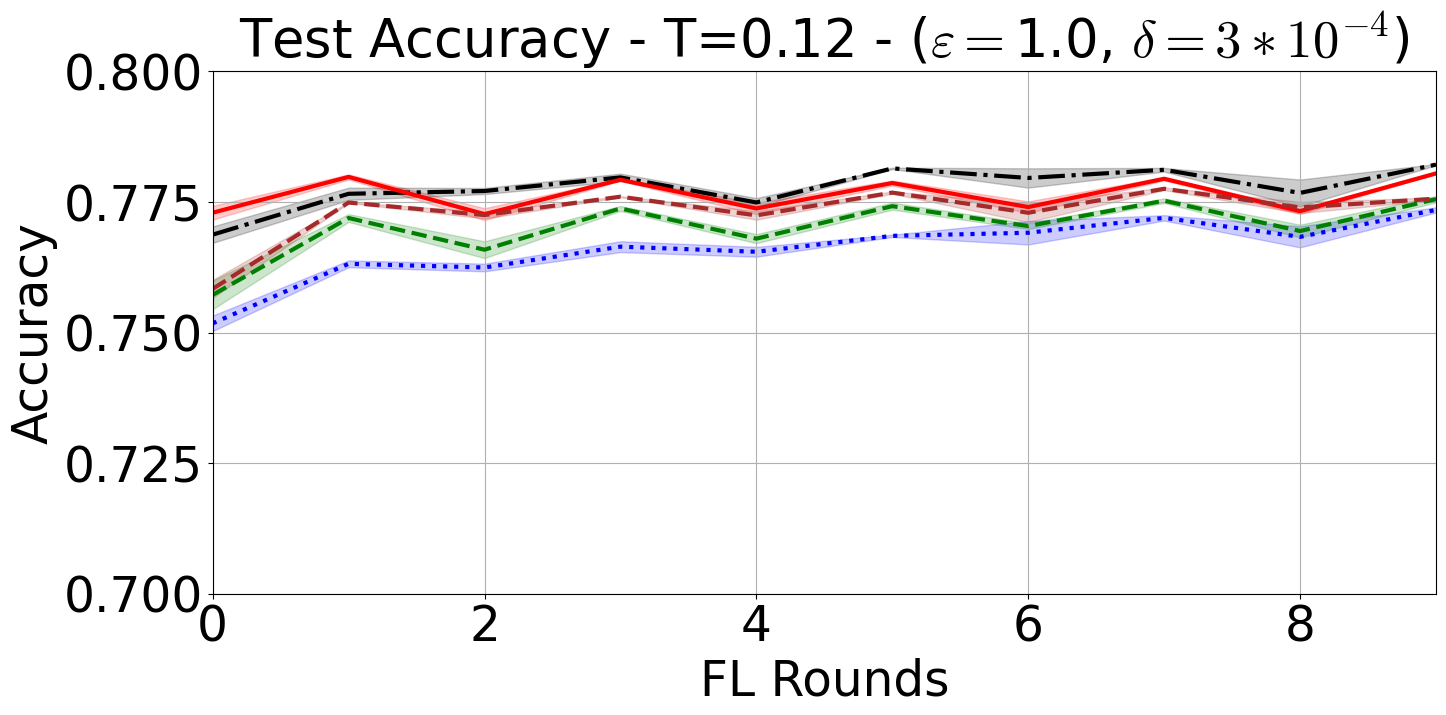}
         \caption{}
         \label{fig:income_1:accuracy_3}
     \end{subfigure}
     \hfill
     \begin{subfigure}[b]{0.18\textwidth}
         \centering
         \includegraphics[width=\textwidth]{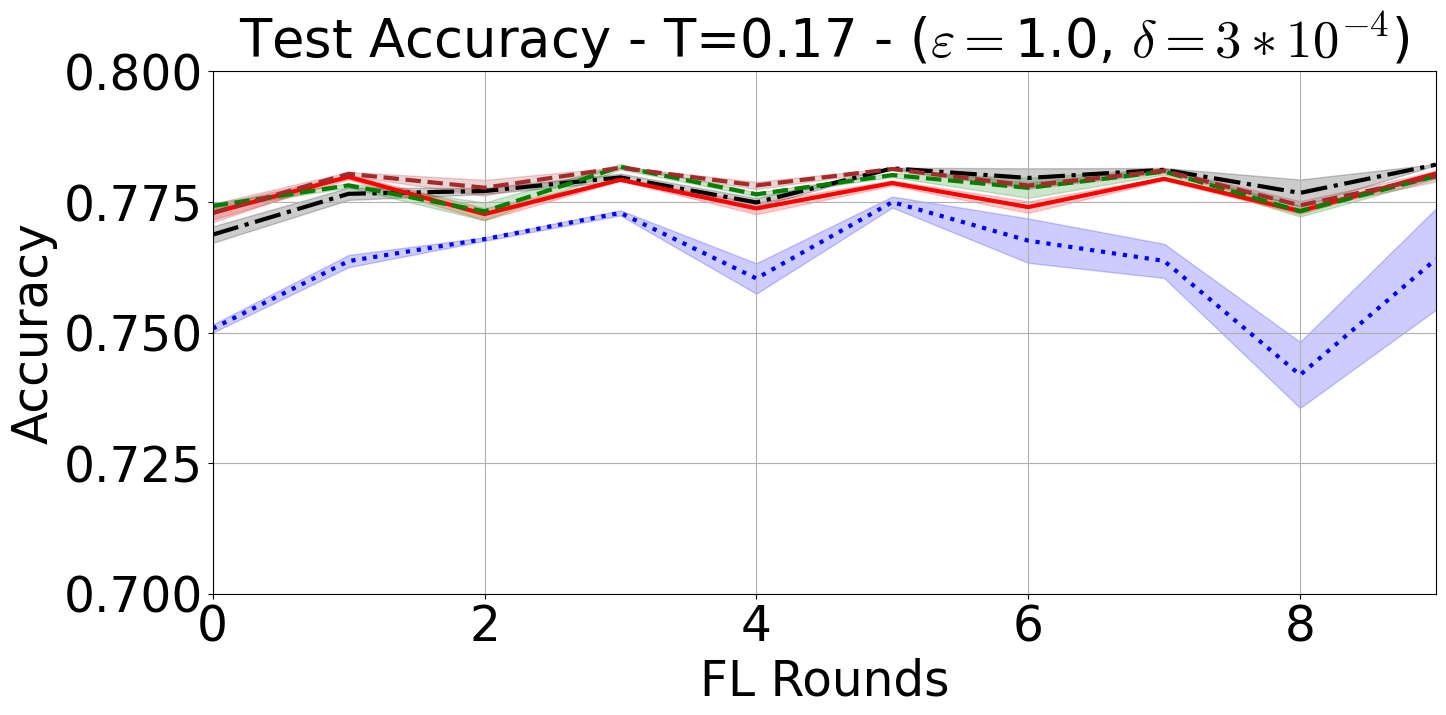}
         \caption{}
         \label{fig:income_1:accuracy_4}
     \end{subfigure}
     \hfill
     \begin{subfigure}[b]{0.18\textwidth}
         \centering
         \includegraphics[width=\textwidth]{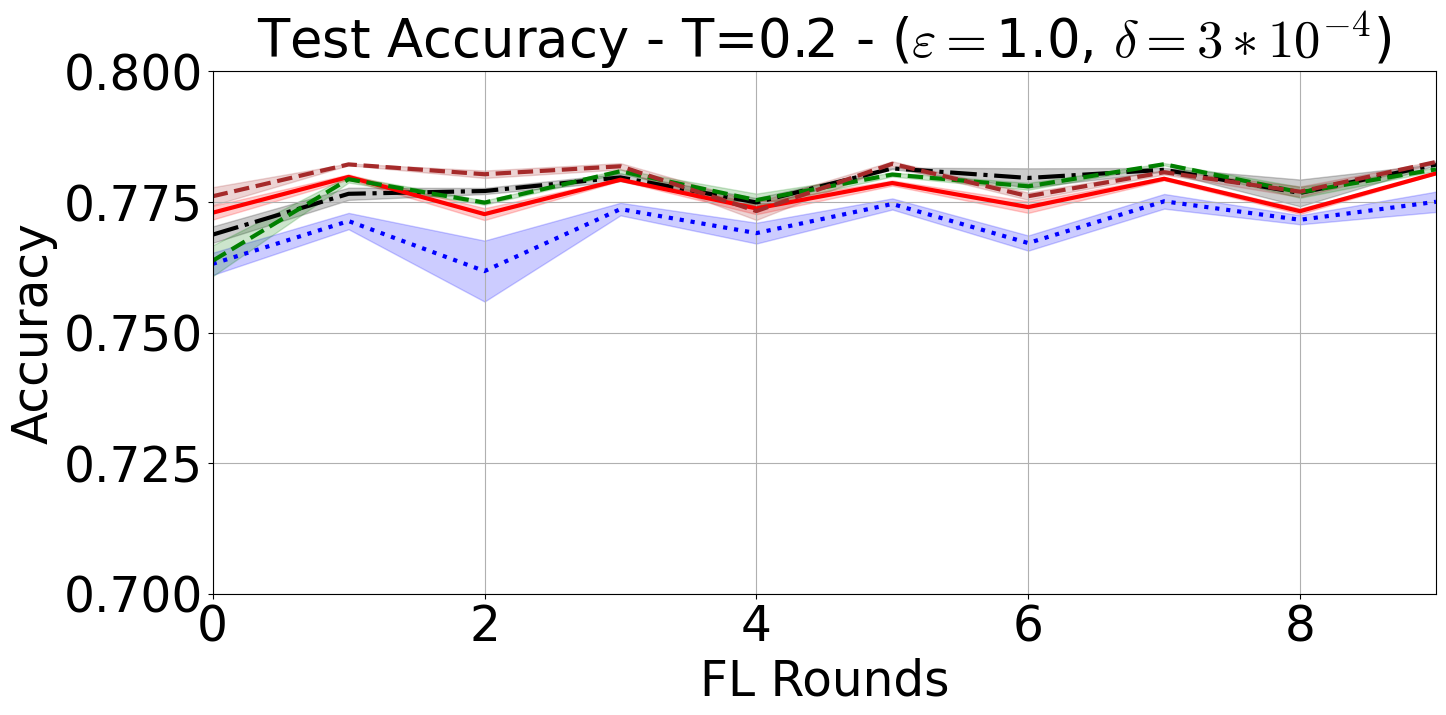}
         \caption{}
         \label{fig:income_1:accuracy_5}
     \end{subfigure}\\
     \vspace{0.30cm}

    \begin{subfigure}[b]{0.18\textwidth}
         \centering
         \includegraphics[width=\textwidth]{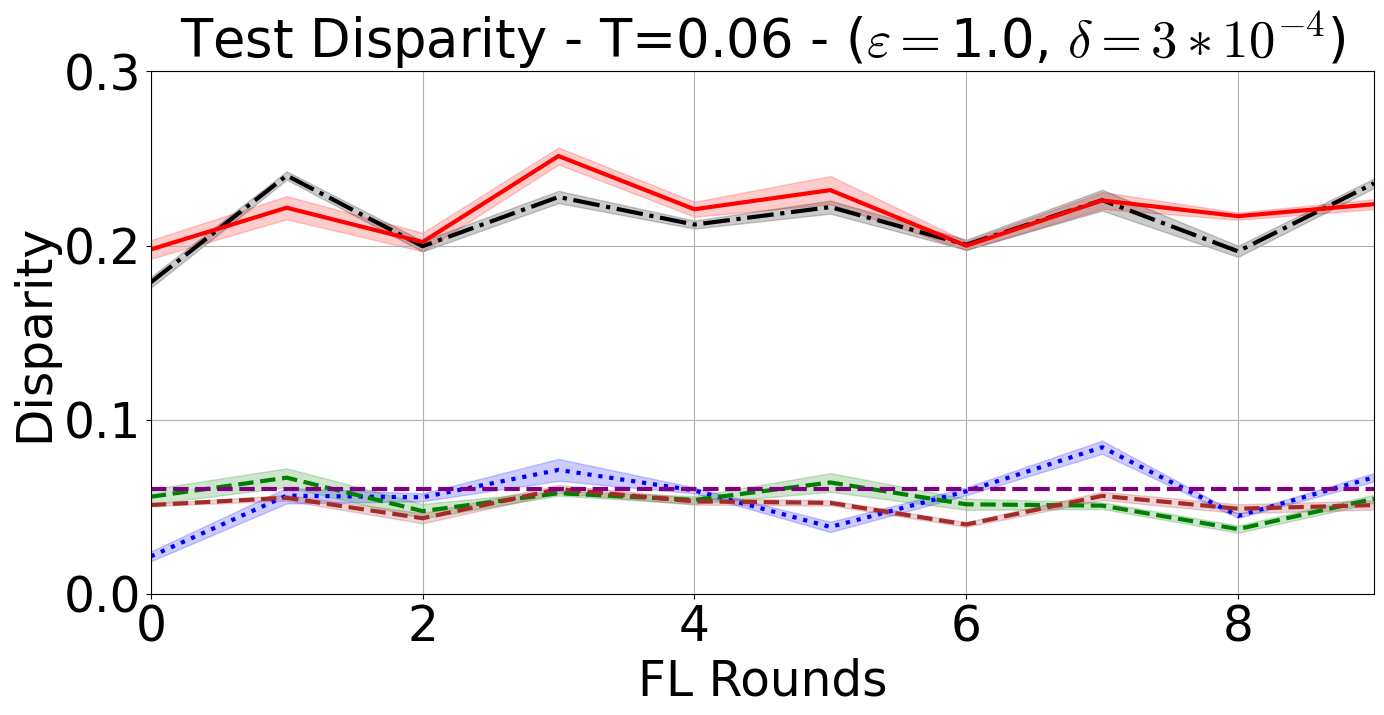}
         \caption{}
         \label{fig:income_1:disparity_1}
     \end{subfigure}
     \hfill
     \begin{subfigure}[b]{0.18\textwidth}
         \centering
         \includegraphics[width=\textwidth]{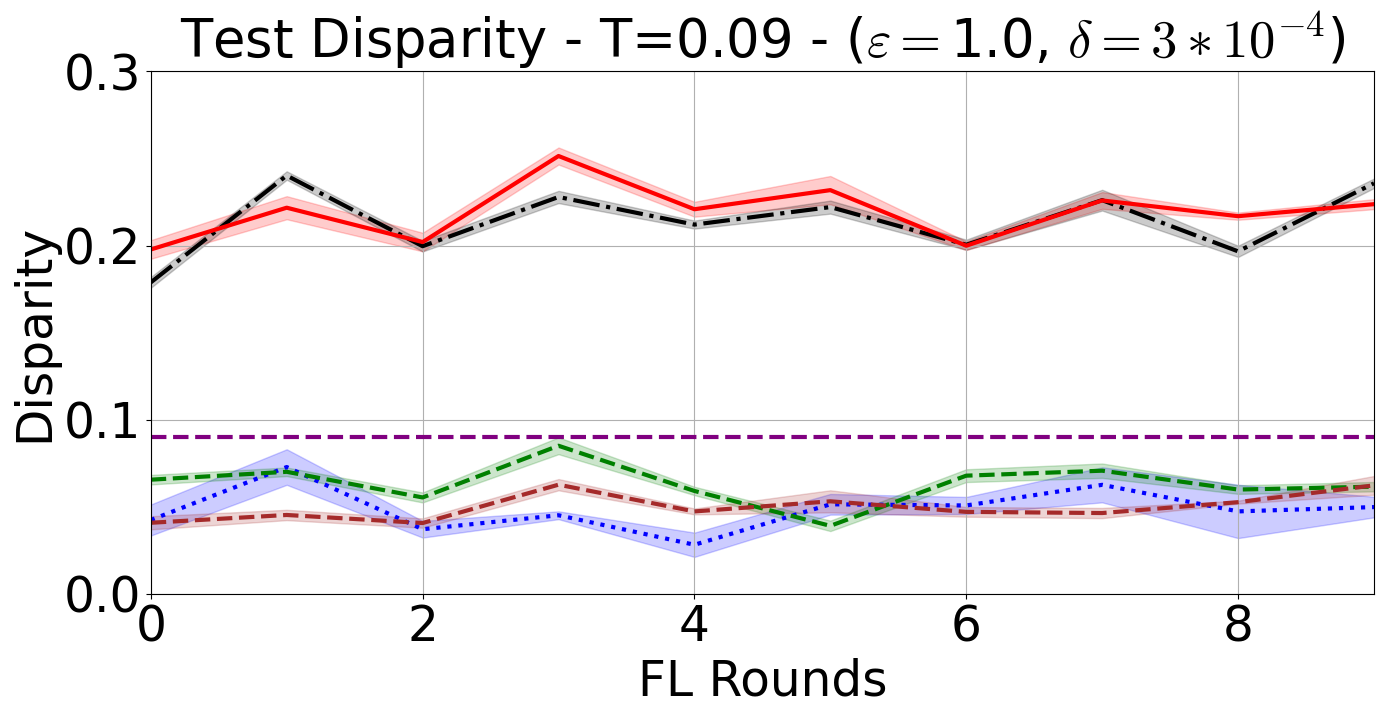}
         \caption{}
         \label{fig:income_1:disparity_2}
     \end{subfigure}
     \hfill
     \begin{subfigure}[b]{0.18\textwidth}
         \centering
         \includegraphics[width=\textwidth]{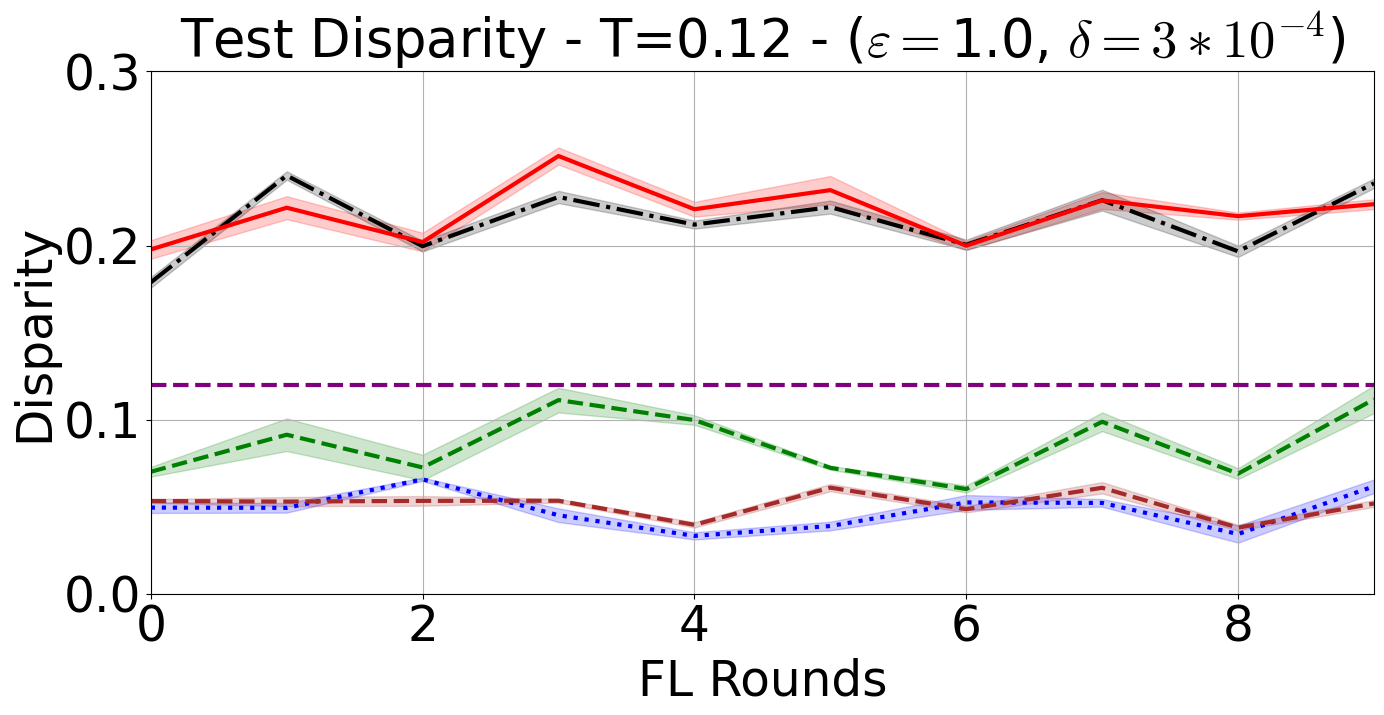}
         \caption{}
         \label{fig:income_1:disparity_3}
     \end{subfigure}
     \hfill
     \begin{subfigure}[b]{0.18\textwidth}
         \centering
         \includegraphics[width=\textwidth]{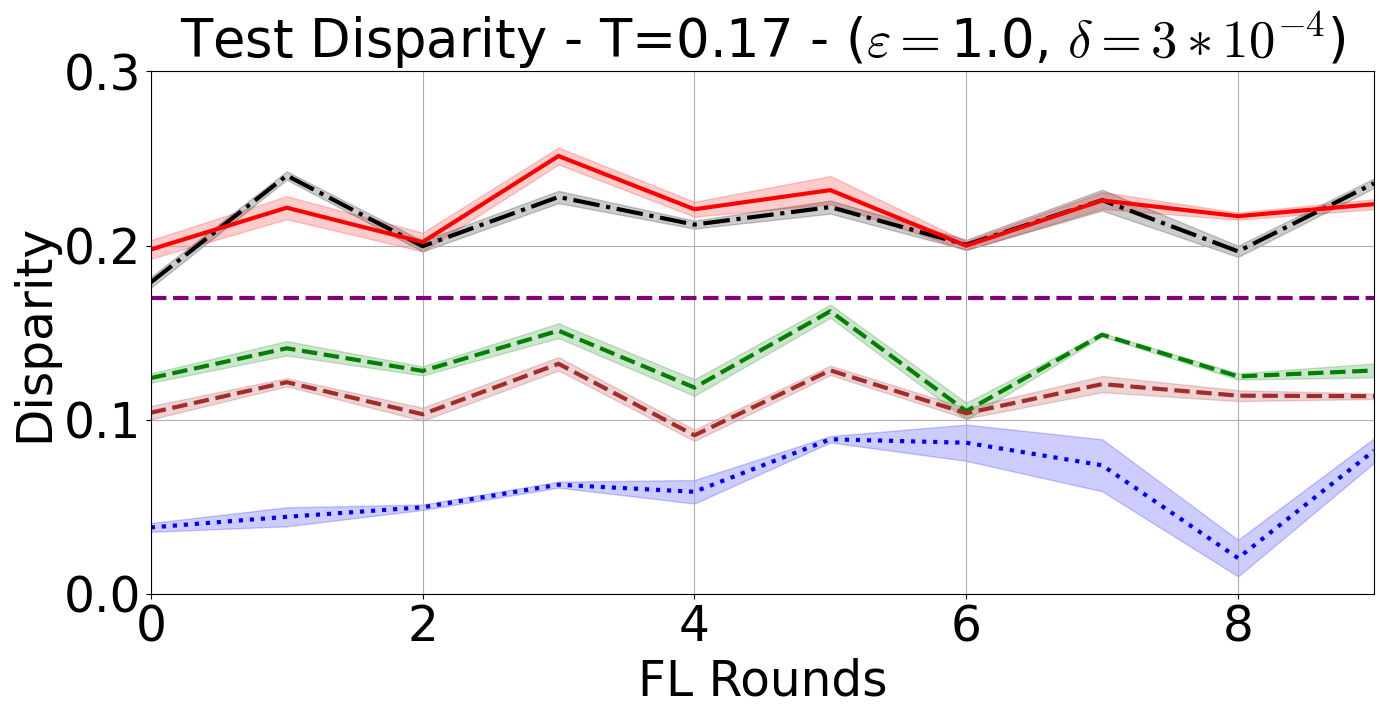}
         \caption{}
         \label{fig:income_1:disparity_4}
     \end{subfigure}
     \hfill
     \begin{subfigure}[b]{0.18\textwidth}
         \centering
         \includegraphics[width=\textwidth]{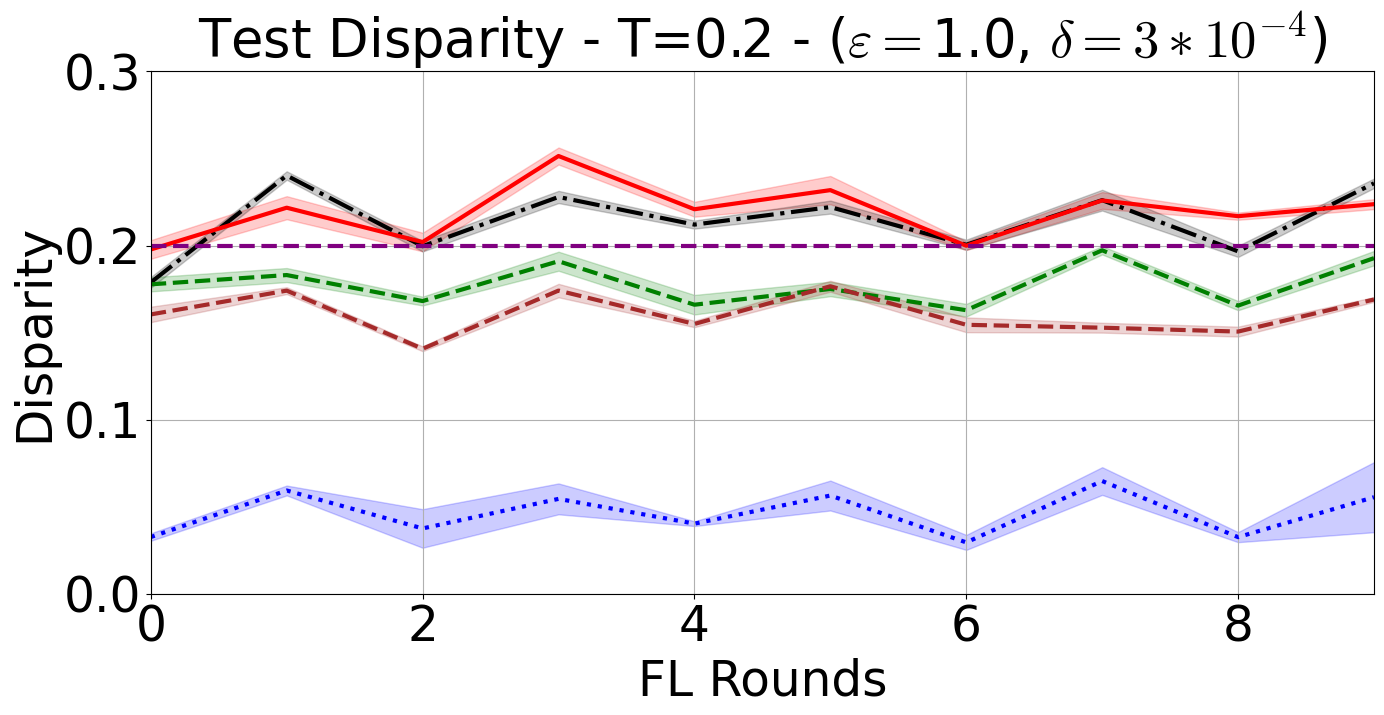}
         \caption{}
         \label{fig:income_1:disparity_5}
     \end{subfigure}\\
     \vspace{0.3cm}
     \includegraphics[width=0.60\linewidth]{images/legend_disparity.png}\\
        \vspace{0.15cm}
        \caption{Experiment with Income. Fairness parameters $T=0.06$, $T=0.08$, $T=0.12$, $T=0.18$ and $T=0.22$, privacy parameters ($\epsilon=0.5$, $\delta=7 \times  10^{-3}$). Figures ~\ref{fig:income_1:accuracy_1}, ~\ref{fig:income_1:accuracy_2}, ~\ref{fig:income_1:accuracy_3}, ~\ref{fig:income_1:accuracy_4} and ~\ref{fig:income_1:accuracy_5} show the test accuracy of training model while Figures ~\ref{fig:dutch_05:disparity_1}, ~\ref{fig:income_1:disparity_2}, ~\ref{fig:income_1:disparity_3}, ~\ref{fig:income_1:disparity_4},  and ~\ref{fig:income_1:disparity_5} show the model disparity.}
        \label{fig:income_1}
\end{figure*}

\begin{figure*}
     \centering
     \captionsetup{justification=justified}
     \begin{subfigure}[b]{0.18\textwidth}
         \centering
         \includegraphics[width=\textwidth]{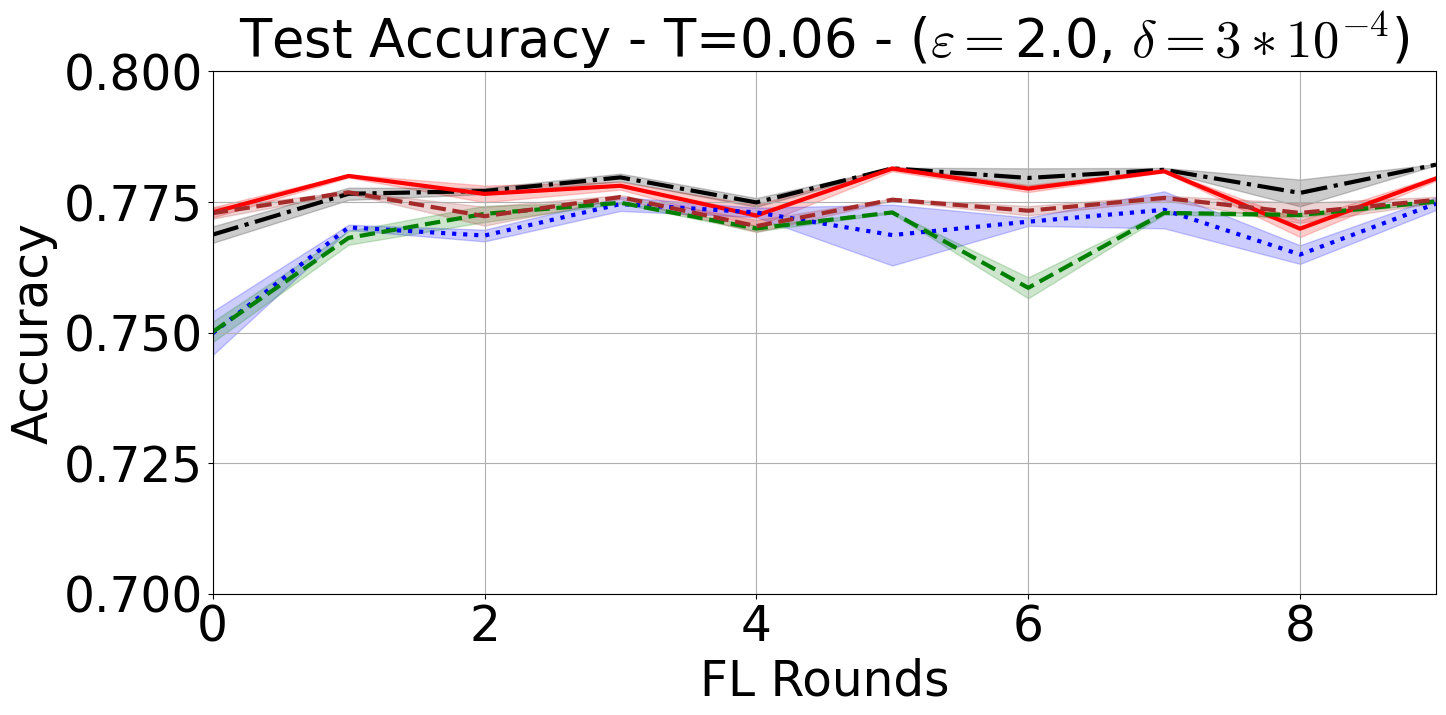}
         \caption{}
         \label{fig:income_2:accuracy_1}
     \end{subfigure}
     \hfill
     \begin{subfigure}[b]{0.18\textwidth}
         \centering
         \includegraphics[width=\textwidth]{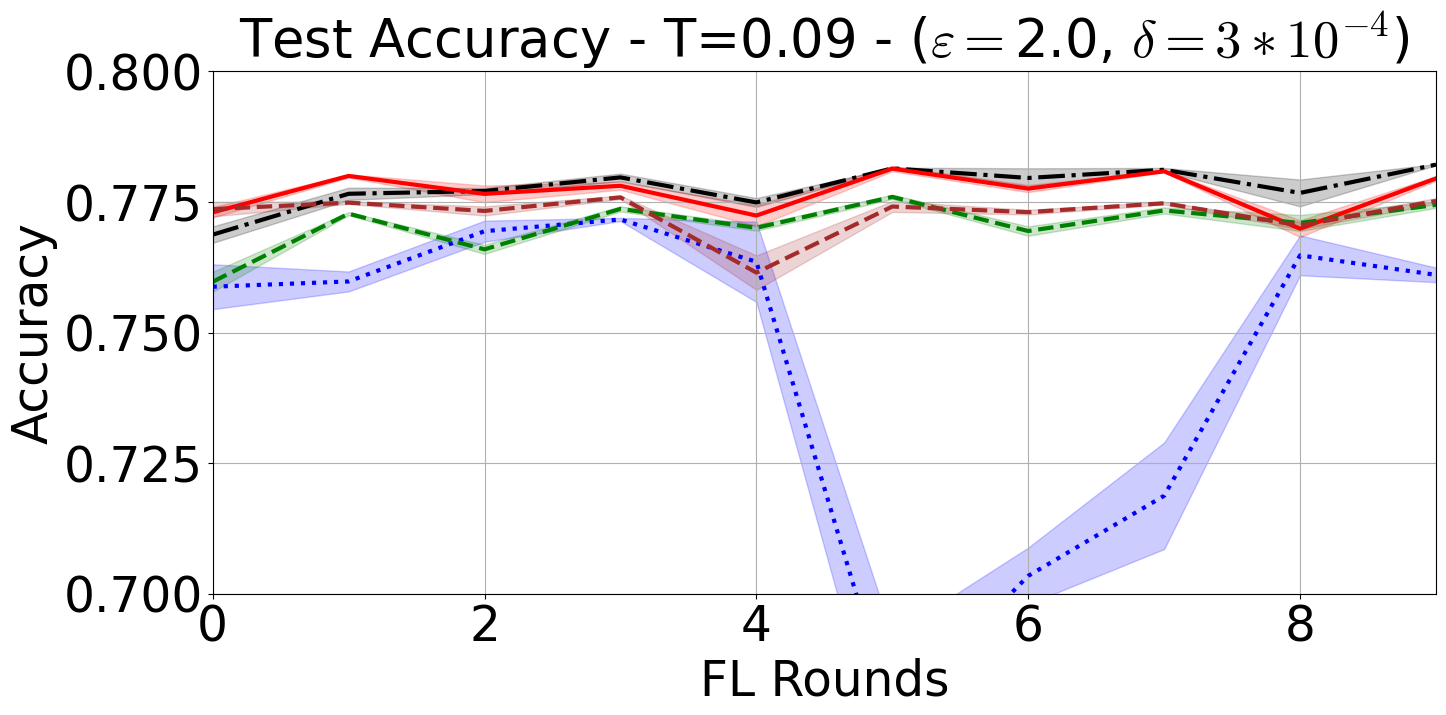}
         \caption{}
         \label{fig:income_2:accuracy_2}
     \end{subfigure}
     \hfill
     \begin{subfigure}[b]{0.18\textwidth}
         \centering
         \includegraphics[width=\textwidth]{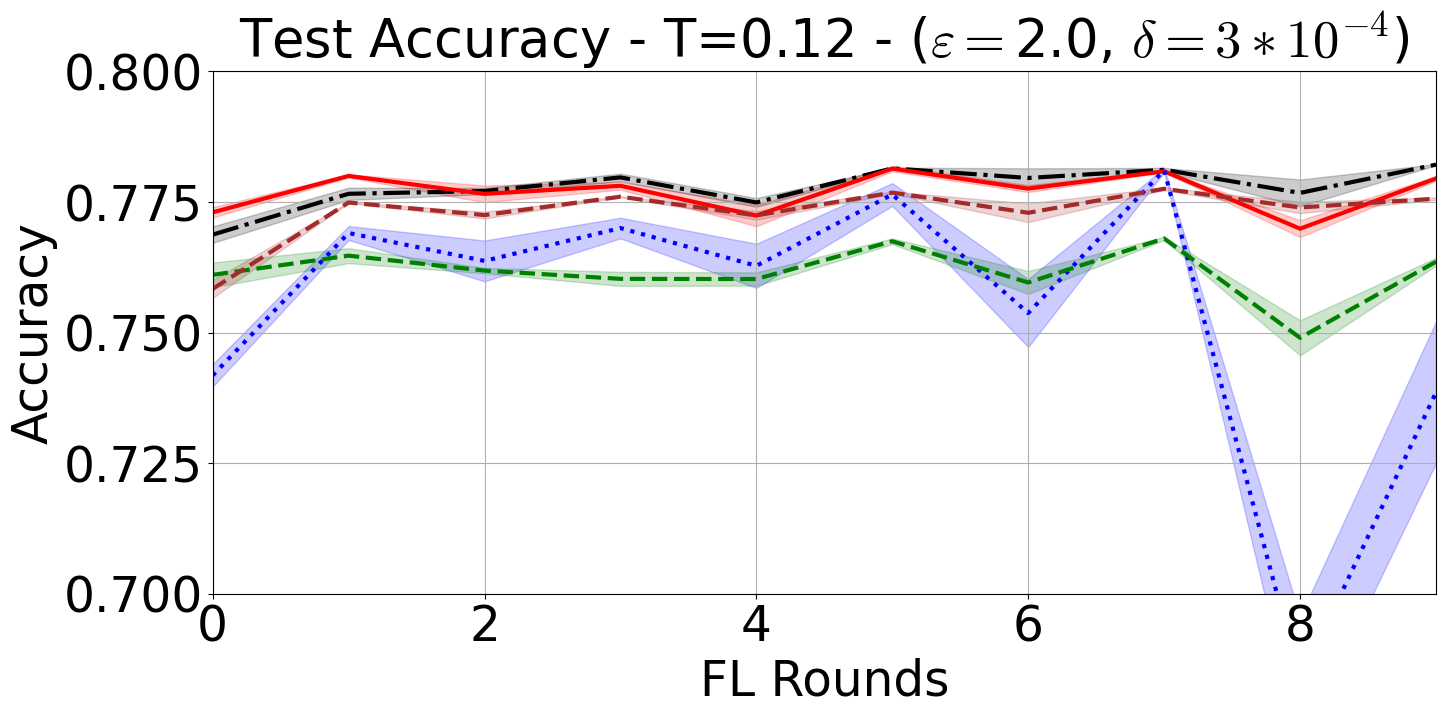}
         \caption{}
         \label{fig:income_2:accuracy_3}
     \end{subfigure}
     \hfill
     \begin{subfigure}[b]{0.18\textwidth}
         \centering
         \includegraphics[width=\textwidth]{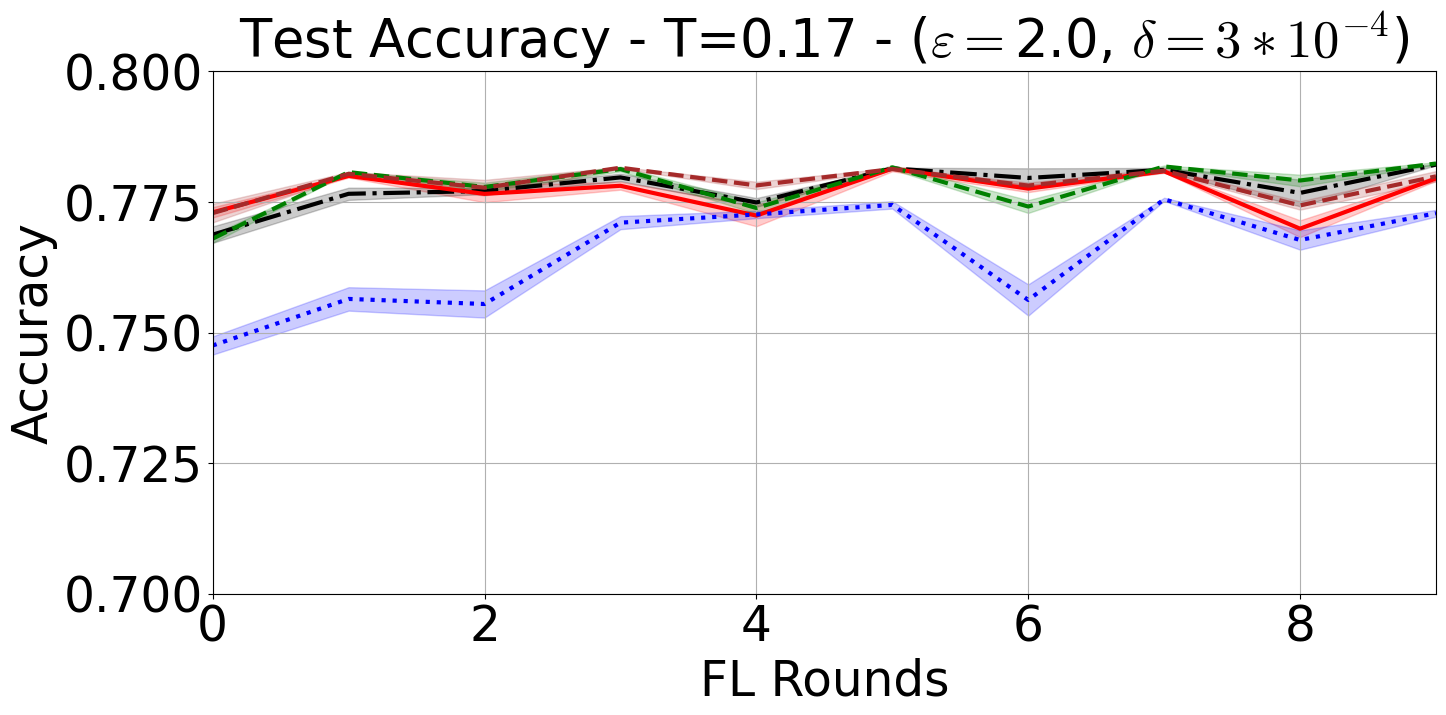}
         \caption{}
         \label{fig:income_2:accuracy_4}
     \end{subfigure}
     \hfill
     \begin{subfigure}[b]{0.18\textwidth}
         \centering
         \includegraphics[width=\textwidth]{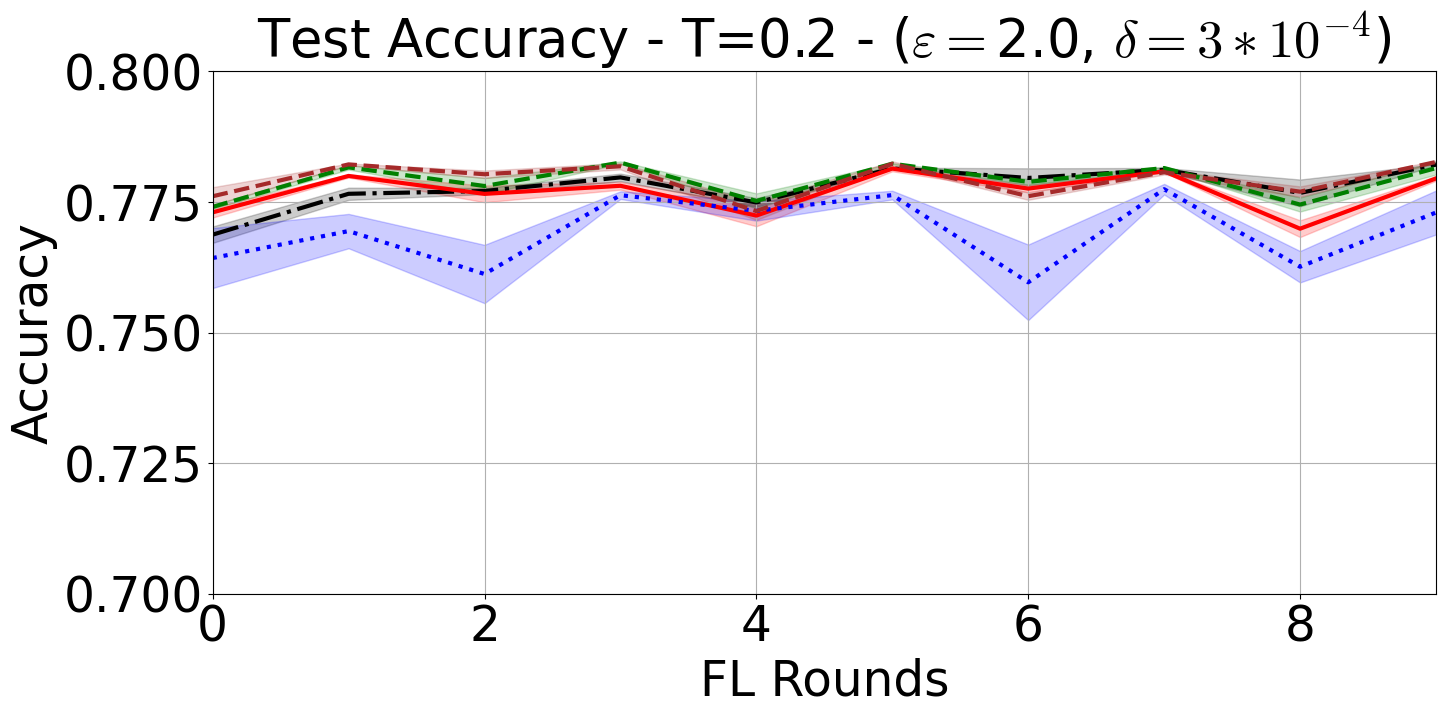}
         \caption{}
         \label{fig:income_2:accuracy_5}
     \end{subfigure}\\
     \vspace{0.30cm}
 
    \begin{subfigure}[b]{0.18\textwidth}
         \centering
         \includegraphics[width=\textwidth]{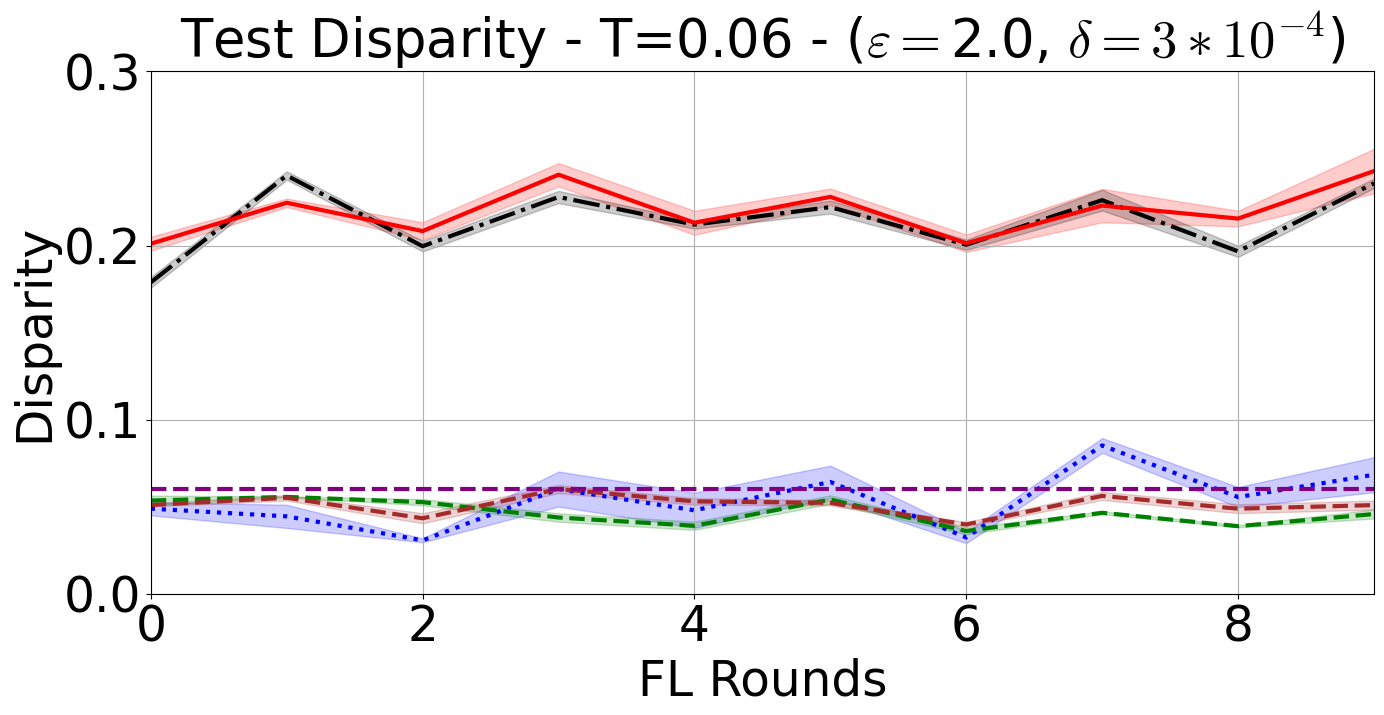}
         \caption{}
         \label{fig:income_2:disparity_1}
     \end{subfigure}
     \hfill
     \begin{subfigure}[b]{0.18\textwidth}
         \centering
         \includegraphics[width=\textwidth]{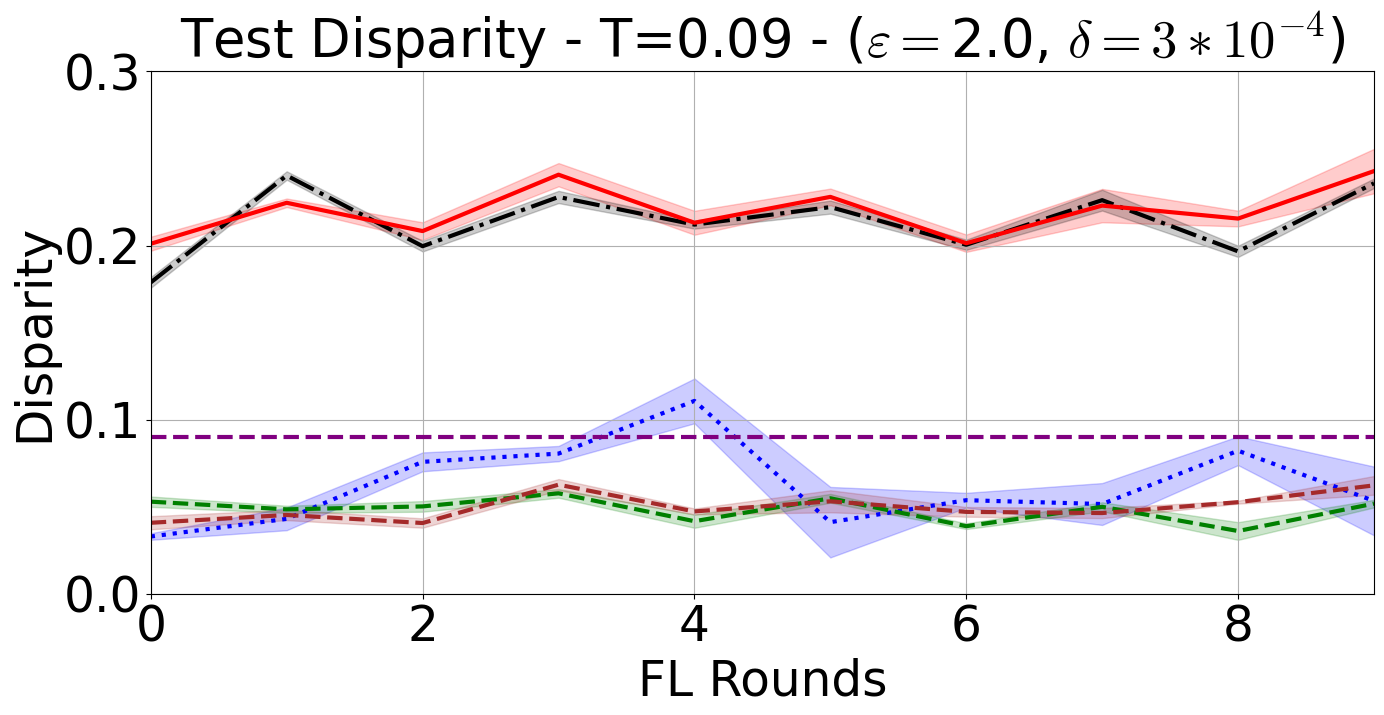}
         \caption{}
         \label{fig:income_2:disparity_2}
     \end{subfigure}
     \hfill
     \begin{subfigure}[b]{0.18\textwidth}
         \centering
         \includegraphics[width=\textwidth]{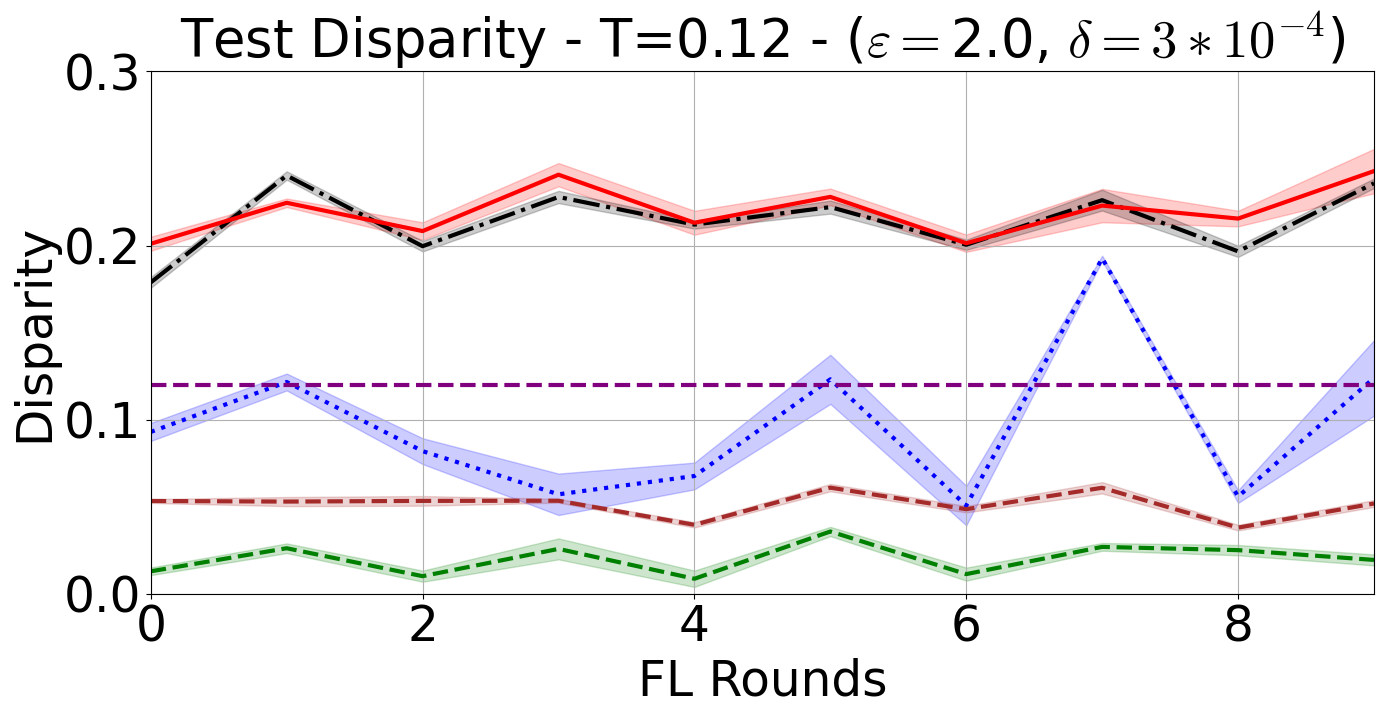}
         \caption{}
         \label{fig:income_2:disparity_3}
     \end{subfigure}
     \hfill
     \begin{subfigure}[b]{0.18\textwidth}
         \centering
         \includegraphics[width=\textwidth]{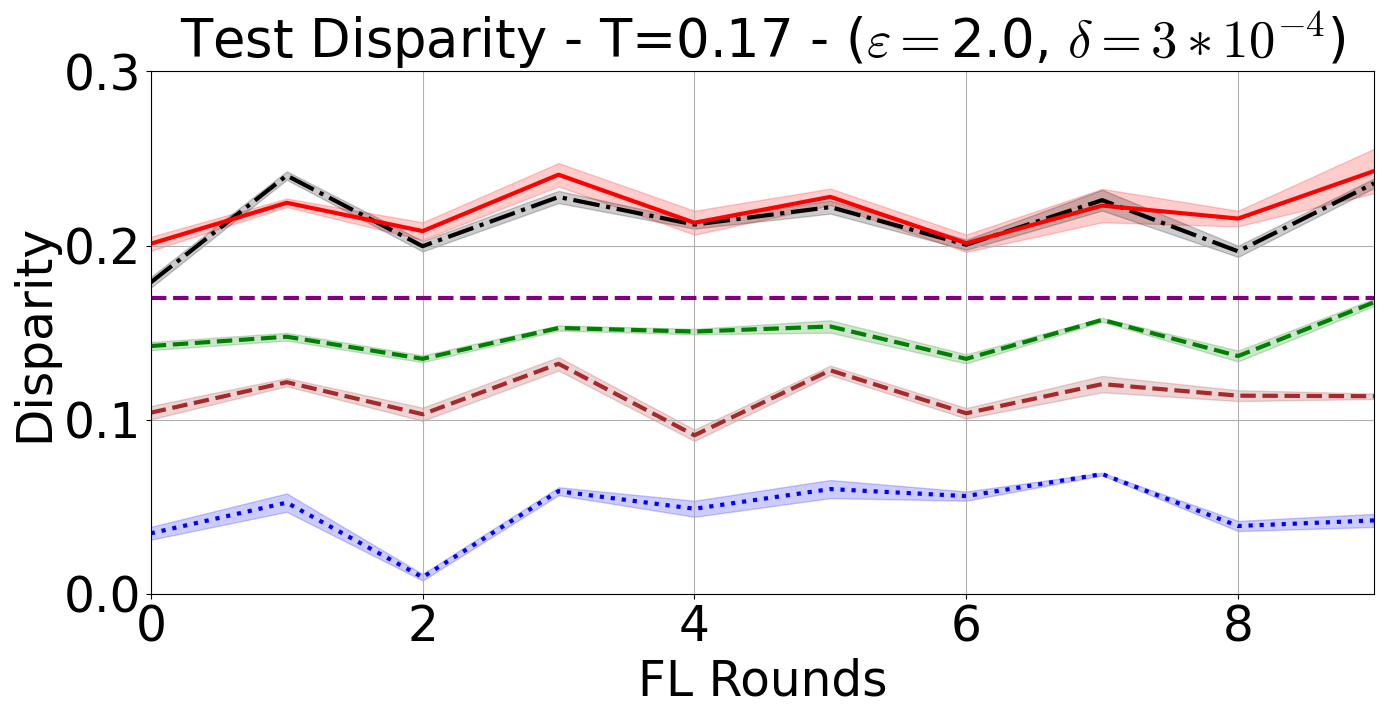}
         \caption{}
         \label{fig:income_2:disparity_4}
     \end{subfigure}
     \hfill
     \begin{subfigure}[b]{0.18\textwidth}
         \centering
         \includegraphics[width=\textwidth]{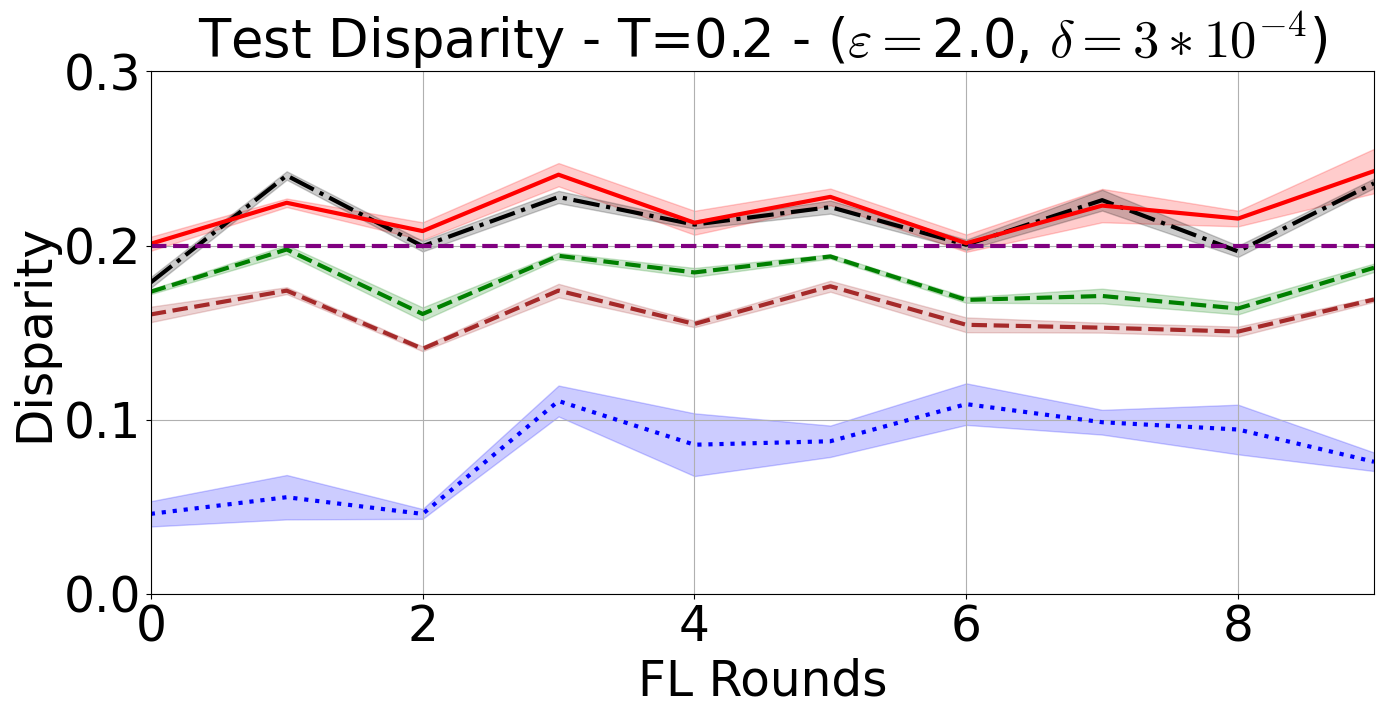}
         \caption{}
         \label{fig:income_2:disparity_5}
     \end{subfigure}\\
     \vspace{0.3cm}
     \includegraphics[width=0.60\linewidth]{images/legend_disparity.png}\\
    \vspace{0.15cm}
        \caption{Experiment with Income. Fairness parameters $T=0.06$, $T=0.08$, $T=0.12$, $T=0.18$ and $T=0.22$, privacy parameters ($\epsilon=0.5$, $\delta=7 \times  10^{-3}$). Figures ~\ref{fig:income_2:accuracy_1}, ~\ref{fig:income_2:accuracy_2},~\ref{fig:income_2:accuracy_3}, ~\ref{fig:income_2:accuracy_4} and ~\ref{fig:income_2:accuracy_5} show the test accuracy of training model while Figures ~\ref{fig:income_1:disparity_1}, ~\ref{fig:income_2:disparity_2}, ~\ref{fig:income_2:disparity_3}, ~\ref{fig:income_2:disparity_4} and ~\ref{fig:income_2:disparity_5} show the model disparity.}
        \label{fig:income_2}
\end{figure*}

Figures ~\ref{fig:income_1} and ~\ref{fig:income_2} show the results we obtained testing our proposed approach on the Income Dataset. 
We report in Table ~\ref{tab:results_income} a comparison of the different use of regularization and DP. In particular, here we compare the following experiments: Baseline, Only DP, Only Fairness, Fairness and DP both with Tunable and Fixed $\lambda$.

\begin{table*}[!t]
    \centering
    \captionsetup{justification=justified}
    \begin{tabular}{|c|c|c|c|c|}\hline
         & \textbf{($\varepsilon$, $\delta$)} & \textbf{$T$}& \textbf{Accuracy} & \textbf{Disparity}\\\hline
        Baseline & - & - & 0.782+-0.0& 0.236+-0.003\\\hline\hline
        
        DP & (1.0, $\mathrm{3*10^{-4}}$) & - & 0.78+-0.0& 0.224+-0.003\\\hline
        DP & (2.0, $\mathrm{3*10^{-4}}$) & - & 0.78+-0.0& 0.243+-0.013\\\hline\hline
        
        Fair & - & 0.06 & 0.775+-0.001& 0.051+-0.003\\\hline
        Fair & - & 0.09 & 0.775+-0.001& 0.062+-0.005\\\hline
        Fair & - & 0.12 & 0.776+-0.0& 0.052+-0.002\\\hline
        Fair & - & 0.17 & 0.78+-0.001& 0.113+-0.002\\\hline
        Fair & - & 0.2 & 0.783+-0.0& 0.169+-0.001\\\hline\hline
        
        PUFFLE (Fixed $\lambda$) & (1.0, $\mathrm{3*10^{-4}}$) & 0.06 & 0.776+-0.0& 0.055+-0.002\\\hline
        PUFFLE (Tunable $\lambda$) & (1.0, $\mathrm{3*10^{-4}}$) & 0.06 & 0.771+-0.001& 0.067+-0.002\\\hline
        PUFFLE (Fixed $\lambda$) & (2.0, $\mathrm{3*10^{-4}}$) & 0.06 & 0.775+-0.0& 0.046+-0.003\\\hline
        PUFFLE (Tunable $\lambda$) & (2.0, $\mathrm{3*10^{-4}}$) & 0.06 & 0.775+-0.001& 0.068+-0.01\\\hline\hline
        
        PUFFLE (Fixed $\lambda$) & (1.0, $\mathrm{3*10^{-4}}$) & 0.09 & 0.771+-0.001& 0.061+-0.003\\\hline
        PUFFLE (Tunable $\lambda$) & (1.0, $\mathrm{3*10^{-4}}$) & 0.09 & 0.772+-0.001& 0.05+-0.006\\\hline
        PUFFLE (Fixed $\lambda$) & (2.0, $\mathrm{3*10^{-4}}$) & 0.09 & 0.774+-0.001& 0.052+-0.002\\\hline
        PUFFLE (Tunable $\lambda$) & (2.0, $\mathrm{3*10^{-4}}$) & 0.09 & 0.761+-0.001& 0.053+-0.02\\\hline\hline

        PUFFLE (Fixed $\lambda$) & (1.0, $\mathrm{3*10^{-4}}$) & 0.12 & 0.776+-0.001& 0.112+-0.008\\\hline
        PUFFLE (Tunable $\lambda$) & (1.0, $\mathrm{3*10^{-4}}$) & 0.12 & 0.773+-0.0& 0.062+-0.004\\\hline
        PUFFLE (Fixed $\lambda$) & (2.0, $\mathrm{3*10^{-4}}$) & 0.12 & 0.764+-0.001& 0.019+-0.003\\\hline
        PUFFLE (Tunable $\lambda$) & (2.0, $\mathrm{3*10^{-4}}$) & 0.12 & 0.739+-0.014& 0.124+-0.022\\\hline\hline

        PUFFLE (Fixed $\lambda$) & (1.0, $\mathrm{3*10^{-4}}$) & 0.17 & 0.78+-0.0& 0.128+-0.004\\\hline
        PUFFLE (Tunable $\lambda$) & (1.0, $\mathrm{3*10^{-4}}$) & 0.17 & 0.764+-0.01& 0.082+-0.007\\\hline
        PUFFLE (Fixed $\lambda$) & (2.0, $\mathrm{3*10^{-4}}$) & 0.17 & 0.782+-0.0& 0.167+-0.002\\\hline
        PUFFLE (Tunable $\lambda$) & (2.0, $\mathrm{3*10^{-4}}$) & 0.17 & 0.773+-0.001& 0.042+-0.004\\\hline\hline

        PUFFLE (Fixed $\lambda$) & (1.0, $\mathrm{3*10^{-4}}$) & 0.2 & 0.781+-0.0& 0.193+-0.004\\\hline
        PUFFLE (Tunable $\lambda$) & (1.0, $\mathrm{3*10^{-4}}$) & 0.2 & 0.775+-0.002& 0.055+-0.02\\\hline
        PUFFLE (Fixed $\lambda$) & (2.0, $\mathrm{3*10^{-4}}$) & 0.2 & 0.781+-0.001& 0.187+-0.002\\\hline
        PUFFLE (Tunable $\lambda$) & (2.0, $\mathrm{3*10^{-4}}$) & 0.2 & 0.773+-0.004& 0.076+-0.005\\\hline
    \end{tabular}
    \caption{Income dataset: A comparison of the final test accuracy and the final test disparity across all the possible combinations. Specifically, the analysis encompasses two possible privacy parameters, namely ($\epsilon=1.0$, $3 \times  10^{-4}$) and ($\epsilon=2.0$, $3 \times  10^{-4}$) alongside three target fairness disparity, denoted as $T=0.06$, $T=0.09$, $T=0.12$, $T=0.17$, and $T=0.20$  corresponding to a reduction of 10\%, 25\%, 50\%, 65\%, 75\% with respect to the Baseline disparity.}
    \label{tab:results_income}
\end{table*}

\begin{figure*}
     \centering
     \captionsetup{justification=justified}
     \begin{subfigure}[b]{0.18\textwidth}
         \centering
         \includegraphics[width=\textwidth]{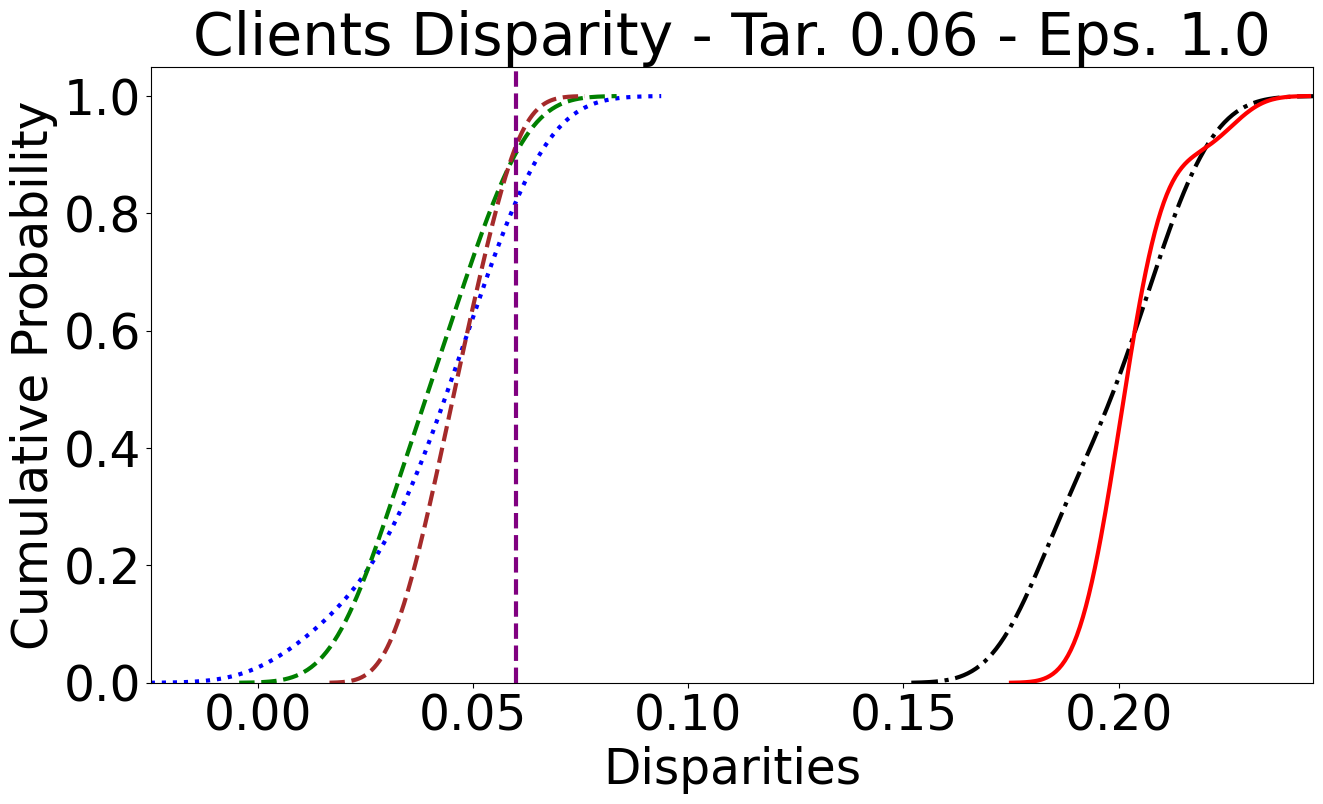}
         \caption{}
         \label{fig:income_local_1:cdf_1}
     \end{subfigure}
     \hfill
     \begin{subfigure}[b]{0.18\textwidth}
         \centering
         \includegraphics[width=\textwidth]{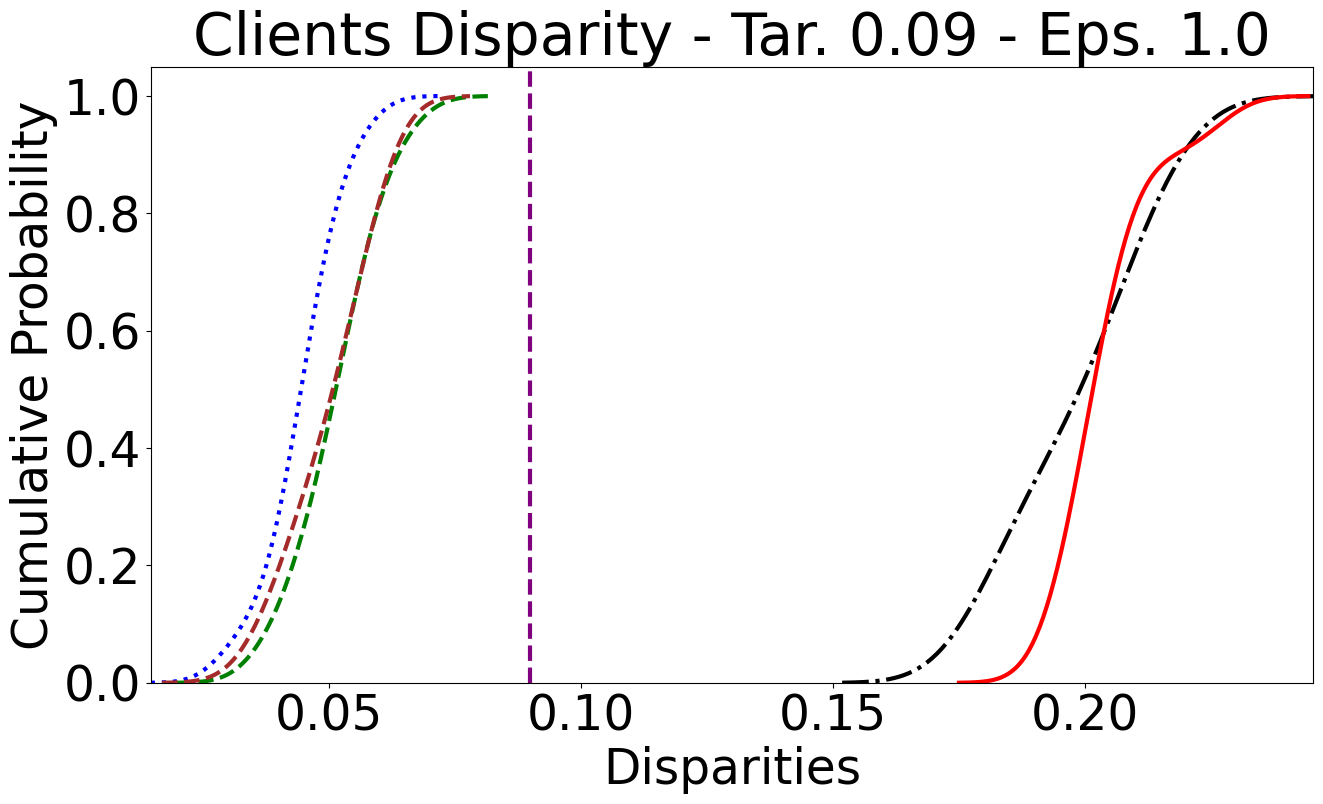}
         \caption{}
         \label{fig:income_local_1:cdf_2}
     \end{subfigure}
     \hfill
     \begin{subfigure}[b]{0.18\textwidth}
         \centering
         \includegraphics[width=\textwidth]{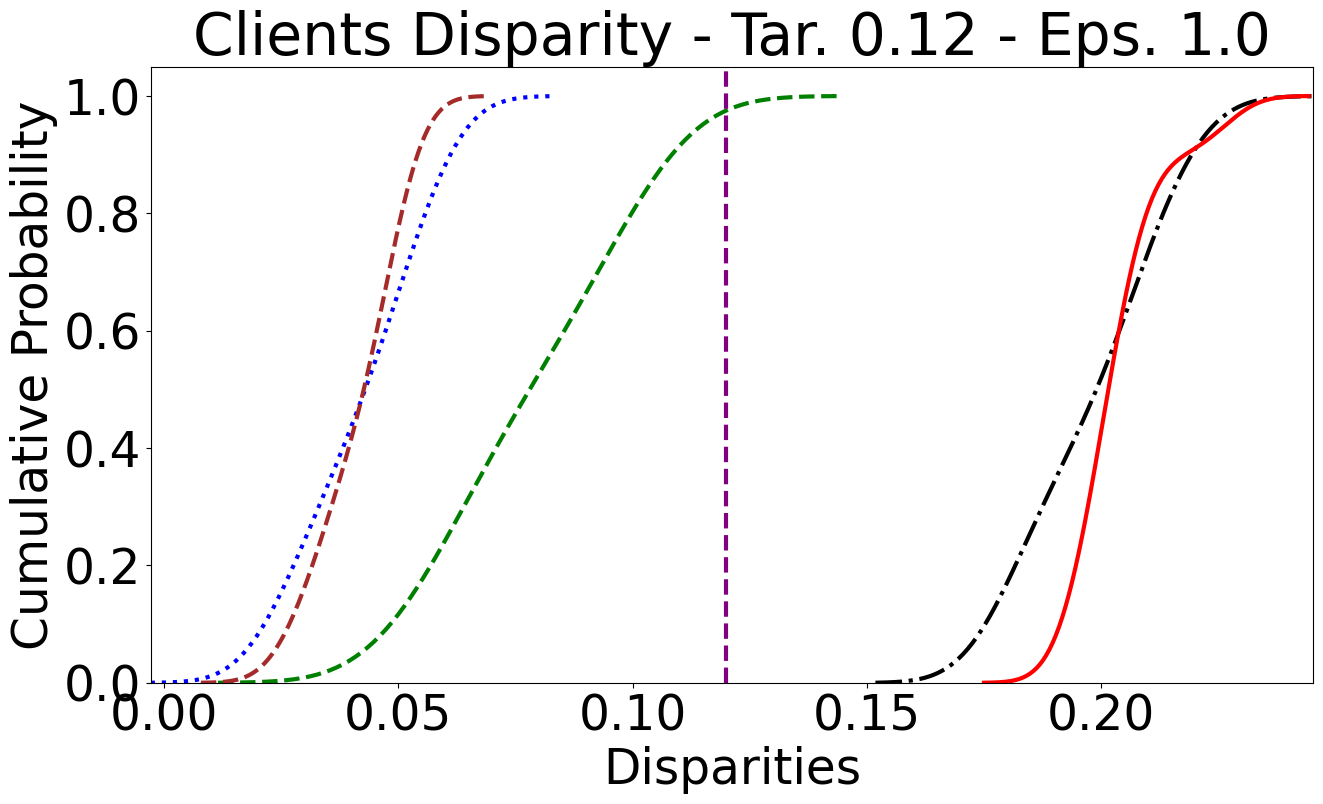}
         \caption{}
         \label{fig:income_local_1:cdf_3}
     \end{subfigure}
     \hfill
     \begin{subfigure}[b]{0.18\textwidth}
         \centering
         \includegraphics[width=\textwidth]{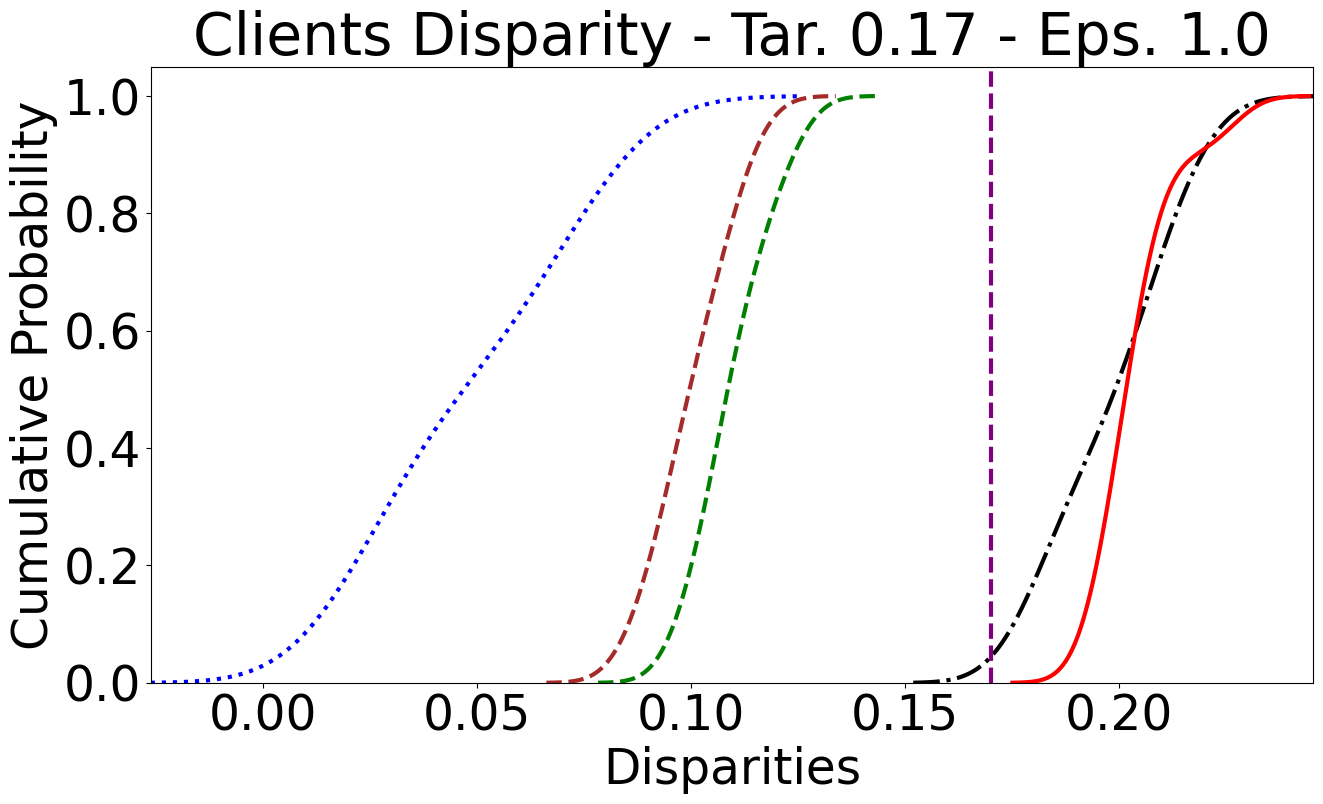}
         \caption{}
         \label{fig:income_local_1:cdf_4}
     \end{subfigure}
     \hfill
     \begin{subfigure}[b]{0.18\textwidth}
         \centering
         \includegraphics[width=\textwidth]{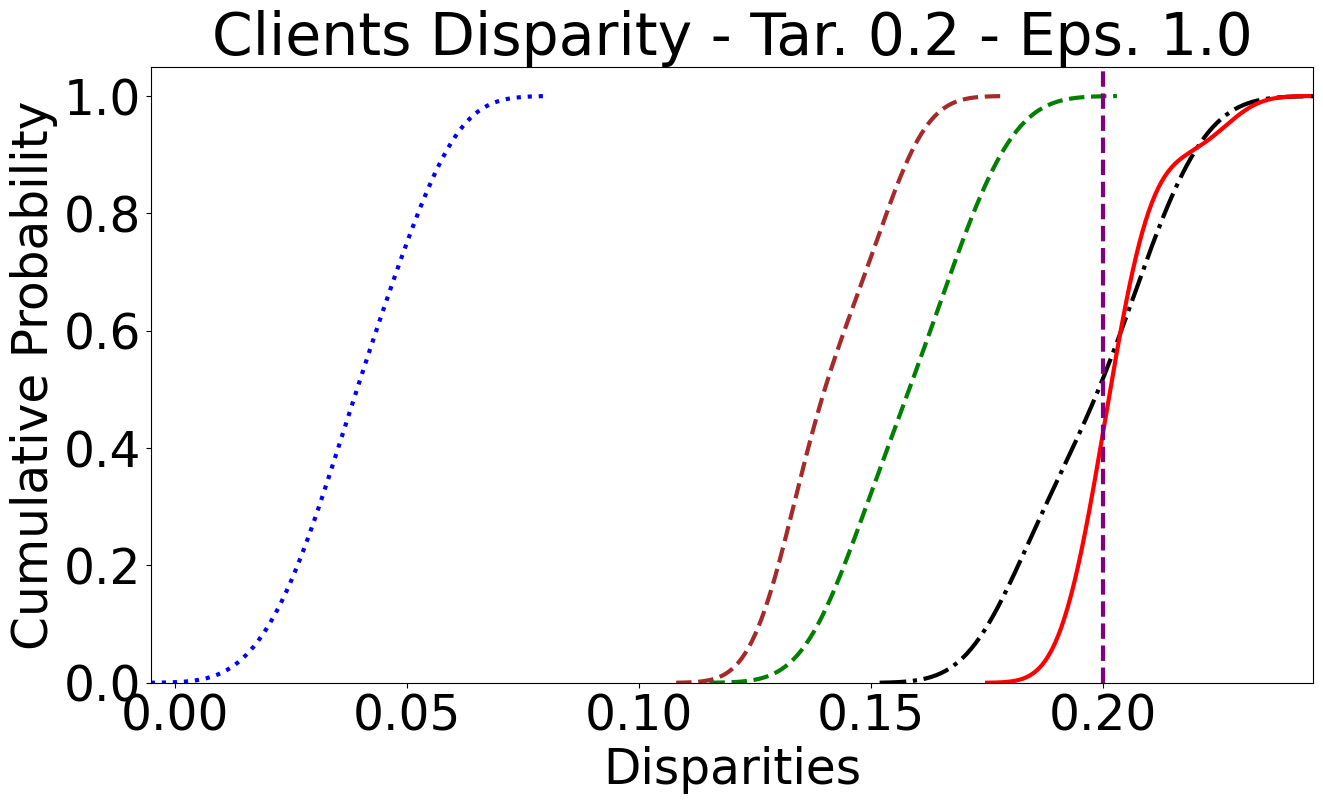}
         \caption{}
         \label{fig:income_local_1:cdf_5}
     \end{subfigure}\\
    \vspace{0.3cm}
     \includegraphics[width=0.60\linewidth]{images/legend_disparity.png}\\
    \vspace{0.15cm}
        \caption{Experiment with Income. Fairness parameters $T=0.06$, $T=0.08$, $T=0.12$, $T=0.18$ and $T=0.22$, privacy parameters ($\epsilon=0.5$, $\delta=7 \times  10^{-3}$). The Figures show the cumulative distribution function of the local disparities of the clients.}
        \label{fig:income_local_1}
\end{figure*}

\begin{figure*}
     \centering
     \captionsetup{justification=justified}
     \begin{subfigure}[b]{0.18\textwidth}
         \centering
         \includegraphics[width=\textwidth]{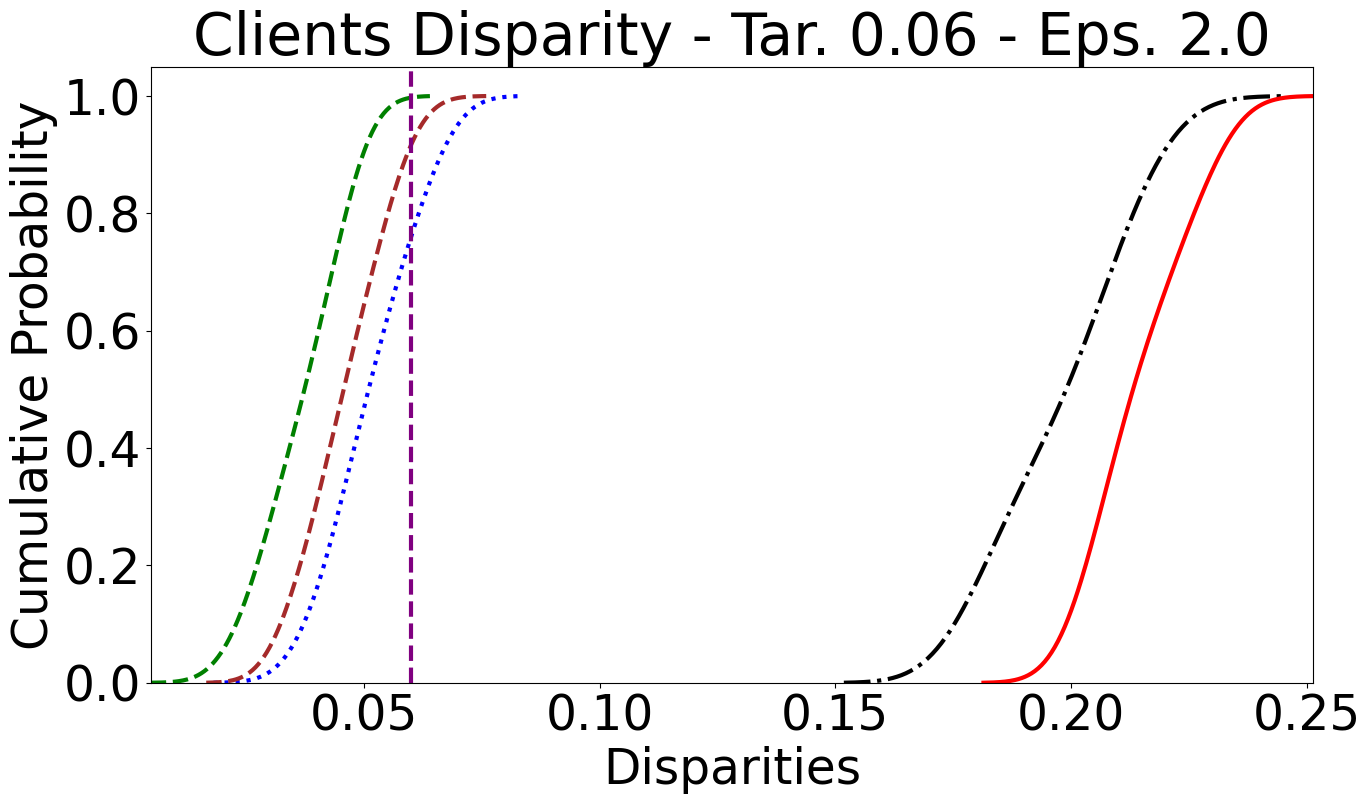}
         \caption{}
         \label{fig:income_local_2:cdf_1}
     \end{subfigure}
     \hfill
     \begin{subfigure}[b]{0.18\textwidth}
         \centering
         \includegraphics[width=\textwidth]{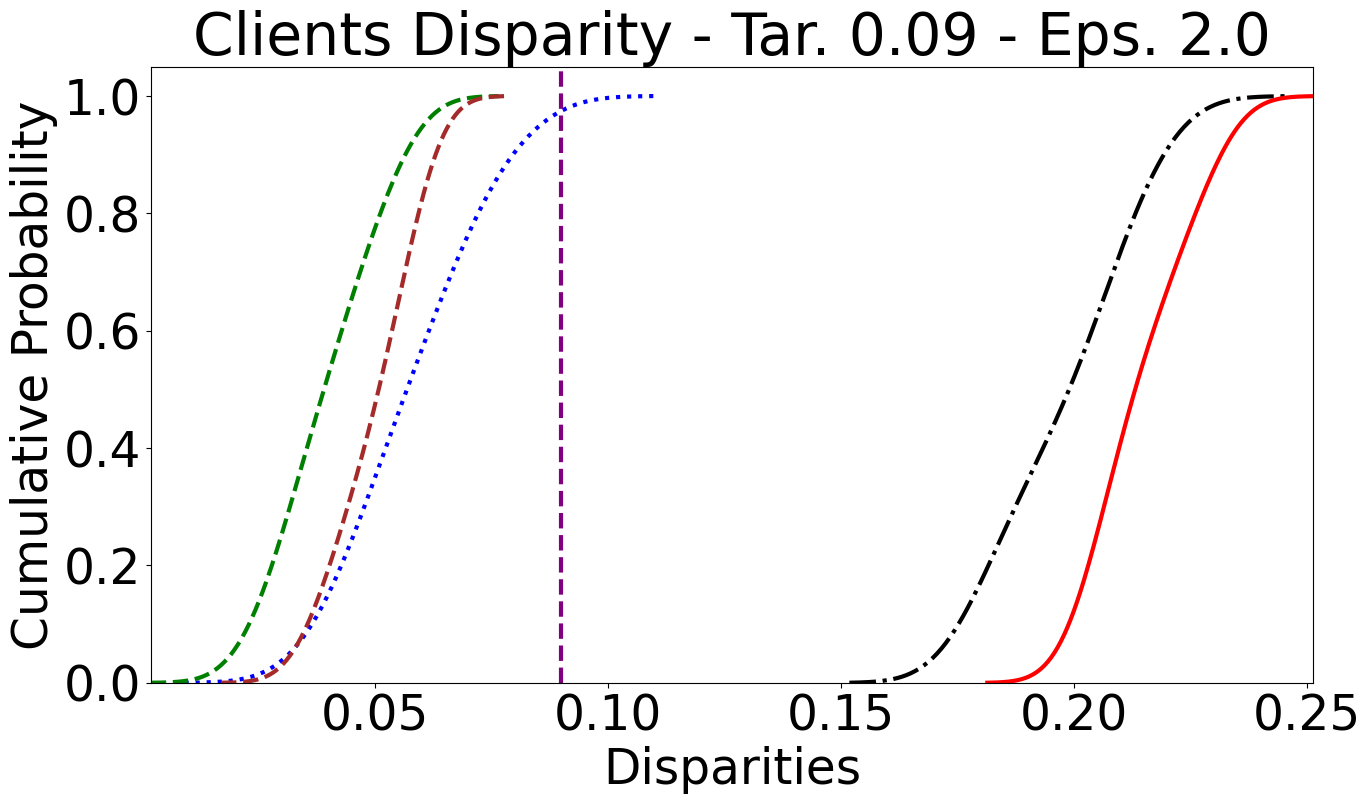}
         \caption{}
         \label{fig:income_local_2:cdf_2}
     \end{subfigure}
     \hfill
     \begin{subfigure}[b]{0.18\textwidth}
         \centering
         \includegraphics[width=\textwidth]{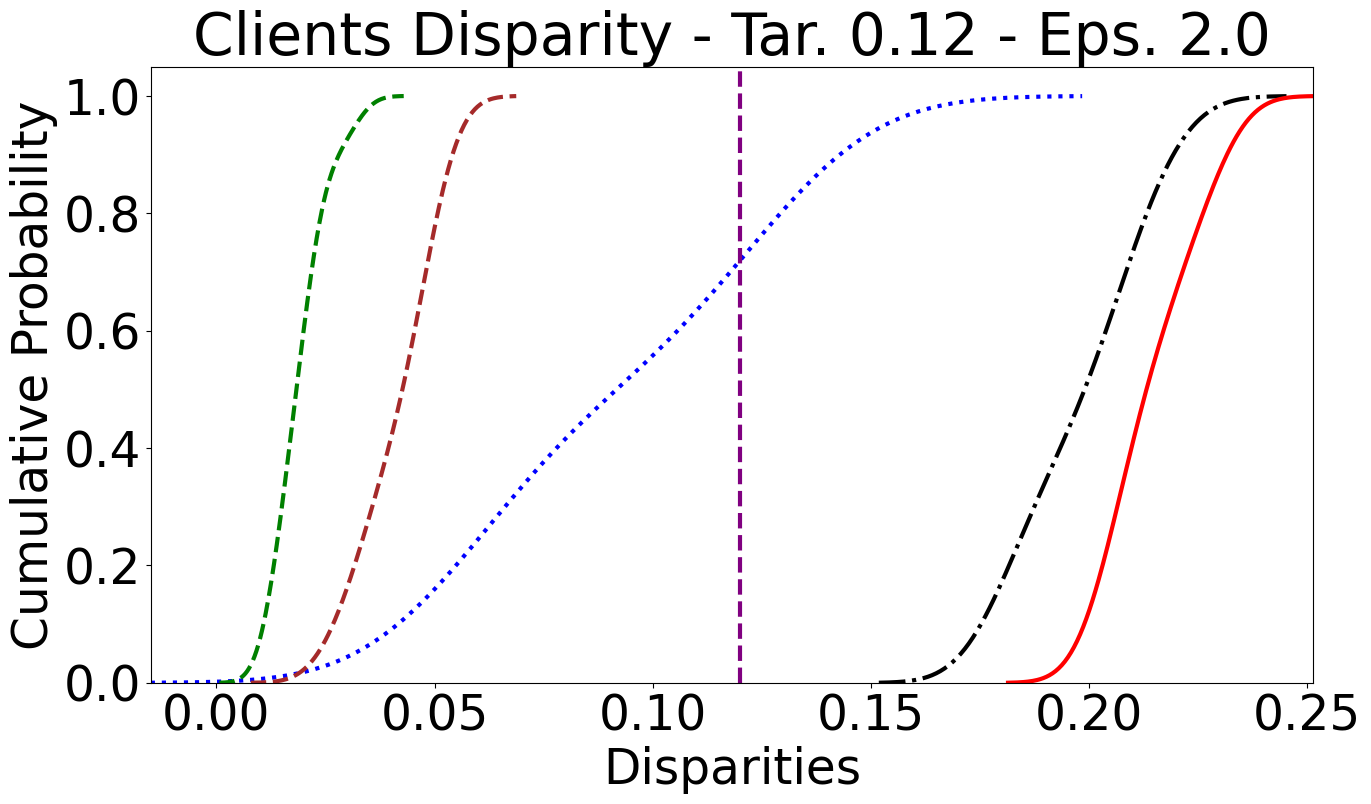}
         \caption{}
         \label{fig:income_local_2:cdf_3}
     \end{subfigure}
     \hfill
     \begin{subfigure}[b]{0.18\textwidth}
         \centering
         \includegraphics[width=\textwidth]{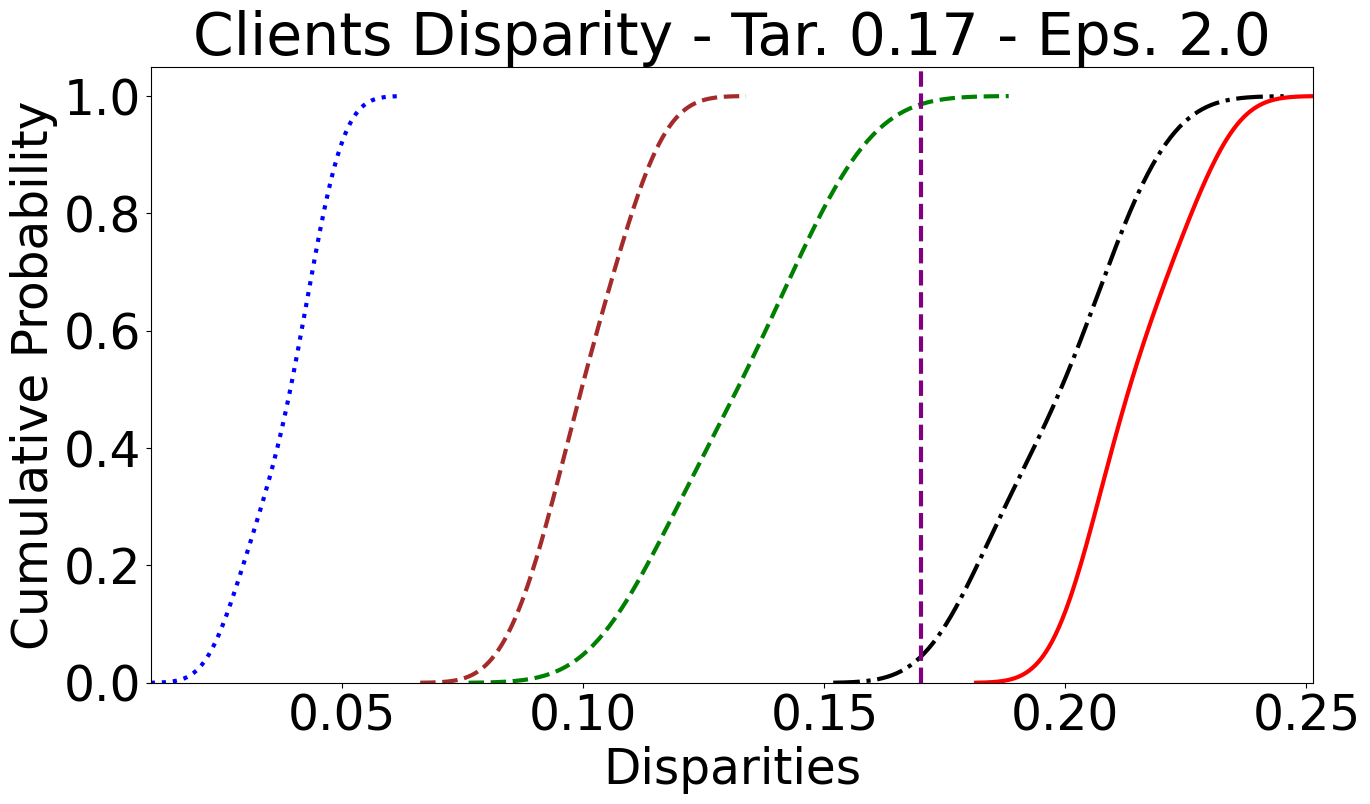}
         \caption{}
         \label{fig:income_local_2:cdf_4}
     \end{subfigure}
     \hfill
     \begin{subfigure}[b]{0.18\textwidth}
         \centering
         \includegraphics[width=\textwidth]{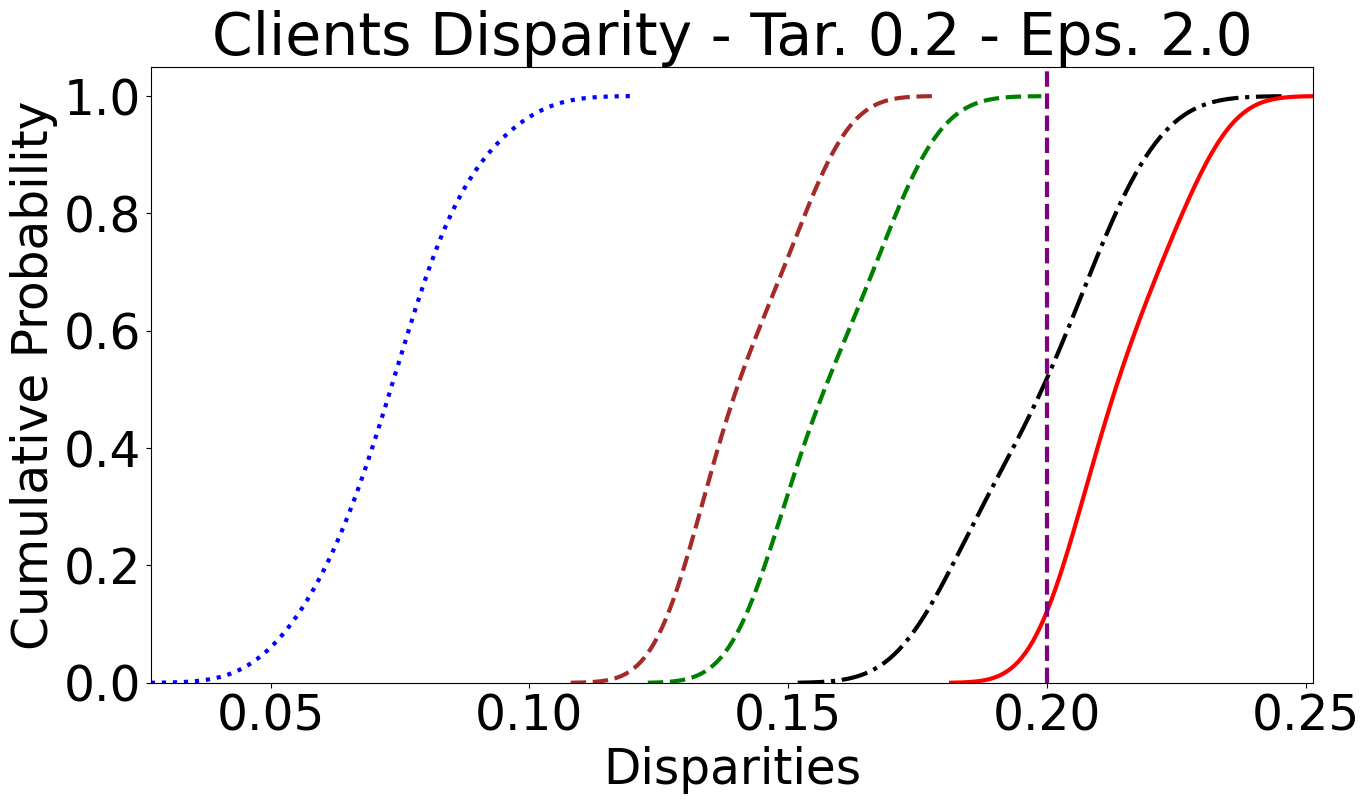}
         \caption{}
         \label{fig:income_local_2:cdf_5}
     \end{subfigure}\\
    \vspace{0.3cm}
     \includegraphics[width=0.60\linewidth]{images/legend_disparity.png}\\
    \vspace{0.15cm}
        \caption{Experiment with Income. Fairness parameters $T=0.06$, $T=0.08$, $T=0.12$, $T=0.18$ and $T=0.22$, privacy parameters ($\epsilon=0.5$, $\delta=7 \times  10^{-3}$). The Figures show the cumulative distribution function of the local disparities of the clients.}
        \label{fig:income_local_2}
\end{figure*}

We observe that the results in terms of unfairness mitigation are coherent with the ones that we have shown with the CelabA and Dutch Dataset. In this case, the task is easier than with Dutch and CelebA and, as we can see from Figures~\ref{fig:income_1:accuracy_1}, ~\ref{fig:income_1:accuracy_2}, ~\ref{fig:income_1:accuracy_3}, ~\ref{fig:income_1:accuracy_4},  ~\ref{fig:income_1:accuracy_5}, ~\ref{fig:income_2:accuracy_1}, ~\ref{fig:income_2:accuracy_2}, ~\ref{fig:income_2:accuracy_3}, ~\ref{fig:income_2:accuracy_4} and ~\ref{fig:income_2:accuracy_5} the use of DP and regularization do not degrade the model accuracy.
We report the plots of the CDF in Figures ~\ref{fig:income_local_1} and ~\ref{fig:income_local_2}.

\section{Research Ethics and Social Impact}

In this paper, we have addressed the problem of designing a \textit{trustworthy} machine learning process in a decentralized, federated learning setting. Our approach provides by design an acceptable trade-off between privacy, fairness and accuracy of the learned model. 
The scope of this research is to provide technological grounds and solutions for the development of ethical, lawful and trustworthy artificial intelligence compliant with regulations related to privacy protection, such as the EU GDPR~\cite{GDPR2016a}, and to Artificial Intelligence, such as the upcoming EU AI Act~\cite{madiega2021artificial}. 

AI systems based on machine learning models trained by exploiting our methodology demonstrated efficacy in avoiding the negative impact of learning biases from data. This work may therefore benefit underrepresented groups such as ethnic and gender minorities avoiding their algorithmic discrimination.   

Moreover, our approach enables the opportunity for humans to set both desired privacy and fairness requirements. As a consequence, in case data used for training the models contain sensitive information, the potential leakage of such information is mitigated through both the distributed setting that minimizes the sharing of data and the privacy mitigation strategy which ensures the confidentiality of any shared information.    

This work is the result of a collaboration of researchers with expertise ranging from privacy in big data and machine learning to responsible AI. This blending of expertise has allowed us to better understand the intricate landscape of AI ethics, considering not only the technical nuances but also the social implications. The presence of researchers with a mathematical background allowed us to apply differential privacy in our methodology in a rigorous way.

Lastly, the reproducibility of our results is guaranteed by: \textit{i)} the use of publicly available datasets; \textit{ii)} the sharing of the source code of our approach by a publicly available repository; \textit{iii)} the detailed description of each technical setting in the paper and appendix.

\end{document}